\documentclass[10pt,twocolumn,letterpaper]{article}

\usepackage{cvpr}              

\usepackage[dvipsnames]{xcolor}

\definecolor{cvprblue}{rgb}{0.21,0.49,0.74}
\usepackage[pagebackref,breaklinks,colorlinks,citecolor=cvprblue]{hyperref}
\usepackage{times}
\usepackage{epsfig}
\usepackage{graphicx}
\usepackage{amsmath}
\usepackage{amssymb}
\usepackage{booktabs}
\usepackage{multirow}
\usepackage{makecell}
\usepackage[ruled]{algorithm}
\usepackage{algorithmic}
\usepackage{latexsym}
\usepackage{hhline}
\usepackage{pifont}
\usepackage[accsupp]{axessibility} 
\def\paperID{4838} 
\def\confName{CVPR}
\def\confYear{2024}

\title{Real-World Efficient Blind Motion Deblurring via Blur Pixel Discretization}

\author{Insoo Kim\textsuperscript{1,}\textsuperscript{2} \;\;
Jae Seok Choi\textsuperscript{1} \;\;
Geonseok Seo\textsuperscript{1} \;\;
Kinam Kwon\textsuperscript{1} \;\;
Jinwoo Shin\textsuperscript{2}\thanks{Corresponding authors} \;\;
Hyong-Euk Lee\textsuperscript{1}\footnotemark[1] \\
\textsuperscript{1}Samsung Advanced Institute of Technology (SAIT), South Korea\\
\textsuperscript{2}Korea Advanced Institute of Science and Technology (KAIST), South Korea\\
}

\begin{document}
\maketitle
\begin{abstract}
As recent advances in mobile camera technology have enabled the capability to capture high-resolution images, such as 4K images, the demand for an efficient deblurring model handling large motion has increased. In this paper, we discover that the image residual errors, i.e., blur-sharp pixel differences, can be grouped into some categories according to their motion blur type and how complex their neighboring pixels are. Inspired by this, we decompose the deblurring (regression) task into blur pixel discretization (pixel-level blur classification) and discrete-to-continuous conversion (regression with blur class map) tasks. Specifically, we generate the discretized image residual errors by identifying the blur pixels and then transform them to a continuous form, which is computationally more efficient than naively solving the original regression problem with continuous values. Here, we found that the discretization result, i.e., blur segmentation map, remarkably exhibits visual similarity with the image residual errors. As a result, our efficient model shows comparable performance to state-of-the-art methods in realistic benchmarks, while our method is up to 10 times computationally more efficient.
\end{abstract}
    
\vspace{-0.2cm}
\section{Introduction}\label{sec:intro}
Motion blur is caused by the camera motion and object movement within the exposure time.
As the long exposure time is required to ensure the amount of light in low-light environments, it leads to significant blur degradation, which is a challenge to overcome.
Given blur images, the blind image deblurring procedure aims to reconstruct sharp images. 

In earlier years, many researchers have investigated accurate blur kernel estimations~\cite{motionflow,adaptivebasis,bd1,bluroperator,exposuretrajectory,bd2}.
Those kernel-based methods focus on estimating blur kernels and then exploiting them in the deconvolution process~\cite{richard_lucy,hyperlaplacian}, resulting in the final sharp images. They mainly train with synthesized motion blur images that may not hold in practice.
\begin{figure}[t]
  \centering
  \centerline{\includegraphics[width=8.6cm]{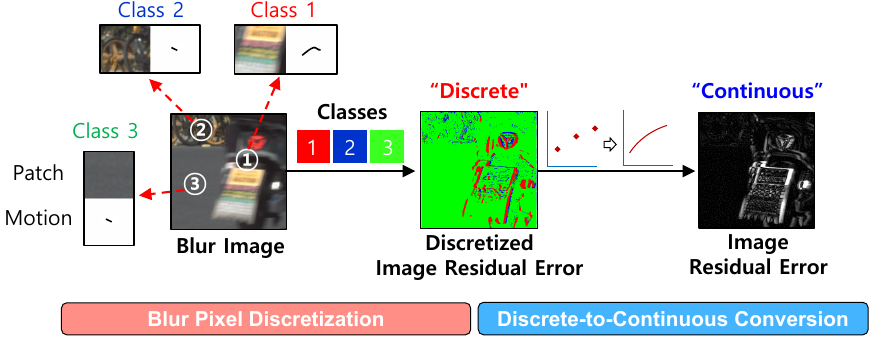}}
  \vspace{-0.3cm}
  \caption{Blind motion deblurring via blur pixel discretization. The image residual error, i.e., blur-sharp pixel differences is estimated by our blur pixel discretization and discrete-to-continuous conversion, whose result is used to produce final deblurred images.}  \label{fig:concept_proposed}
\vspace{-0.5cm}
\end{figure}
In recent years, kernel-free methods have been more widely studied in motion deblurring tasks~\cite{deblurgan,deblurganv2,deeprft,gopro,maxim,mimounet,mprnet,nafnet,reblur,restormer,srndeblurnet,stripformer,uformer,grl,fftformer,ufp,msdi}.
They achieve state-of-art performance despite their simplicity where they do not consider such blur kernels.
However, their performance highly relies on network capacities~\cite{mimounet,grl,maxim,uformer}, which may restrict from practical usage.
Some literature introduces self-generated prior information to further improve deblurring performance, but generating the prior information consumes additional huge computational costs~\cite{ufp,msdi}. 

As recent advances in camera technology have enabled the capability to capture high-resolution images, such as 4K images, dealing with large motion has become increasingly important. To deploy a deblurring model on resource-constrained devices (e.g., mobile devices), an efficient model capable of handling large motion is required. Nevertheless, these models have not been thoroughly explored. In our earlier experiments, we found that the small-scale model suffers from significant performance drops in the case of large motion scenarios as shown in Table~\ref{tbl:efficient}.
In fact, the small-scale deblurring model without any prior information may be susceptible to distortions.

\begin{figure}[t]
\begin{minipage}[!t]{.24\linewidth}
  \centering
  \centerline{\includegraphics[width=2.04cm]{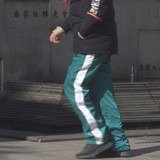}}
  \centerline{\includegraphics[width=2.04cm]{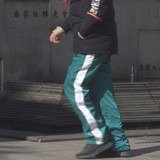}}
  \centerline{\small{(a)}}\medskip
\end{minipage}
\begin{minipage}[!t]{.24\linewidth}
  \centering
  \centerline{\includegraphics[width=2.04cm]{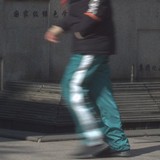}}
  \centerline{\includegraphics[width=2.04cm]{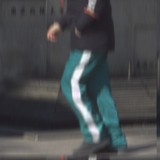}}
  \centerline{\small{(b)}}\medskip
\end{minipage}
\begin{minipage}[!t]{.24\linewidth}
  \centering
  \centerline{\includegraphics[width=2.04cm]{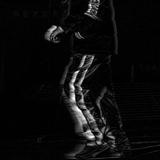}}
  \centerline{\includegraphics[width=2.04cm]{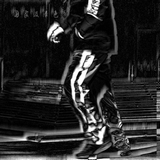}}
  \centerline{\small{(c)}}\medskip
\end{minipage}
\begin{minipage}[!t]{.24\linewidth}
  \centering
  \centerline{\includegraphics[width=2.04cm]{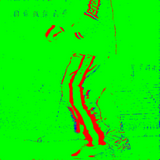}}
  \centerline{\includegraphics[width=2.04cm]{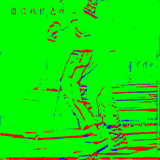}}
  \centerline{\small{(d)}}\medskip
\end{minipage}
\vspace{-0.3cm}
\caption{Visual comparison in the object motion blur (above) and uniform blur (below): (a) Sharp image, (b) Blur image, (c) Image residual error, and (d) Blur segmentation map. We utilize three classes and the colors in blur segmentation map indicate classes.}
\vspace{-0.5cm}
\label{fig:x3}
\end{figure}

In this paper, we rethink the deblurring task by decomposing the original regression problem into blur pixel discretization and discrete-to-continuous (D2C) conversion problems to explore an efficient deblurring model for practical usage, as shown in Fig.~\ref{fig:concept_proposed}.
Our intuition is that we build a discrete version of the ground truth (GT), namely, a discretized image residual error (e.g., discrete blur-sharp pixel differences) by identifying the blur pixels and then transform it into a continuous form, which is computationally more efficient than na\"ively solving the deblurring regression problem.
As shown in the upper side of Fig.~\ref{fig:x3} (c), the object movement regions (e.g., legs) highlight image residual errors, i.e., blur-sharp pixel differences, when compared to non-moving regions. In contrast, as shown in the lower side of Fig.~\ref{fig:x3} (c), despite the same blur for all pixels (i.e., uniform blur), the high-frequency region (e.g., edges) shows noticeable image residual errors rather than those of the low-frequency region (e.g., wall).
Namely, the image residual error is characterized by its motion type  (e.g., motion amount and direction) and spatial frequency (e.g., complexity of neighboring pixels).
Using this \textit{motion-frequency} property, the image residual error may be grouped into some categories, e.g., low or high levels.

\begin{figure*}[t]
  \centering
  \centerline{\includegraphics[width=17.2cm]{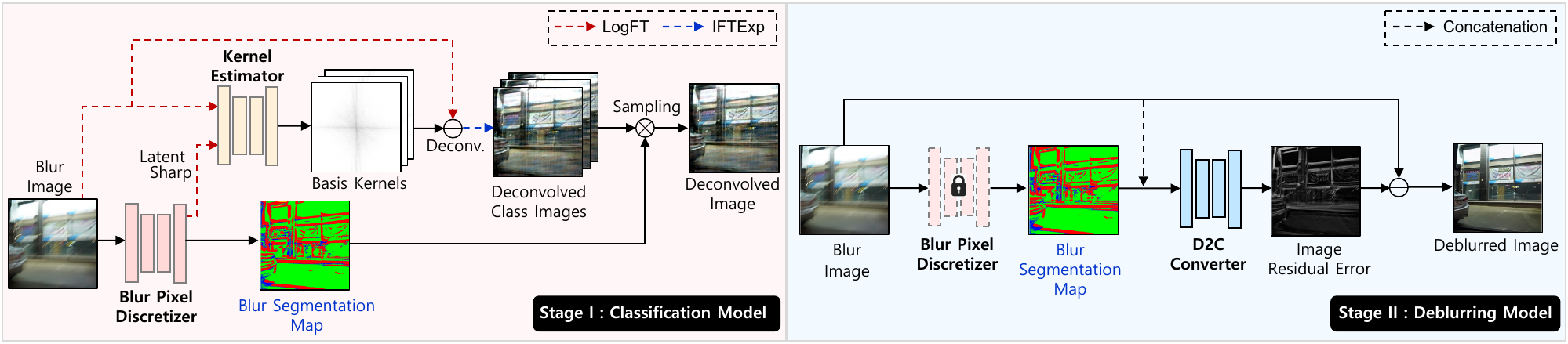}}
  \vspace{-0.3cm}
  \caption{Our network architectures. The proposed method consists of two stages and each stage is shaded in different colors. In the first stage, we train our blur pixel discretizer and kernel estimator. In the second stage, we proceed to train our D2C converter with the frozen blur pixel discretizer. LogFT indicates the logarithmic fourier transform and IFTExp means the inverse logarithmic fourier transform.}
  \label{fig:proposed_method}
\vspace{-0.5cm}
\end{figure*}

Inspired by those observations, we propose a new deblurring scheme that decomposes the deblurring model into \textit{blur pixel discretizer} and \textit{D2C converter} as shown in Fig.~\ref{fig:proposed_method}.
In the first stage, our blur pixel discretizer is trained to yield the discretized image residual error at a low computational cost, which is referred to as \textit{blur segmentation map}.
In the second stage, the D2C converter efficiently transforms the discrete version of image residual error into a continuous form, leading to better deblurring results.
In particular, the blur pixels are identified by the following steps:
(1) To reflect the nature of the image residual error, e.g., motion-frequency property, we first predict a \textit{motion-related} several basis kernels and they are used for a deconvolution procedure with the \textit{frequency-related} neighboring pixels, resulting in a set of deconvolved images, i.e., deconvolved class images.
(2) We introduce a blur segmentation map that contains a per-pixel class index. Based on this information, we perform pixel-wise sampling from the deconvolved class images to predict the optimal deconvolved image.
(3) As a result, all blur pixels are optimally categorized into classes through our blur segmentation map, which visually aligns with the image residual error as shown in Fig.~\ref{fig:x3} (c) and (d).
Furthermore, we pay attention to the logarithmic fourier space to simplify the relationship between blur and sharp images so that the basis kernels are easily trained in our kernel estimator as shown in Fig.~\ref{fig:proposed_method}.
Our contributions are summarized as follows:
\begin{itemize}
    \item We propose a new deblurring scheme that decomposes a deblurring regression task into simpler tasks: blur pixel discretization and D2C conversion tasks, which is computationally more efficient than na\"ively solving the deblurring regression problem.
    \item Our blur pixel discretizer produces the blur segmentation map, which reflects the nature of the image residual error. Hence, the proposed method can be interpreted as deblurring with GT-like information, leading to better deblurring results at a low computational cost.
    \item Our efficient deblurring model demonstrates its competitiveness not only even with a reduction of up to 10$\times$ in the computational cost compared to larger deblurring methods in realistic benchmarks but also in commercial applications such as Samsung EnhanceX and Google Unblur.
\end{itemize}
\begin{figure}[t]
\begin{minipage}[!t]{.32\linewidth}
  \centering
  \centerline{\includegraphics[width=2.6cm]{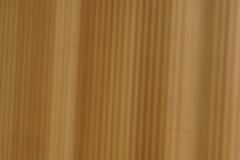}}
  \centerline{\includegraphics[width=2.6cm]{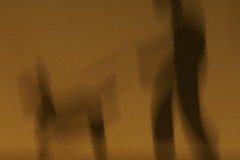}}
  \centerline{\small{(a)}}\medskip
\end{minipage}
\begin{minipage}[!t]{.32\linewidth}
  \centering
  \centerline{\includegraphics[width=2.6cm]{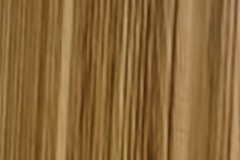}}
  \centerline{\includegraphics[width=2.6cm]{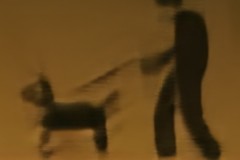}}
  \centerline{\small{(b)}}\medskip
\end{minipage}
\begin{minipage}[!t]{.32\linewidth}
  \centering
  \centerline{\includegraphics[width=2.6cm]{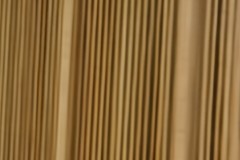}}
  \centerline{\includegraphics[width=2.6cm]{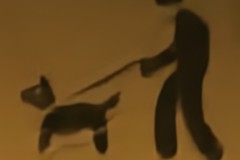}}
  \centerline{\small{(c)}}\medskip
\end{minipage}
\vspace{-0.3cm}
\caption{Visual comparison results on the efficient models: (a) Blur image, (b) NAFNet~\cite{nafnet}, and (c) Proposed method. The computational cost of both models is around $16$ GMACs. NAFNet is vulnerable to distortions since it has no blur class information, whereas our method produces more natural deblurring results.}
\vspace{-0.3cm}
\label{fig:distortion}
\end{figure}
\section{Related Works} \label{sec:related}
\vspace{0.02in}
\setlength{\parindent}{0in}\textbf{Kernel-based methods.}
In blind motion deblurring tasks, kernel-based methods~\cite{bd1,bd2,deepkernel1} directly learn a per-pixel kernel and then produce sharp images using the existing non-blind deblurring methods~\cite{richard_lucy,hyperlaplacian,deepwiener}.
To estimate the blur kernels easier
, a motion flow estimation~\cite{motionflow} is studied to produce a motion flow map. Similarly, the exposure trajectory estimation framework~\cite{exposuretrajectory} is investigated to estimate a set of motion offsets.
The adaptive basis decomposition scheme~\cite{adaptivebasis} introduces a set of pixel-shared basis kernels and a set of pixel-wise mixing coefficients to efficiently estimate a per-pixel kernel.
Overall, those kernel-based methods aim to generate sharp images by utilizing motion flows, trajectories, and basis kernels at a large cost, whereas ours focuses on providing per-pixel blur class information (distinct from the motion flows, trajectories and basis kernels) at a small cost.

\vspace{0.02in}
\setlength{\parindent}{0in}\textbf{Kernel-free methods.} The kernel-free methods~\cite{deepkernelfree,gopro,mimounet,nafnet} treat the image deblurring tasks as image-to-image translation tasks that map from blur to sharp in realistic blur datasets~\cite{realblur,gopro}. Some pioneering works~\cite{gopro,srndeblurnet} introduce the coarse-to-fine strategy that gradually recovers the sharp image from low-resolution to high-resolution images.
Some frameworks that use multi-input and multi-output U-Net~\cite{unet} are proposed to handle the coarse-to-fine strategy in an efficient way~\cite{mimounet,deeprft,mprnet}. Transformer-based image deblurring methods~\cite{restormer,uformer,stripformer,fftformer,grl} are studied to recover more image details by presenting their own ways to overcome the quadratic complexity of the self-attention modules. Some literature investigates the network architecture design to capture local and global information effectively~\cite{maxim,restormer,mprnet, nafnet}. 
Recently, some methods introduce additional self-generated priors to improve deblurring performance~\cite{msdi,ufp}. However, their priors are not directly related to the image residual error (e.g., GT) which may limit the performance improvement.
On the other hand, our method generates a discretized image residual error (e.g. discrete GT), which results in better performance.
\section{Real-World Efficient Motion Deblurring}
\label{sec:method}
\setlength{\parindent}{0.4cm}
In this section, we describe a new deblurring scheme that consists of blur pixel discretizer and discrete-to-continuous (D2C) converter.
First, we explore a new perspective of motion deblurring tasks in Section~\ref{sec:formulation}. Then, we introduce a blur pixel discretizer that yields blur segmentation map in Section~\ref{sec:kest}. In Section~\ref{sec:deblurring}, a D2C converter with blur segmentation map is explained. Finally, we present the implementation details of our method in Section~\ref{sec:implementation}.

\subsection{A new perspective of motion deblurring} \label{sec:formulation}
As discussed in Section~\ref{sec:intro}, the image residual error is determined by the motion (e.g., motion amount and direction) and frequency properties (e.g., complexity of neighboring pixels). To produce meaningful discrete image residual errors, we revisit a kernel-based method that is able to estimate motion kernels and consider neighboring pixels via a deconvolution procedure, resulting in sharp images.

Consider a dataset $\mathcal{D}=\{(x,y)\}$, which contains a pair of blur image $y$ and sharp image $x$ in the continuous domain $\mathcal{X}$. Given a uniform blur kernel $k$, the sharp image $x$ is reconstructed by $x=y\odot k$ where $\odot$ is a deconvolution operation.
To extend it to non-uniform deblurring, the sharp pixel value $x_i$ can be obtained by a per-pixel deconvolution, i.e., $x_{i}=\mathcal{P}_iy\odot k^{(i)}$ where $i=1,2,\ldots,N$ is the pixel index, $\mathcal{P}_i$ is an operator to extract the patch (i.e., \textit{neighboring pixels}) at a pixel $i$ and $k^{(i)}$ is a per-pixel kernel at a pixel $i$.
Since we aim to categorize the blur pixels into classes, we instead consider a set of uniform motion basis kernels, $k=\{k^{(r)}\}^{R}_{r=1}$ where $r$ is the basis kernel index. Given the basis kernels $k$, a set of deconvolved images, i.e., deconvolved class images $\nu=\{\nu^{(1)},\nu^{(2)},\ldots,\nu^{(R)}\}$ where $\nu^{(r)}=\{\nu^{(r)}_i\}^{N}_{i=1}$ are computed by $\nu=y\odot k$, and a deconvolved class pixel value $\nu^{(r)}_i$ is defined by
\begin{equation} \label{eq:uniform_basis}
\nu^{(r)}_i=\mathcal{P}_iy\odot k^{(r)}.
\end{equation}
To determine which class $r$ is more relevant to a given pixel $i$, we introduce blur segmentation map $\rho=\{\rho_{i}\}^{N}_{i=1}$ that contains a per-pixel class index $\rho_i\in\{1,2,\ldots,R\}$ in the discrete domain $\mathcal{Z}$.
Then, the deconvolved image $\tilde{x}=\{\nu^{(\rho_{i})}_i\}^N_{i=1}$ is recovered by combining the relevant deconvolved class pixel values using individual per-pixel class indices.
Since the deconvolved class pixel values are determined by the motion-related basis kernels $k^{(r)}$ and frequency-related neighboring pixels $\mathcal{P}_iy$ in~\eqref{eq:uniform_basis}, our blur segmentation map implicitly reflects the characteristics of the image residual errors, as shown in Fig.~\ref{fig:x3} (c) and (d).

\begin{figure*}[htb]
  \centering
\begin{minipage}[!t]{.260\linewidth}
  \centering
  \centerline{\includegraphics[width=4.60cm]{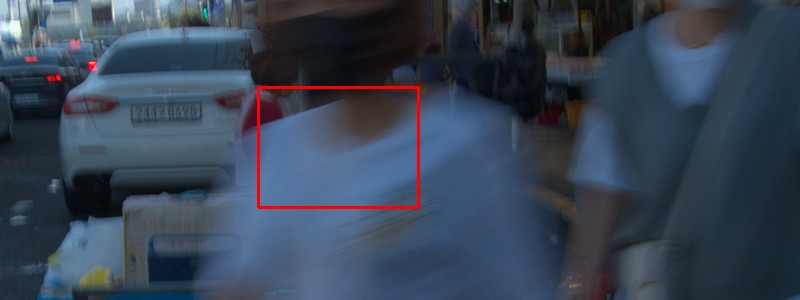}}
\end{minipage}
\begin{minipage}[!t]{.130\linewidth}
  \centering
  \centerline{\includegraphics[width=2.30cm]{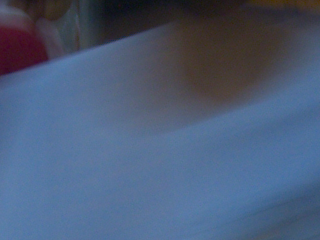}}
\end{minipage}
\begin{minipage}[!t]{.130\linewidth}
  \centering
  \centerline{\includegraphics[width=2.30cm]{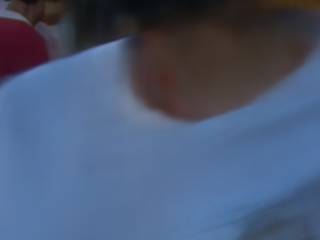}}
\end{minipage}
\begin{minipage}[!t]{.130\linewidth}
  \centering
  \centerline{\includegraphics[width=2.30cm]{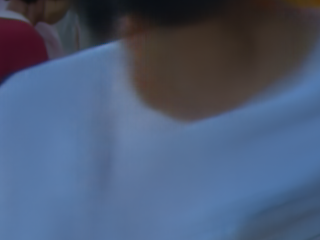}}
\end{minipage}
\begin{minipage}[!t]{.130\linewidth}
  \centering
  \centerline{\includegraphics[width=2.30cm]{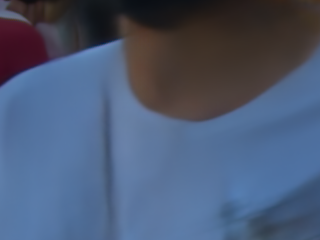}}
\end{minipage}
\begin{minipage}[!t]{.130\linewidth}
  \centering
  \centerline{\includegraphics[width=2.30cm]{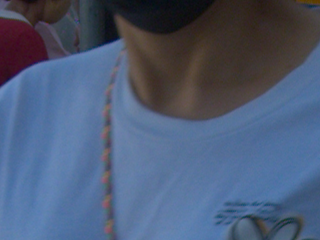}}
\end{minipage}
\vspace{-0.35cm}
\end{figure*}
\begin{figure*}[htb]
  \centering
\begin{minipage}[!t]{.260\linewidth}
  \centering
  \centerline{\includegraphics[width=4.60cm]{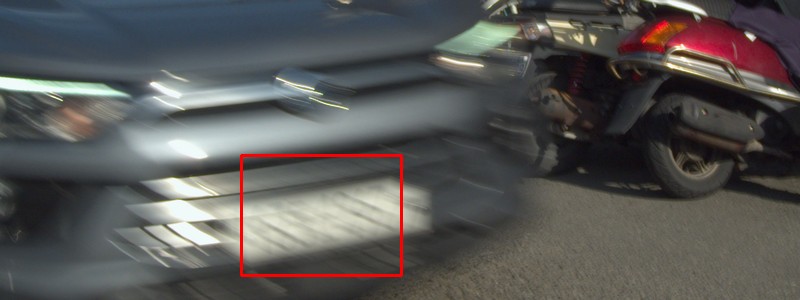}}
\end{minipage}
\begin{minipage}[!t]{.130\linewidth}
  \centering
  \centerline{\includegraphics[width=2.30cm]{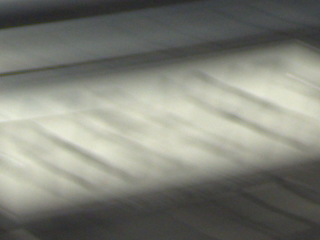}}
\end{minipage}
\begin{minipage}[!t]{.130\linewidth}
  \centering
  \centerline{\includegraphics[width=2.30cm]{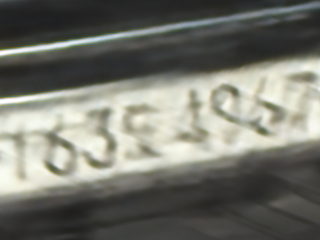}}
\end{minipage}
\begin{minipage}[!t]{.130\linewidth}
  \centering
  \centerline{\includegraphics[width=2.30cm]{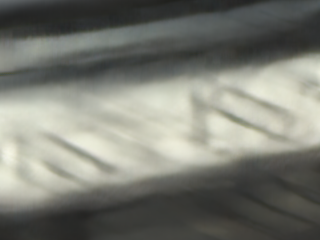}}
\end{minipage}
\begin{minipage}[!t]{.130\linewidth}
  \centering
  \centerline{\includegraphics[width=2.30cm]{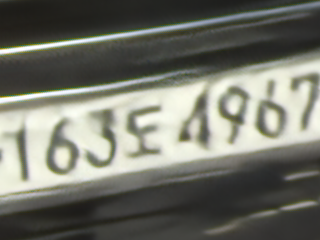}}
\end{minipage}
\begin{minipage}[!t]{.130\linewidth}
  \centering
  \centerline{\includegraphics[width=2.30cm]{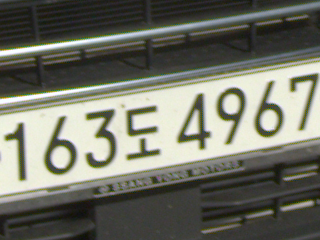}}
\end{minipage}
\vspace{-0.35cm}
\end{figure*}
\begin{figure*}[!htb]
  \centering
\begin{minipage}[!t]{.260\linewidth}
  \centering
  \centerline{\includegraphics[width=4.60cm]{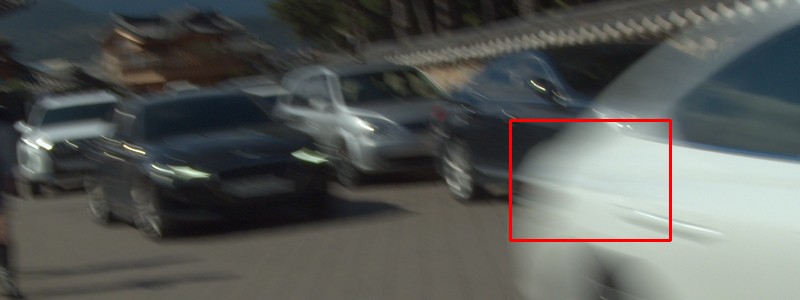}}
\end{minipage}
\begin{minipage}[!t]{.130\linewidth}
  \centering
  \centerline{\includegraphics[width=2.30cm]{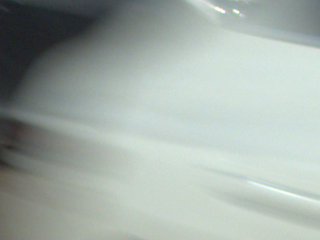}}
\end{minipage}
\begin{minipage}[!t]{.130\linewidth}
  \centering
  \centerline{\includegraphics[width=2.30cm]{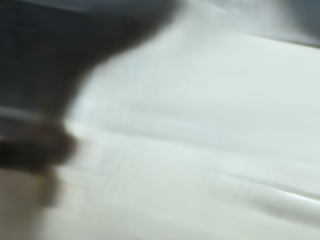}}
\end{minipage}
\begin{minipage}[!t]{.130\linewidth}
  \centering
  \centerline{\includegraphics[width=2.30cm]{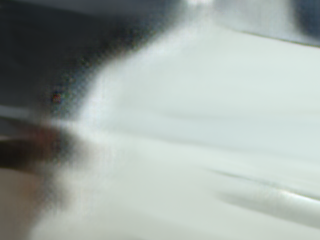}}
\end{minipage}
\begin{minipage}[!t]{.130\linewidth}
  \centering
  \centerline{\includegraphics[width=2.30cm]{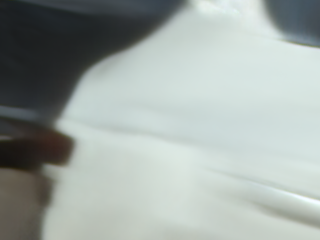}}
\end{minipage}
\begin{minipage}[!t]{.130\linewidth}
  \centering
  \centerline{\includegraphics[width=2.30cm]{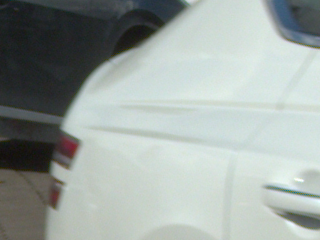}}
\end{minipage}
\vspace{-0.35cm}
\end{figure*}
\begin{figure*}[!t]
\begin{minipage}[!t]{.990\linewidth}
  \centering
\leftline{\;\;\;\;\;\;\;\;\;\;\;\;\;\;\;\;\;\;\;\;\;\;\;\;\;\;\;\;\;\;\;\;\;\;\;\;\;\;\;\;\;\;\;\;\;\;\;\;\;\;\;\;\;\;\;\;\;\;\;\;\;\;\small{Blur}\;\;\;\;\;\;\;\;\;\;\;\;\small{FFTFormer-16}\;\;\;\;\;\;\;\;\;\small{NAFNet-32}\;\;\;\;\;\;\;\;\;\small{SegDeblur-S}\;\;\;\;\;\;\;\;\;\;\;\;\small{Sharp}}\medskip
\end{minipage}
\vspace{-0.3cm}
\caption{Visual comparison results on RSBlur~\cite{rsblur}. We compare our SegDeblur-S ($14.44$ GMACs) with FFTFormer-16~\cite{fftformer} ($16.41$ GMACs) and NAFNet-32~\cite{nafnet} ($16.25$ GMACs). Note that all methods are trained with RSBlur.}
\vspace{-0.3cm}
\label{fig:vis_rsblur}
\end{figure*}
\subsection{Logarithmic fourier discretization model} \label{sec:kest}
To successfully train our classification model as shown in Fig.~\ref{fig:proposed_method}, we present two key strategies to reduce the nature of ill-posedness: (i) we utilize the logarithmic fourier space~\cite{cepstrum} to simplify the relationship between blur and sharp images so that the basis kernels are easily estimated, and (ii) we introduce a latent sharp image in order to restrict the number of feasible kernel solutions, thereby making it simpler to train our classification model. Those key techniques facilitate implementing our classification model at a low computational cost.
Hence, we first consider a fourier operator $\mathcal{F}$. It has an inherent property that converts from deconvolution to division, i.e., $y \odot k \leftrightarrows \mathcal{F}(y) / \mathcal{F}(k)$~\cite{fft}.
To further simplify the relationship between two fourier samples, we apply the logarithmic operation to the fourier space (i.e., logarithmic fourier transform $\mathcal{F_L}$ of sample images), and then we obtain the relationship $y \odot k \leftrightarrows \mathcal{F_L}(y) - \mathcal{F_L}(k)$~\cite{cepstrum}.

Let $X=\mathcal{F_L}(x)$ and $Y=\mathcal{F_L}(y)$ be sharp and blur samples by the logarithmic fourier operator $\mathcal{F_L}$.
We are interested in estimating a set of logarithmic fourier basis kernels, $\tilde{K}=\{\tilde{K}^{(r)}\}^R_{r=1}$ where $r$ is a basis kernel index, and blur segmentation map $\rho$ modeled by classification model $g$, i.e., 
\begin{equation} \label{eq:cluster_model}
g:y\rightarrow (\tilde{K},\rho).
\end{equation}
Our classification model $g_{\psi,\phi}$ consists of two sub-modules for a blur pixel discretizer $h_\phi$ and kernel estimator $k_\psi$ as shown in Fig.~\ref{fig:proposed_method}.
To restrict the number of feasible kernel solutions, i.e., reduce the nature of ill-posedness, the blur pixel discretizer additionally produces the latent sharp image $\tilde{x}_\ell$ that is $h_\phi:y\rightarrow (\rho, \tilde{x}_\ell)$.
Then, the kernel estimator $k_\psi$ takes two inputs, e.g., blur and latent sharp samples ($Y,\mathcal{F_L}(\tilde{x}_\ell)$) in the logarithmic fourier space and predicts a set of logarithmic fourier basis kernels $\tilde{K}$.
Given the basis kernels $\tilde{K}$, a set of deconvolved images, i.e., deconvolved class images $\nu=\{\nu^{(1)},\nu^{(2)},\ldots,\nu^{(R)}\}$, is generated by using simple subtraction (e.g., deconvolution) and inverse logarithmic fourier transform $\mathcal{F_L}^{-1}$, i.e.,
\begin{equation} \label{eq:cluster}
\nu=\mathcal{F_L}^{-1}(Y-\tilde{K}).
\end{equation}
Finally, we can extract the deconvolved pixels $\tilde{x}_d=\{\nu^{(\rho_{i})}_i\}^N_{i=1}$ by choosing individual pixels within the deconvolved class images $\nu$ based on a per-pixel class index $\rho_i$.
Namely, the deconvolved $i$th pixel of $\tilde{x}_d$ is determined by $i$th pixel of the chosen deconvolved class image $\nu^{(\rho_i)}$.
Then, the model parameters $\psi$ and $\phi$ are jointly optimized by minimizing the following loss:
\begin{equation} \label{eq:kernel_loss}
L_{\tt class}(\psi,\phi;\mathcal{D})=\frac{1}{|\mathcal{D}|}\sum_{(x,y)\in \mathcal{D}} d(\tilde{x}_d,x)+\lambda d(\tilde{x}_\ell,x),
\end{equation}
where $d$ is some distance or divergence in the image domain, e.g., the PSNR loss~\cite{nafnet} and $\lambda$ is a hyperparameter to control the contribution of a latent sharp image.
In our method, the deblurring regression problem is tackled by reframing it as a blur classification task, which allows us to build our model $g_{\psi,\phi}$ at a low computational cost.

\begin{figure*}[htb]
  \centering
\begin{minipage}[!t]{.260\linewidth}
  \centering
  \centerline{\includegraphics[width=4.60cm]{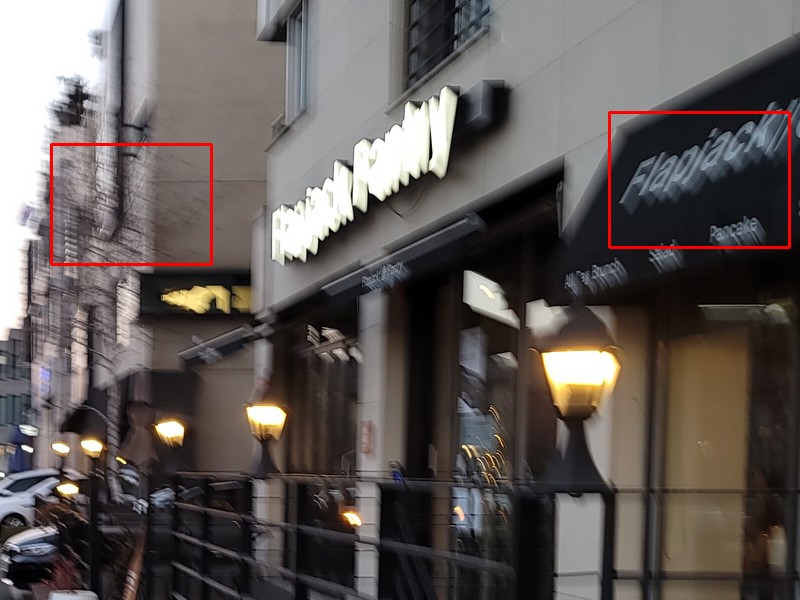}}
\end{minipage}
\begin{minipage}[!t]{.130\linewidth}
  \centering
  \centerline{\includegraphics[width=2.30cm]{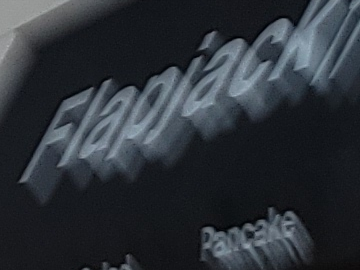}}
  \centerline{\includegraphics[width=2.30cm]{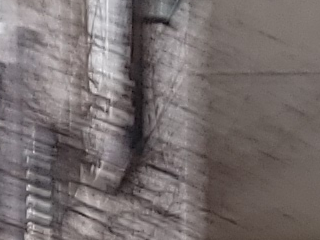}}
\end{minipage}
\begin{minipage}[!t]{.130\linewidth}
  \centering
  \centerline{\includegraphics[width=2.30cm]{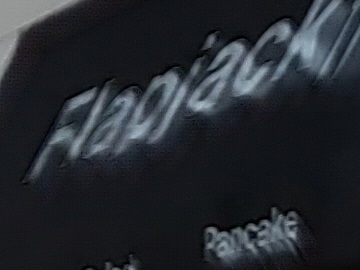}}
  \centerline{\includegraphics[width=2.30cm]{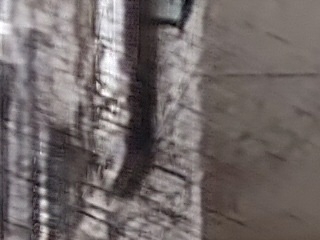}}
\end{minipage}
\begin{minipage}[!t]{.130\linewidth}
  \centering
  \centerline{\includegraphics[width=2.30cm]{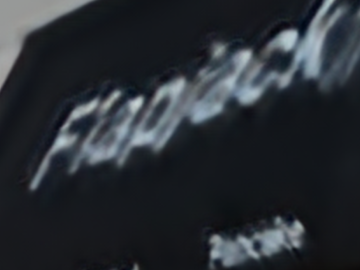}}
  \centerline{\includegraphics[width=2.30cm]{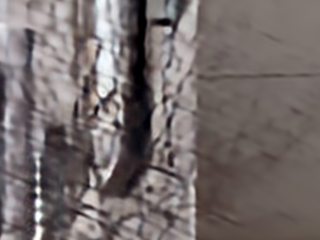}}
\end{minipage}
\begin{minipage}[!t]{.130\linewidth}
  \centering
  \centerline{\includegraphics[width=2.30cm]{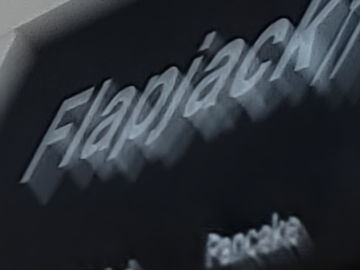}}
  \centerline{\includegraphics[width=2.30cm]{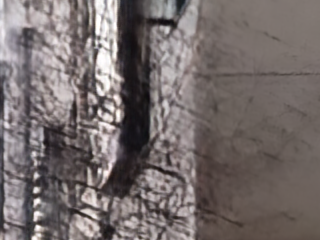}}
\end{minipage}
\begin{minipage}[!t]{.130\linewidth}
  \centering
  \centerline{\includegraphics[width=2.30cm]{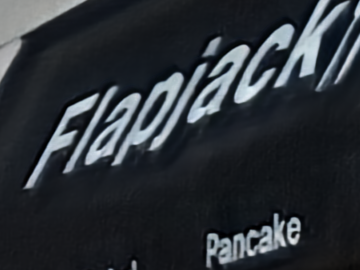}}
  \centerline{\includegraphics[width=2.30cm]{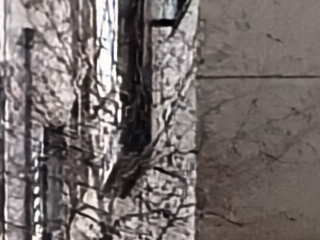}}
\end{minipage}
\vspace{-0.35cm}
\end{figure*}
\begin{figure*}[htb]
  \centering
\begin{minipage}[!t]{.260\linewidth}
  \centering
  \centerline{\includegraphics[width=4.60cm]{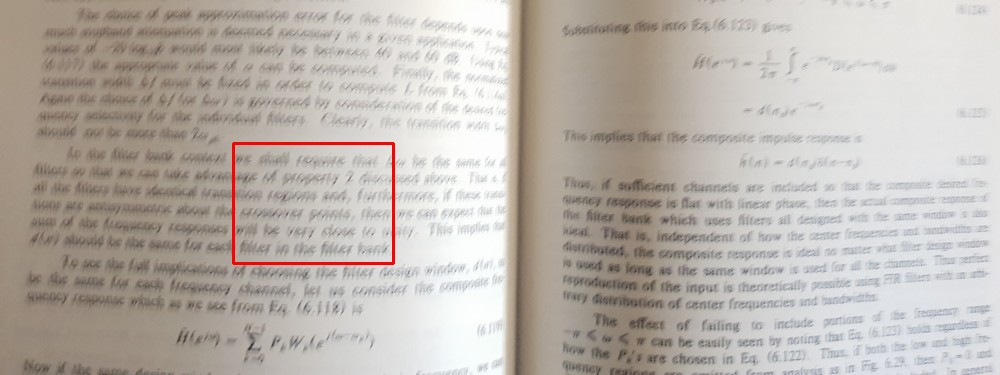}}
\end{minipage}
\begin{minipage}[!t]{.130\linewidth}
  \centering
  \centerline{\includegraphics[width=2.30cm]{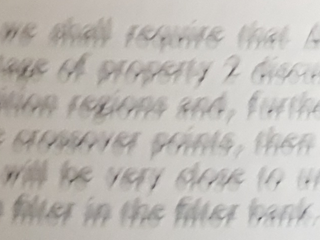}}
\end{minipage}
\begin{minipage}[!t]{.130\linewidth}
  \centering
  \centerline{\includegraphics[width=2.30cm]{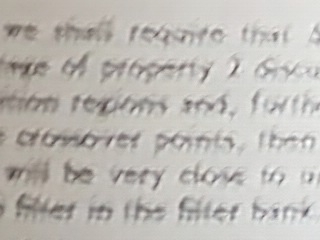}}
\end{minipage}
\begin{minipage}[!t]{.130\linewidth}
  \centering
  \centerline{\includegraphics[width=2.30cm]{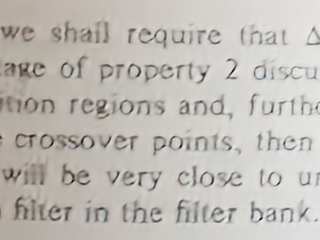}}
\end{minipage}
\begin{minipage}[!t]{.130\linewidth}
  \centering
  \centerline{\includegraphics[width=2.30cm]{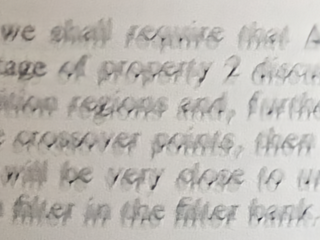}}
\end{minipage}
\begin{minipage}[!t]{.130\linewidth}
  \centering
  \centerline{\includegraphics[width=2.30cm]{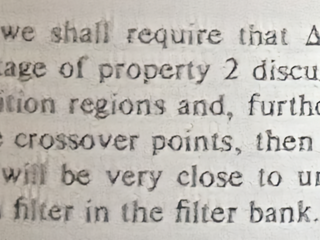}}
\end{minipage}
\vspace{-0.35cm}
\end{figure*}
\begin{figure*}[!htb]
  \centering
\begin{minipage}[!t]{.260\linewidth}
  \centering
  \centerline{\includegraphics[width=4.60cm]{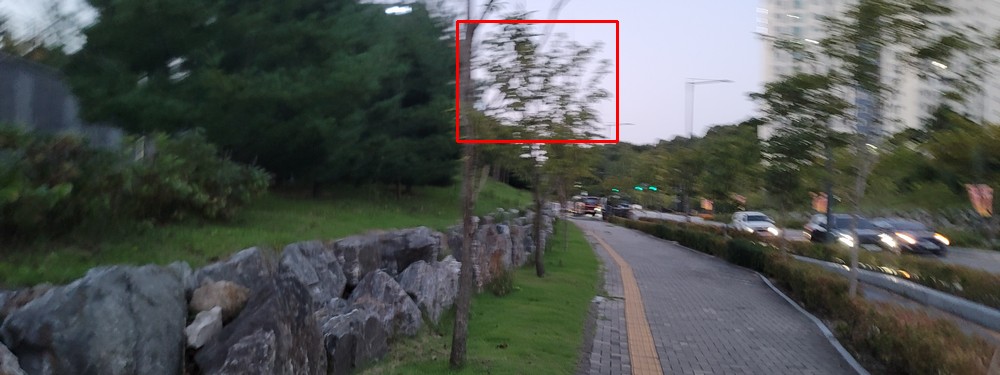}}
\end{minipage}
\begin{minipage}[!t]{.130\linewidth}
  \centering
  \centerline{\includegraphics[width=2.30cm]{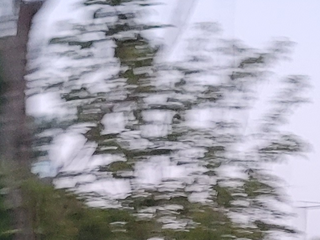}}
\end{minipage}
\begin{minipage}[!t]{.130\linewidth}
  \centering
  \centerline{\includegraphics[width=2.30cm]{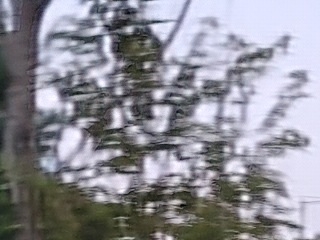}}
\end{minipage}
\begin{minipage}[!t]{.130\linewidth}
  \centering
  \centerline{\includegraphics[width=2.30cm]{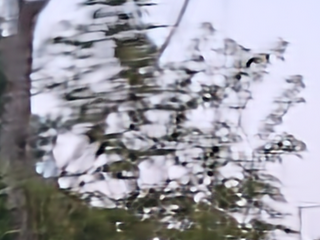}}
\end{minipage}
\begin{minipage}[!t]{.130\linewidth}
  \centering
  \centerline{\includegraphics[width=2.30cm]{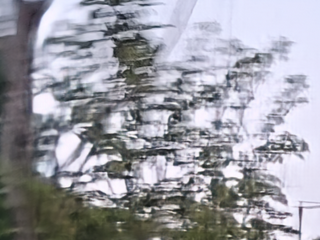}}
\end{minipage}
\begin{minipage}[!t]{.130\linewidth}
  \centering
  \centerline{\includegraphics[width=2.30cm]{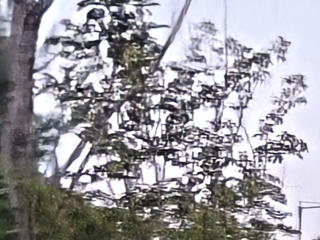}}
\end{minipage}
\vspace{-0.35cm}
\end{figure*}
\begin{figure*}[!t]
\begin{minipage}[!t]{.990\linewidth}
\centering\leftline{\;\;\;\;\;\;\;\;\;\;\;\;\;\;\;\;\;\;\;\;\;\;\;\;\;\;\;\;\;\;\;\;\;\;\;\;\;\;\;\;\;\;\;\;\;\;\;\;\;\;\;\;\;\;\;\;\;\;\;\;\;\;\small{Blur}\;\;\;\;\;\;\;\;\;\small{Samsung EnhanceX}\;\;\;\small{Google Unblur}\;\;\;\;\;\small{NAFNet-32+}\;\;\;\;\;\;\;\small{SegDeblur-S+}}\medskip
\end{minipage}
\vspace{-0.3cm}
\caption{Visual comparison results on real-world blur images. NAFNet-32+ and SegDeblur-S+ are the accelerated (faster) models for the deployment and they are evaluated for fair comparison with the commercial applications. This will be more discussed in Section~\ref{sec:experiment_commercial}.}
\vspace{-0.3cm}
\label{fig:vis_realworld}
\end{figure*}

\subsection{Discrete-to-continuous conversion model} \label{sec:deblurring}
As discussed in Section.~\ref{sec:kest}, we demonstrate that our classification model produces a discretized image residual error (i.e., discrete version of GT) under a small computational cost ($4$ GMACs).
Therefore, it may allow for less computational cost of the subsequent task, i.e., D2C converter, to transform such discretized image residual error $\rho$ in the discrete domain into the image residual error $e$ in the continuous domain, i.e., $f_\theta:\mathcal{Z}\rightarrow \mathcal{X}$, rather than directly regressing the image residual error without such GT-like information.
As a result, we can achieve an efficient deblurring model ($14$ GMACs) whose performance is still comparable to that of state-of-the-art methods as shown in Table~\ref{tbl:realblur}.

Given the discretized image residual error (i.e., blur segmentation map $\rho$) and blur image $y$, the final deblurred image $\hat{x}$ is estimated via our D2C converter ${f_\theta}$ as follows:
\begin{equation} \label{eq:res2}
\hat{x}=y+f_{\theta}(\rho,y),
\end{equation}
where $f_{\theta}(\rho,y)$ is the estimated image residual error $\hat{e}$ and $\hat{x}$ denotes the final deblurred image.
Then, we derive our deblur loss that minimizes the distance between the real sharp and deblurred images by
\begin{equation} \label{eq:deblur_loss}
L_{\tt deblur}(\theta;\mathcal{D})=\frac{1}{|\mathcal{D}|}\sum_{(x,y)\in \mathcal{D}}d(y+f_{\theta}(\rho,y),x),
\end{equation}
where $d$ is some distance or divergence in the image domain, e.g., the PSNR loss~\cite{nafnet}.
Note that the classification model $g_{\psi,\phi}$ is not updated by~\eqref{eq:deblur_loss}.
Since our classification model produces a blur segmentation map based on a deconvolution procedure, it may be more grounded in physical blur modeling than the simple blur-to-sharp mapping that the kernel-free methods do. Hence, Under an efficient model, the proposed method equipped with our blur segmentation map reconstructs more natural deblurred results as shown in Fig.~\ref{fig:distortion} (c) than the kernel-free method without any physical blur information as shown in Fig.~\ref{fig:distortion} (b).

\subsection{Implementation details} \label{sec:implementation}
In this section, we present the implementation details of our method. We use NAFNet~\cite{nafnet} as our blur pixel discretizer and D2C converter while U-Net~\cite{unet} is chosen as our kernel estimator.
We adopt a two-stage training strategy for our method.
In the first stage, we jointly train our blur pixel discretizer $h_{\phi}$ and kernel estimator $k_{\psi}$, with blur images $y$ by minimizing \eqref{eq:kernel_loss}, as shown in Fig.~\ref{fig:proposed_method}. The element of the blur segmentation map, i.e., $\rho_i$, is structured as a one-hot vector. We use \textit{argmax} operation to obtain the per-pixel class index. Then, with this class information, we conduct pixel-wise sampling from the deconvolved class images to recover the deconvolved image $\tilde{x}_d$.
In the second stage, we optimize our D2C converter $f_{\theta}$ with the blur segmentation map $\rho$ and blur images $y$ by using \eqref{eq:deblur_loss}, as shown in Fig.~\ref{fig:proposed_method}.
Note that the blur segmentation map estimated by the blur pixel discretizer is represented as probabilities, for example, [$0.5$,$0.2$,$0.2$,$0.1$] for each pixel.
In the second and test stages, it is reconstructed by setting the maximum value to $1$ and all other values to $0$ to provide clearer blur class information to the following D2C converter.
Furthermore, our blur pixel discretizer $h_{\phi}$ is frozen, and our kernel estimator $k_{\psi}$ and the following operations such as logarithmic fourier transform, deconvolution, and sampling are discarded in the second and test stages.
\vspace{-0.15cm}
\section{Experiments} \label{sec:experiments}
\subsection{Experimental setup} \label{sec:experimental_setup}
\vspace{0.02in}
\setlength{\parindent}{0in}\textbf{Dataset and evaluation metrics.}
We use RealBlur~\cite{realblur}, RSBlur~\cite{rsblur}, ReLoBlur~\cite{reloblur} and GoPro~\cite{gopro} datasets for training and evaluation. RealBlur, RSBlur and ReLoBlur are realistic motion blur datasets that capture blur and sharp 
images in the same scene by using the beam splitter, whereas GoPro is a synthetic dataset where the blur image is generated by averaging sharp video frames captured by a high-speed camera.
RealBlur consists of RealBlur-J (sRGB domain) and RealBlur-R (RAW domain). Each RealBlur type comprises 3,758 and 980 image pairs for training and test sets, respectively. RSBlur contains 8,878 and 3,360 blur-sharp image pairs for training and test sets, respectively.
ReLoBlur consists of 2,010 and 395 image pairs for training and test sets, respectively.
GoPro contains 2,103 and 1,111 blur-sharp image pairs for training and test sets, respectively.
To verify the performance of the proposed method, we measure Peak Signal to Noise Ratio (PSNR) and Structural SIMilarity (SSIM)~\cite{ssim}.
To measure model efficiency, we compute the number of network parameters and Multiply–ACcumulate operations (MACs) based on the image size of $256 \times 256$.
Also, we compute on-chip execution time based on the image size of $2000$ $\times$ $2000$ when compared with commercial applications.

\begin{table}[!t]\centering
\scalebox{0.85}{
\begin{tabular}{c|c|c|c}
\hline \multirow{2}{*}{Methods}	& \multirow{2}{*}{Params} & {GMACs}  & {RealBlur-J} \\
\cline{3-4}                     & & {Prior\;Deblur\;Total}                         & {Total\;\;Large}\\
\hline {NAFNet-32~\cite{nafnet}}	            &{$16.00$}& {\;\;-\;\;\;\;\;$16.25$\;\;$16.25$} & {$31.99\;\;28.27$}\\
{NAFNet-64~\cite{nafnet}}	            &{$63.50$}& {\;\;-\;\;\;\;\;$63.64$\;\;$63.64$} & {$32.50\;\;29.06$}\\
{SegDeblur-S (ours)}	&{$12.30$}& {$4.37$\;\;$10.07$\;\;$14.44$} & {$32.53\;\;29.12$}\\
{SegDeblur-L (ours)}	&{$55.40$}& {$4.37$\;\;$58.31$\;\;$62.68$} & {$\textbf{32.95}\;\;\textbf{29.77}$}\\
\hline		
\end{tabular}
}
\vspace{0.02in}
\vspace{-0.3cm}
\caption{The effect of blur segmentation map compared against NAFNet~\cite{nafnet}. We present prior (blur pixel discretizer), deblur (D2C converter or NAFNet itself), and total model GMACs, respectively. We evaluate the methods on the whole and large motion test set of RealBlur-J~\cite{realblur}, which denotes as ``Total'' and ``Large''.}
\label{tbl:efficient}
\vspace{-0.3cm}
\end{table}
\begin{table}
    \centering
\scalebox{0.85}{
    \begin{tabular}{c|c|c|c}
\hline         &  Prior&  GMACs&\;PSNR $\uparrow$\;\;SSIM $\uparrow$\\
\hline         HiNet~\cite{hinet}&  &  $170.71$& $32.12$\;\;\;\;\;$0.921$\\
         MSDI-Net~\cite{msdi}&  \checkmark&  $336.43$& $32.35$\;\;\;\;\;$0.923$\\
\hline         NAFNet*~\cite{nafnet}&  &  $63.64$& $33.12$\;\;\;\;\;$0.930$\\
         UFPNet*~\cite{ufp}&  \checkmark&  $243.33$& $33.35$\;\;\;\;\;$0.934$\\
         SegDeblur-L* (ours)&  \checkmark&  $62.68$& $\textbf{33.51}$\;\;\;\;\;$\textbf{0.938}$\\
\hline
    \end{tabular}}
    \vspace{0.02in}
    \vspace{-0.3cm}
    \caption{Comparison with prior-based methods~\cite{ufp,msdi}. ``*'' denotes that the methods use the test-time local converter (TLC)~\cite{local}. Although our prior knowledge (blur segmentation map) is generated from a small-scale model, it significantly improves performance compared to the others.}
    \label{tbl:prior_result}
    \vspace{-0.3cm}
\end{table}
\vspace{0.02in}
\setlength{\parindent}{0in}\textbf{Network architecture variants.}
NAFNet~\cite{nafnet} consists of encoder blocks $\{1,1,1,28\}$, middle block $\{1\}$ and decoder blocks $\{1,1,1,28\}$. NAFNet with 32 and 64 widths are denoted as NAFNet-32 and NAFNet-64.
Similarly, we build SegDeblur-S ($14.44$ GMACs) and SegDeblur-L ($62.68$ GMACs) for realistic datasets such as RealBlur~\cite{realblur}, RSBlur~\cite{rsblur} and ReLoBlur~\cite{reloblur}. Since FFTFormer~\cite{fftformer} is the best performance on GoPro~\cite{gopro}, we build SegFFTFormer ($135.81$ GMACs) for GoPro.
Our network variants are described in detail in Section~\ref{sec:app_variants} of Appendix.

\vspace{0.02in}
\setlength{\parindent}{0in}\textbf{Implementation details.} 
We train with RealBlur~\cite{realblur}, RSBlur~\cite{rsblur}, ReLoBlur~\cite{reloblur} and GoPro~\cite{gopro} randomly cropped by $256 \times 256$.
We train our SegDeblur-S and L up to $1,000$ and $2,000$ epochs while our SegFFTFormer is trained up to $60,000$ iterations as in~\cite{fftformer}. The batch size is $16$ with $1$ GPU for SegDeblur-S, $32$ with $4$ GPUs for SegDeblur-L, and $16$ with $8$ GPUs for SegFFTFormer.
Our blur pixel discretizer, kernel estimator and D2C converter are optimized by the AdamW~\cite{adamw} algorithm ($\beta_1 = 0.9$, $\beta_2 = 0.9$ and weight decay $1e^{-3}$) with the cosine annealing schedule ($1e^{-3}$ to $1e^{-7}$)~\cite{cosineanneal} gradually reduced for total iterations of each dataset.
We employ the hyperparameter as $\lambda=1.0$. We use the number of classes as $R=16$ for RealBlur and $R=8$ for the other datasets.
We use RealBlur-J for all ablation studies.

\begin{table}[!t]\centering
\scalebox{0.82}{
\begin{tabular}{c|c|c|c}
\hline \multirow{2}{*}{Methods}	& \multirow{2}{*}{GMACs}	& {RealBlur-J} & {RealBlur-R}\\
\cline{3-4}                               &                       	& {PSNR$\uparrow$\;\;SSIM$\uparrow$}  & {PSNR$\uparrow$\;\;SSIM$\uparrow$}\\
\hline {MPRNet~\cite{mprnet}}	    & {$777.01$}	& {${31.76}$\;\;$0.922$} & {${39.31}$\;\;$0.972$}\\
{MIMO-UNet+~\cite{mimounet}}	    & {$154.41$}	& {${31.92}$\;\;${0.919}$} & {\;\;\;\;-\;\;\;\;\;\;\;\;\;-\;\;\;\;\;}\\
{FMIMO-UNet~\cite{deeprft}}	        & {$80.21$}	& {${32.65}$\;\;${0.931}$} & {${40.01}$\;\;$0.972$} \\
{Stripformer~\cite{stripformer}}	    & {$169.89$}	& {${32.48}$\;\;${0.929}$} & {${39.84}$\;\;${0.974}$}\\
{MAXIM-3S~\cite{maxim}}	            & {$169.50$}	& {${32.84}$\;\;${0.935}$} & {${39.45}$\;\;$0.962$} \\
{FFTFormer~\cite{grl}}	        & {$131.45$}	& {${32.62}$\;\;${0.932}$} & {${40.20}$\;\;$0.973$} \\
{GRL-B~\cite{grl}}	        & {$1285.28$}	& {${32.82}$\;\;${0.932}$} & {${40.20}$\;\;$0.974$} \\
{NAFNet-64~\cite{nafnet}}	        & {$63.64$}	& {${32.50}$\;\;${0.928}$} & {${39.89}$\;\;$0.973$} \\
{SegDeblur-S (ours)}	& {$14.44$}	& {${32.53}$\;\;${0.927}$} & {${39.75}$\;\;$0.973$}\\
{SegDeblur-L (ours)}	& {$62.68$}	& {${32.95}$\;\;$0.934$} & {${40.21}$\;\;${0.975}$}\\
{SegDeblur-L* (ours)}	& {$62.68$}	& {$\textbf{33.51}$\;\;$\textbf{0.938}$} & {$\textbf{40.79}$\;\;$\textbf{0.976}$}\\
\hline		
\end{tabular}
}
\vspace{0.02in}
\vspace{-0.3cm}
\caption{Comparison results on RealBlur-J~\cite{realblur} and RealBlur-R~\cite{realblur} datasets. ``*'' denotes that the method uses the test-time local converter (TLC)~\cite{local}. The best results are indicated in bold.}
\label{tbl:realblur}
\vspace{-0.3cm}
\end{table}
\subsection{Blind motion deblurring for practical usage} \label{sec:experiment_efficient}
\vspace{0.02in}
\setlength{\parindent}{0in}\textbf{Model efficiency and large motion scenarios.}
As shown in Table~\ref{tbl:efficient}, simply scaling down model size significantly drops the performance on large motion scenarios. We investigate that our efficient model is robust to large motion scenarios compared to the kernel-free method at the same cost. We manually extract the largest motion $104$ image pairs from RealBlur-J~\cite{realblur} to construct the large motion blur test set. We measure PSNR and SSIM for model performance while MACs is measured for model efficiency. As shown in Table~\ref{tbl:efficient}, our efficient model, i.e., SegDeblur-S, improves PSNR in the large motion set ($28.27\rightarrow29.12$ dB) and total set of RealBlur-J ($31.99\rightarrow32.53$ dB), whose performance is also comparable to that of the larger model, NAFNet-64. This explains the necessity of our blur segmentation map particularly when training an efficient model.

\vspace{0.02in}
\setlength{\parindent}{0in}\textbf{Comparison to other prior-based methods.}
Since our method is viewed as a prior-based deblurring method, we compare it with other prior-based deblurring methods such as UFPNet~\cite{ufp} and MSDI-Net~\cite{msdi}.
UFPNet is based on NAFNet~\cite{nafnet} and produces the non-uniform blur kernel as prior information. MSDI-Net is built upon HiNet~\cite{hinet} and provides the degradation representation as prior information.
Although such prior information is beneficial, the computational costs to estimate them and fuse them are highly expensive as shown in Table~\ref{tbl:prior_result}, which is insufficient for practical usage. In contrast, our method only requires $4$ GMACs to generate a blur segmentation map for prior information. As a result, our SegDeblur-L ($62$ GMACs) achieves $33.51$ dB which is even better than $33.35$ dB in UFPNet ($243$ GMACs). Despite the most cost-effective method for generating prior information, our blur segmentation map has the greatest impact on improving deblurring performance compared to the other0 methods.

\begin{table}[!t]\centering
\scalebox{0.90}{
\begin{tabular}{c|c|c|c}
\hline \multirow{2}{*}{Methods}	& \multirow{2}{*}{GMACs}  & \multicolumn{2}{c}{RSBlur} \\
\cline{3-4}                     &                          & {PSNR $\uparrow$} & {SSIM $\uparrow$}\\
\hline {SRN-Deblur~\cite{srndeblurnet}}	   & {$1434.82$} & {$32.53$} & {$0.840$}\\
{MIMO-UNet+~\cite{mimounet}}	           & {$154.41$} & {$33.37$} & {$0.856$}\\
{MPRNet~\cite{mprnet}}	                   & {$777.01$} & {$33.61$} & {$0.861$}\\
{Restormer~\cite{restormer}}	           & {$141.00$} & {$33.69$} & {$0.863$}\\
{Uformer-B~\cite{uformer}}	           & {$89.50$} & {$33.98$} & {$0.866$}\\
{NAFNet-64~\cite{nafnet}}	           & {$63.64$} & {$33.97$} & {$0.866$}\\
{SegDeblur-S (ours)}	            & {$14.44$} & {$33.96$} & {$0.865$}\\
{SegDeblur-L (ours)}	            & {$62.68$} & {${34.21}$} & {${0.870}$}\\
{SegDeblur-L* (ours)}	            & {$62.68$} & {$\textbf{34.63}$} & {$\textbf{0.876}$}\\
\hline		
\end{tabular}
}
\vspace{0.02in}
\vspace{-0.3cm}
\caption{Comparison results on RSBlur~\cite{rsblur} dataset. ``*'' denotes that the method uses the test-time local converter (TLC)~\cite{local}. The best results are indicated in bold.}
\label{tbl:rsblur}
\vspace{-0.3cm}
\end{table}
\vspace{0.15in}
\vspace{-0.3cm}

\begin{table}[!t]\centering
\scalebox{0.90}{
\begin{tabular}{c|c|c|c}
\hline \multirow{2}{*}{Methods}	& \multirow{2}{*}{GMACs}  & \multicolumn{2}{c}{ReLoBlur} \\
\cline{3-4}                     &                          & {PSNR $\uparrow$} & {SSIM $\uparrow$}\\
\hline {SRN-Deblur~\cite{srndeblurnet}}	   & {$1434.82$} & {$34.30$} & {$0.923$}\\
{DeblurGAN-v2~\cite{deblurganv2}}	           & {$411.34$} & {$33.85$} & {$0.902$}\\
{HiNet~\cite{hinet}}	                   & {$170.71$} & {$34.36$} & {$0.915$}\\
{MIMO-UNet~\cite{mimounet}}	           & {$154.41$} & {$34.52$} & {${0.925}$}\\
{LBAG+~\cite{reloblur}}	           & {$154.44$} & {$34.85$} & {${0.925}$}\\
{NAFNet-64~\cite{nafnet}}	           & {$63.64$} & {$34.55$} & {$0.926$}\\
{SegDeblur-S (ours)}	            & {$14.44$} & {${34.86}$} & {${0.923}$}\\
{SegDeblur-L (ours)}	            & {$62.68$} & {$\textbf{35.34}$} & {$\textbf{0.927}$}\\
\hline		
\end{tabular}
}
\vspace{0.02in}
\vspace{-0.3cm}
\caption{Comparison results on ReLoBlur~\cite{reloblur} dataset. The best results are indicated in bold.}
\label{tbl:reloblur}
\vspace{-0.4cm}
\end{table}

\subsection{Comparison to kernel-free methods} \label{sec:experiment_kernelfree}
\vspace{0.05in}
\setlength{\parindent}{0in}\textbf{Results on RealBlur.}
As discussed in Section~\ref{sec:experiment_efficient}, our blur segmentation map aids in improving deblurring performance under an efficient model. In this section, we confirm whether our blur segmentation map is still beneficial in larger networks. To this end, we train and evaluate with RealBlur-J and -R dataset~\cite{realblur}. We measure PSNR and SSIM for model performance while MACs is measured for model efficiency.
The experimental results of RealBlur-J and -R are summarized in Table~\ref{tbl:realblur}.
The results show that our SegDeblur-L improves performance from $32.50$ to $32.95$ dB (RealBlur-J) and $39.89$ to $40.21$ (RealBlur-R).

\setlength{\parindent}{0in}\textbf{Results on RSBlur.}
RSBlur~\cite{rsblur} enables us to conduct a comprehensive evaluation of deblurring models since it contains realistic high-resolution images around $1920 \times 1200$, large motion scenarios, and various blur types such as object and camera motions. To this end, we train and evaluate with RSBlur.
As shown in Table~\ref{tbl:rsblur}, our SegDeblur-L achieves the best performance against other methods. Furthermore, it is noticeable that the performance of our efficient model, i.e., SegDeblur-S ($14$ GMACs), is comparable to that of the best-performing methods such as NAFNet-64 ($64$ GMACs) and Uformer-B ($89$ GMACs). As illustrated in Fig.~\ref{fig:vis_rsblur}, our efficient model shows better qualitative results compared to the other efficient deblurring models~\cite{fftformer,nafnet}.

\vspace{0.02in}
\setlength{\parindent}{0in}\textbf{Results on ReLoBlur.}
We experiment with ReLoBlur~\cite{reloblur}, which consists of various object motion blur images. To verify that our blur segmentation map is also helpful for object motion blur dataset, we train and evaluate with ReLoBlur.
As shown in Table~\ref{tbl:reloblur}, our SegDeblur-S ($14$ GMACs) achieves the performance comparable to that of the best-performing method, LBAG+ ($154.44$ GMACs) while requiring only $10$\% of the computational cost.

\begin{table}[!t]\centering
\scalebox{0.90}{
\begin{tabular}{c|c|c|c}
\hline \multirow{2}{*}{Methods}	& \multirow{2}{*}{GMACs}   & \multicolumn{2}{c}{GoPro} \\
\cline{3-4}                     &                           & {PSNR $\uparrow$} & {SSIM $\uparrow$} \\
\hline {SRN-DeblurNet~\cite{srndeblurnet}}& {${1434.82}$} & {$30.26$} & {$0.932$}  \\
{DeblurGAN-v2~\cite{deblurganv2}}	        & {${411.34}$} & {$29.55$} & {$0.934$}  \\
{MPRNet~\cite{mprnet}}	        & {${777.01}$} & {$32.66$} & {$0.959$} \\
{MIMO-UNet+~\cite{mimounet}}	    & {${154.41}$} & {$32.45$} & {$0.957$} \\
{HINet~\cite{hinet}}	            & {${170.71}$} & {$32.77$} & {$0.959$}\\
{Restormer~\cite{restormer}}	    & {${141.00}$} & {$32.92$} & {$0.961$}\\
{Uformer-B~\cite{uformer}}	    & {${89.50}$} & {$32.97$} & {$0.967$}\\
{Stripformer~\cite{stripformer}}	    & {${169.89}$} & {$33.08$} & {$0.962$}\\
{MAXIM-3S~\cite{maxim}}	            & {${169.50}$} & {$32.86$} & {$0.961$}\\
{NAFNet-64~\cite{nafnet}}	        & {${63.64}$} & {$33.69$} & {$0.966$}\\
{FNAFNet-64~\cite{deeprft}}	        & {${72.40}$} & {$33.85$} & {$0.967$}\\
{MSDI-Net\cite{msdi}}	        & {${336.43}$} & {$33.28$} & {$0.964$}\\
{UFPNet\cite{ufp}}	        & {${243.33}$} & {$34.06$} & {$0.968$}\\
{GRL-B\cite{grl}}	        & {${1285.28}$} & {$33.93$} & {$0.968$}\\
{FFTformer\cite{fftformer}}	        & {${131.45}$} & {$34.14$} & {$0.968$}\\
{SegFFTformer (ours)}	& {${135.81}$} & {$\textbf{34.38}$} & {$\textbf{0.970}$}  \\
\hline		
\end{tabular}
}
\vspace{0.02in}
\vspace{-0.3cm}
\caption{Comparison results on GoPro~\cite{gopro} dataset. We train FFTformer~\cite{fftformer} with our blur segmentation map which is referred to as SegFFTformer. The best results are indicated in bold.}
\label{tbl:gopro}
\vspace{-0.5cm}
\end{table}
\vspace{0.02in}
\setlength{\parindent}{0in}\textbf{Results on GoPro.}
We do experiments on GoPro~\cite{gopro} which is a synthetic dataset.
To confirm that our method still works well on the transformer-based model, we instead choose the baseline, FFTFormer~\cite{fftformer} which also gives the best performance on this dataset.
We train FFTformer with our blur segmentation map and demonstrate that our method improves performance from $34.14$ to $34.38$ dB only with additional $4$ GMACs, as denoted in Table~\ref{tbl:gopro}. Note that the performance of FFTFormer in our experiment is $34.14$ dB, slightly below the reported performance of $34.21$ dB.

\subsection{Comparison to commercial applications} \label{sec:experiment_commercial}
The goal of this experiment is to confirm whether our method achieves model performance and efficiency comparable to those of commercial applications such as EnhanceX in Samsung Galaxy S23 and Unblur in Google Pixel 8. 
To accelerate our SegDeblur-S, we modify some network architectures of our method (called SegDeblur-S+) because some operations are not supported by AI acceleration, which will be more discussed in Section~\ref{sec:app_deploy} of Appendix.
We train SegDeblur-S+ with a combination of RealBlur~\cite{realblur} and RSBlur~\cite{rsblur}. 
We deploy our SegDeblur-S+ using 32-bit floating point precision on GPU without additional quantization.
As shown in Fig.~\ref{fig:vis_realworld}, our method visually produces compelling results on real-world blur images compared to those of commercial applications and the other accelerated model (NAFNet-32+).
Furthermore, the on-chip execution times of ours is $2.35$s while the execution times of Samsung EnhanceX ($1.64$s) and Google Unblur ($2.03$s) are measured. Note that there is greater potential for further accelerating ours if deployed on Neural Processing Unit (NPU).

\subsection{Ablation study} \label{sec:ablation}
\setlength{\parindent}{0in}\textbf{Contribution by each module.}
In this section, we explore which module of our method has more impact on performance improvement. As described in Table~\ref{tbl:ablation_module}, with no module, it indicates the original NAFNet-32~\cite{nafnet}. When adding (i), it uses blur segmentation map but not considers latent sharp image and logarithmic fourier space.
The method including (i) and (ii) mean SegDeblur-S but uses fourier space instead of logarithmic fourier space.
The final method containing all modules denotes our SegDeblur-S.
As shown in Table~\ref{tbl:ablation_module}, we observe that adding the module such as (ii) or (iii) provides a meaningful performance improvement from $31.96$ to $32.17$ dB and from $32.17$ to $32.53$ dB, respectively.
This explains the necessity of utilizing a latent sharp image and logarithmic fourier space.
Furthermore, simply estimating blur segmentation map is not helpful since it may suffer from the ill-posed problem as discussed in Section~\ref{sec:kest} (see the results of the first and second columns).

\begin{table}[!t]\centering
\scalebox{0.82}{
\begin{tabular}{c|c|c|c|c}
\hline{(i) Blur Segmentation Map}                             & {}	  & {\checkmark} & {\checkmark}      & {\checkmark}\\
{(ii) Latent Sharp Loss}                     & {}	  & {}          & {\checkmark}      & {\checkmark}\\
{(iii) Logarithmic Fourier Transform}		 & {}     & {}          & {}      & {\checkmark}\\
\hline{PSNR (dB)}  & {$31.99$}	& {$31.96$}	        & {$32.17$}           & {$\textbf{32.53}$}\\
\hline		
\end{tabular}
}
\vspace{0.02in}
\vspace{-0.3cm}
\caption{Ablation study on our components. Experimental results show that all components are necessary and positively combined to provide the best performance as indicated in bold.}
\label{tbl:ablation_module}
\vspace{-0.5cm}
\vspace{0.01in}
\end{table}

\begin{figure}[t]
  \centering
  \centerline{\includegraphics[width=8.6cm]{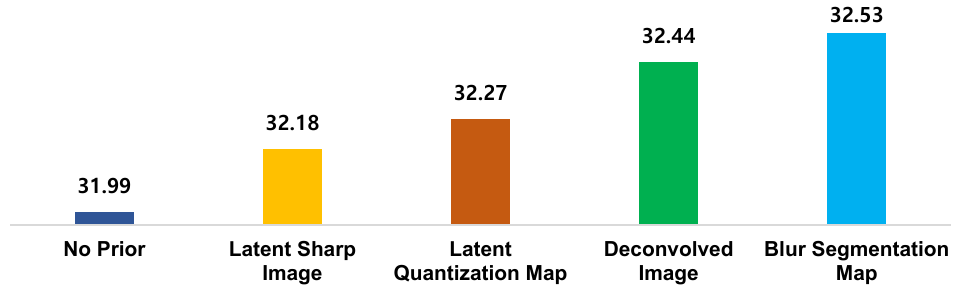}}
    \vspace{-0.3cm}
    \caption{Ablation study on several prior information. Our blur segmentation map gives the most contribution to improving deblurring performance.}
  \label{fig:ablation_prior}
\vspace{-0.1cm}
\end{figure}

\vspace{0.02in}
\setlength{\parindent}{0in}\textbf{Other prior information.}
Our classification model not only generates a blur segmentation map but also provides a latent sharp image and deconvolved image. As they are estimated by the sum of their image residual error and blur image, the latent quantization map (e.g., quantizing the image residual error in the latent sharp image) and our blur segmentation map can be seen as their discrete counterparts, i.e., latent sharp image - latent quantization map and deconvolved image - blur segmentation map.
In essence, the motion-frequency property is used for generating the deconvolved image while the latent sharp image does not consider it as shown in Fig.~\ref{fig:proposed_method}. 
As a result, the deconvolved image yields a better subsequent deblurring performance ($32.44$ dB) than the latent sharp image ($32.18$ dB) as shown in Fig.~\ref{fig:ablation_prior}. Similarly, the case of latent quantization map ($32.27$ dB) lags behind the performance compared with that of our blur segmentation map ($32.53$ dB).
This accounts for the necessity of the motion-frequency properties when generating GT-like priors. Furthermore, the discretized representations such as latent quantization map and blur segmentation map lead to a slight performance improvement compared with the continuous counterparts such as latent sharp and deconvolved images, e.g., $32.18 \rightarrow 32.27$ dB and $32.44 \rightarrow 32.53$ dB as shown in Fig.~\ref{fig:ablation_prior}.

\begin{figure}[t]
\begin{minipage}[!t]{.48\linewidth}
  \centering
  \centerline{\includegraphics[width=4.08cm]{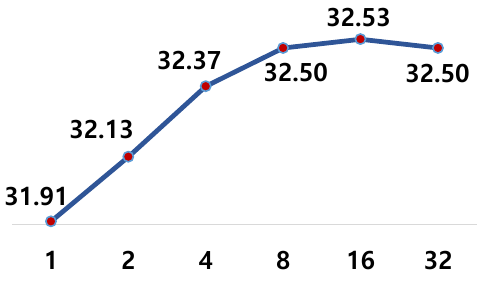}}
  \centerline{\small{(a)}}\medskip
\end{minipage}
\begin{minipage}[!t]{.48\linewidth}
  \centering
  \centerline{\includegraphics[width=3.55cm]{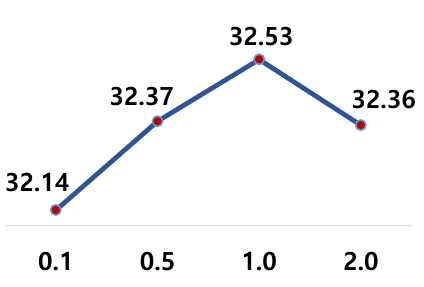}}
  \centerline{\small{(b)}}\medskip
\end{minipage}
\vspace{-0.3cm}
\caption{Ablation study on (a) \# of classes $R$ and (b) hyperparameter $\lambda$. The best results reside on $R=16$ and $\lambda=1.0$.}
\vspace{-0.5cm}
\label{fig:ablation_lambda}
\end{figure}

\vspace{0.02in}
\setlength{\parindent}{0in}\textbf{The number of classes.} We examine the effect of the number of classes $R$. We conduct experiments with $R = \{1, 2, 4, 8, 16, 32\}$.
As shown in Fig.~\ref{fig:ablation_lambda} (a), we observe that the performance of $R=1$ ($31.91$ dB) closely approximates that of deblurring without prior information (NAFNet-32), measured at $31.99$ dB. As the number of classes increases, it positively impacts the performance of the deblurring model.
On the other hand, a sufficient number of classes, e.g., more than $8$, produces a meaningful deblurring performance.

\setlength{\parindent}{0in}\textbf{Effects on $\lambda$.} We investigate which hyperparameter $\lambda$ generates the most optimal blur segmentation map. We experiment with $\lambda=\{0.1,0.5,1.0,2.0\}$.
Given the blur segmentation map varying $\lambda$ from $0.1$ to $2.0$, the final deblurring results are shown in Fig.~\ref{fig:ablation_lambda} (b). The result shows that the constraint of the latent sharp image is significant for reducing ill-posedness as discussed in Section~\ref{sec:kest} and $\lambda=1.0$ gives the best performance. Therefore, we choose the hyperparameter $\lambda$ as $1.0$ in our experiments.
For more insights on the proposed method, we conduct more ablation studies in Section~\ref{sec:app_ablation} of Appendix.
\vspace{-0.15cm}
\section{Conclusions} \label{sec:conclusion}
In this paper, we propose a new deblurring scheme that decomposes a deblurring regression problem into discretization and discrete-to-continuous conversion problems. We present a blur segmentation map that reflects the characteristics of the image residual error, which supports building an efficient model.
We demonstrate the competitiveness of our efficient model when compared with the larger deblurring model and commercial applications.
The proposed method can be extended to other deblurring tasks such as video deblurring~\cite{edvr} and defocus deblurring~\cite{defocus}.

\section*{Acknowledgement}
This work was supported by Institute of Information \& communications Technology Planning \& Evaluation (IITP) grant funded by the Korea government (MSIT) (No.2019-0-00075, Artificial Intelligence Graduate School Program (KAIST); No.2021-0-02068, Artificial Intelligence Innovation Hub).
{
    \small
    \bibliographystyle{ieeenat_fullname}
    \bibliography{main}

\begin{thebibliography}{41}
\providecommand{\natexlab}[1]{#1}
\providecommand{\url}[1]{\texttt{#1}}
\expandafter\ifx\csname urlstyle\endcsname\relax
  \providecommand{\doi}[1]{doi: #1}\else
  \providecommand{\doi}{doi: \begingroup \urlstyle{rm}\Url}\fi

\bibitem[Abuolaim and Brown(2020)]{defocus}
Abdullah Abuolaim and Michael~S Brown.
\newblock Defocus deblurring using dual-pixel data.
\newblock \emph{The European Conference on Computer Vision (ECCV)}, pages 111--126, 2020.

\bibitem[Brigham and Morrow(1967)]{fft}
E~Oran Brigham and RE Morrow.
\newblock The fast fourier transform.
\newblock \emph{IEEE spectrum}, 4\penalty0 (12):\penalty0 63--70, 1967.

\bibitem[Carbajal et~al.(2021)Carbajal, Vitoria, Delbracio, Mus{\'e}, and Lezama]{adaptivebasis}
Guillermo Carbajal, Patricia Vitoria, Mauricio Delbracio, Pablo Mus{\'e}, and Jos{\'e} Lezama.
\newblock Non-uniform blur kernel estimation via adaptive basis decomposition.
\newblock \emph{arXiv preprint arXiv:2102.01026}, 2021.

\bibitem[Chakrabarti(2016)]{bd1}
Ayan Chakrabarti.
\newblock A neural approach to blind motion deblurring.
\newblock \emph{The European Conference on Computer Vision (ECCV)}, pages 221--235, 2016.

\bibitem[Chen et~al.(2021)Chen, Lu, Zhang, Chu, and Chen]{hinet}
Liangyu Chen, Xin Lu, Jie Zhang, Xiaojie Chu, and Chengpeng Chen.
\newblock Hinet: Half instance normalization network for image restoration.
\newblock \emph{The IEEE Conference on Computer Vision and Pattern Recognition Workshop (CVPRW)}, pages 182--192, 2021.

\bibitem[Chen et~al.(2022)Chen, Chu, Zhang, and Sun]{nafnet}
Liangyu Chen, Xiaojie Chu, Xiangyu Zhang, and Jian Sun.
\newblock Simple baselines for image restoration.
\newblock \emph{The European Conference on Computer Vision (ECCV)}, 2022.

\bibitem[Childers et~al.(1977)Childers, Skinner, and Kemerait]{cepstrum}
Donald~G Childers, David~P Skinner, and Robert~C Kemerait.
\newblock The cepstrum: A guide to processing.
\newblock \emph{Proceedings of the IEEE}, 65\penalty0 (10):\penalty0 1428--1443, 1977.

\bibitem[Cho et~al.(2021)Cho, Ji, Hong, Jung, and Ko]{mimounet}
Sung-Jin Cho, Seo-Won Ji, Jun-Pyo Hong, Seung-Won Jung, and Sung-Jea Ko.
\newblock Rethinking coarse-to-fine approach in single image deblurring.
\newblock \emph{The IEEE International Conference on Computer Vision (ICCV)}, pages 4641--4650, 2021.

\bibitem[Chu et~al.(2022)Chu, Chen, Chen, and Lu]{local}
Xiaojie Chu, Liangyu Chen, Chengpeng Chen, and Xin Lu.
\newblock Revisiting global statistics aggregation for improving image restoration.
\newblock \emph{The European Conference on Computer Vision (ECCV)}, 2022.

\bibitem[Dong et~al.(2020)Dong, Roth, and Schiele]{deepwiener}
Jiangxin Dong, Stefan Roth, and Bernt Schiele.
\newblock Deep wiener deconvolution: Wiener meets deep learning for image deblurring.
\newblock \emph{Advances in Neural Information Processing Systems (NeurIPS)}, 33:\penalty0 1048--1059, 2020.

\bibitem[Fang et~al.(2023)Fang, Wu, Dong, Li, Wu, and Shi]{ufp}
Zhenxuan Fang, Fangfang Wu, Weisheng Dong, Xin Li, Jinjian Wu, and Guangming Shi.
\newblock Self-supervised non-uniform kernel estimation with flow-based motion prior for blind image deblurring.
\newblock \emph{The IEEE Conference on Computer Vision and Pattern Recognition (CVPR)}, pages 18105--18114, 2023.

\bibitem[Fish et~al.(1995)Fish, Brinicombe, Pike, and Walker]{richard_lucy}
DA Fish, AM Brinicombe, ER Pike, and JG Walker.
\newblock Blind deconvolution by means of the richardson--lucy algorithm.
\newblock \emph{JOSA A}, 12\penalty0 (1):\penalty0 58--65, 1995.

\bibitem[Gong et~al.(2017)Gong, Yang, Liu, Zhang, Reid, Shen, Van Den~Hengel, and Shi]{motionflow}
Dong Gong, Jie Yang, Lingqiao Liu, Yanning Zhang, Ian Reid, Chunhua Shen, Anton Van Den~Hengel, and Qinfeng Shi.
\newblock From motion blur to motion flow: A deep learning solution for removing heterogeneous motion blur.
\newblock \emph{The IEEE Conference on Computer Vision and Pattern Recognition (CVPR)}, pages 2319--2328, 2017.

\bibitem[Hradi{\v{s}} et~al.(2015)Hradi{\v{s}}, Kotera, Zemc{\i}k, and {\v{S}}roubek]{deepkernelfree}
Michal Hradi{\v{s}}, Jan Kotera, Pavel Zemc{\i}k, and Filip {\v{S}}roubek.
\newblock Convolutional neural networks for direct text deblurring.
\newblock \emph{The British Machine Vision Conference (BMVC)}, 10\penalty0 (2), 2015.

\bibitem[Kong et~al.(2023)Kong, Dong, Li, Ge, and Pan]{fftformer}
Lingshun Kong, Jiangxin Dong, Mingqiang Li, Jianjun Ge, and Jinshan Pan.
\newblock Efficient frequency domain-based transformers for high-quality image deblurring, 2023.

\bibitem[Krishnan and Fergus(2009)]{hyperlaplacian}
Dilip Krishnan and Rob Fergus.
\newblock Fast image deconvolution using hyper-laplacian priors.
\newblock \emph{Advances in neural information processing systems (NIPS)}, 22, 2009.

\bibitem[Kupyn et~al.(2018)Kupyn, Budzan, Mykhailych, Mishkin, and Matas]{deblurgan}
Orest Kupyn, Volodymyr Budzan, Mykola Mykhailych, Dmytro Mishkin, and Ji{\v{r}}{\'\i} Matas.
\newblock Deblurgan: Blind motion deblurring using conditional adversarial networks.
\newblock \emph{The IEEE Conference on Computer Vision and Pattern Recognition (CVPR)}, pages 8183--8192, 2018.

\bibitem[Kupyn et~al.(2019)Kupyn, Martyniuk, Wu, and Wang]{deblurganv2}
Orest Kupyn, Tetiana Martyniuk, Junru Wu, and Zhangyang Wang.
\newblock Deblurgan-v2: Deblurring (orders-of-magnitude) faster and better.
\newblock \emph{The IEEE International Conference on Computer Vision (ICCV)}, pages 8878--8887, 2019.

\bibitem[Li et~al.(2022)Li, Zhang, Cheung, Wang, Qin, and Li]{msdi}
Dasong Li, Yi Zhang, Ka~Chun Cheung, Xiaogang Wang, Hongwei Qin, and Hongsheng Li.
\newblock Learning degradation representations for.
\newblock \emph{The European Conference on Computer Vision (ECCV)}, pages 736--753, 2022.

\bibitem[Li et~al.(2023{\natexlab{a}})Li, Zhang, Jiang, Luo, Feng, and Xu]{reloblur}
Haoying Li, Ziran Zhang, Tingting Jiang, Peng Luo, Huajun Feng, and Zhihai Xu.
\newblock Real-world deep local motion deblurring.
\newblock \emph{Association for the Advancement of Artificial Intelligence (AAAI)}, 2023{\natexlab{a}}.

\bibitem[Li et~al.(2023{\natexlab{b}})Li, Fan, Xiang, Demandolx, Ranjan, Timofte, and Gool]{grl}
Yawei Li, Yuchen Fan, Xiaoyu Xiang, Denis Demandolx, Rakesh Ranjan, Radu Timofte, and Luc~Van Gool.
\newblock Efficient and explicit modelling of image hierarchies for image restoration.
\newblock \emph{The IEEE Conference on Computer Vision and Pattern Recognition (CVPR)}, 2023{\natexlab{b}}.

\bibitem[Loshchilov and Hutter(2016)]{cosineanneal}
Ilya Loshchilov and Frank Hutter.
\newblock Sgdr: Stochastic gradient descent with warm restarts.
\newblock \emph{arXiv preprint arXiv:1608.03983}, 2016.

\bibitem[Loshchilov and Hutter(2019)]{adamw}
Ilya Loshchilov and Frank Hutter.
\newblock Decoupled weight decay regularization.
\newblock \emph{International Conference on Learning Representations (ICLR)}, 2019.

\bibitem[Mao et~al.(2023)Mao, Liu, Liu, Li, Shen, and Wang]{deeprft}
Xintian Mao, Yiming Liu, Fengze Liu, Qingli Li, Wei Shen, and Yan Wang.
\newblock Intriguing findings of frequency selection for image deblurring.
\newblock \emph{Association for the Advancement of Artificial Intelligence (AAAI)}, 2023.

\bibitem[Nah et~al.(2017)Nah, Hyun~Kim, and Mu~Lee]{gopro}
Seungjun Nah, Tae Hyun~Kim, and Kyoung Mu~Lee.
\newblock Deep multi-scale convolutional neural network for dynamic scene deblurring.
\newblock \emph{The IEEE Conference on Computer Vision and Pattern Recognition (CVPR)}, pages 3883--3891, 2017.

\bibitem[Nah et~al.(2022)Nah, Son, Lee, and Lee]{reblur}
Seungjun Nah, Sanghyun Son, Jaerin Lee, and Kyoung~Mu Lee.
\newblock Clean images are hard to reblur: Exploiting the ill-posed inverse task for dynamic scene deblurring.
\newblock \emph{International Conference on Learning Representations (ICLR)}, 2022.

\bibitem[Rim et~al.(2020)Rim, Lee, Won, and Cho]{realblur}
Jaesung Rim, Haeyun Lee, Jucheol Won, and Sunghyun Cho.
\newblock Real-world blur dataset for learning and benchmarking deblurring algorithms.
\newblock \emph{The European Conference on Computer Vision (ECCV)}, pages 184--201, 2020.

\bibitem[Rim et~al.(2022)Rim, Kim, Kim, Lee, Lee, and Cho]{rsblur}
Jaesung Rim, Geonung Kim, Jungeon Kim, Junyong Lee, Seungyong Lee, and Sunghyun Cho.
\newblock Realistic blur synthesis for learning image deblurring.
\newblock \emph{The European Conference on Computer Vision (ECCV)}, 2022.

\bibitem[Ronneberger et~al.(2015)Ronneberger, Fischer, and Brox]{unet}
Olaf Ronneberger, Philipp Fischer, and Thomas Brox.
\newblock U-net: Convolutional networks for biomedical image segmentation.
\newblock \emph{International Conference on Medical image computing and computer-assisted intervention}, pages 234--241, 2015.

\bibitem[Schuler et~al.(2015)Schuler, Hirsch, Harmeling, and Sch{\"o}lkopf]{bd2}
Christian~J Schuler, Michael Hirsch, Stefan Harmeling, and Bernhard Sch{\"o}lkopf.
\newblock Learning to deblur.
\newblock \emph{IEEE transactions on pattern analysis and machine intelligence}, 38\penalty0 (7):\penalty0 1439--1451, 2015.

\bibitem[Sun et~al.(2015)Sun, Cao, Xu, and Ponce]{deepkernel1}
Jian Sun, Wenfei Cao, Zongben Xu, and Jean Ponce.
\newblock Learning a convolutional neural network for non-uniform motion blur removal.
\newblock \emph{The IEEE Conference on Computer Vision and Pattern Recognition (CVPR)}, pages 769--777, 2015.

\bibitem[Tao et~al.(2018)Tao, Gao, Shen, Wang, and Jia]{srndeblurnet}
Xin Tao, Hongyun Gao, Xiaoyong Shen, Jue Wang, and Jiaya Jia.
\newblock Scale-recurrent network for deep image deblurring.
\newblock \emph{The IEEE Conference on Computer Vision and Pattern Recognition (CVPR)}, pages 8174--8182, 2018.

\bibitem[Tran et~al.(2021)Tran, Tran, Phung, and Hoai]{bluroperator}
Phong Tran, Anh~Tuan Tran, Quynh Phung, and Minh Hoai.
\newblock Explore image deblurring via encoded blur kernel space.
\newblock \emph{The IEEE Conference on Computer Vision and Pattern Recognition (CVPR)}, pages 11956--11965, 2021.

\bibitem[Tsai et~al.(2022)Tsai, Peng, Lin, Tsai, and Lin]{stripformer}
Fu-Jen Tsai, Yan-Tsung Peng, Yen-Yu Lin, Chung-Chi Tsai, and Chia-Wen Lin.
\newblock Stripformer: Strip transformer for fast image deblurring.
\newblock \emph{The European Conference on Computer Vision (ECCV)}, 2022.

\bibitem[Tu et~al.(2022)Tu, Talebi, Zhang, Yang, Milanfar, Bovik, and Li]{maxim}
Zhengzhong Tu, Hossein Talebi, Han Zhang, Feng Yang, Peyman Milanfar, Alan Bovik, and Yinxiao Li.
\newblock Maxim: Multi-axis mlp for image processing.
\newblock \emph{The IEEE Conference on Computer Vision and Pattern Recognition (CVPR)}, pages 5769--5780, 2022.

\bibitem[Wang et~al.(2019)Wang, Chan, Yu, Dong, and Change~Loy]{edvr}
Xintao Wang, Kelvin~CK Chan, Ke Yu, Chao Dong, and Chen Change~Loy.
\newblock Edvr: Video restoration with enhanced deformable convolutional networks.
\newblock \emph{The IEEE Conference on Computer Vision and Pattern Recognition Workshops (CVPRW)}, 2019.

\bibitem[Wang et~al.(2004)Wang, Bovik, Sheikh, and Simoncelli]{ssim}
Zhou Wang, Alan~C Bovik, Hamid~R Sheikh, and Eero~P Simoncelli.
\newblock Image quality assessment: from error visibility to structural similarity.
\newblock \emph{IEEE transactions on image processing}, 13\penalty0 (4):\penalty0 600--612, 2004.

\bibitem[Wang et~al.(2022)Wang, Cun, Bao, Zhou, Liu, and Li]{uformer}
Zhendong Wang, Xiaodong Cun, Jianmin Bao, Wengang Zhou, Jianzhuang Liu, and Houqiang Li.
\newblock Uformer: A general u-shaped transformer for image restoration.
\newblock \emph{The IEEE Conference on Computer Vision and Pattern Recognition (CVPR)}, pages 17683--17693, 2022.

\bibitem[Zamir et~al.(2021)Zamir, Arora, Khan, Hayat, Khan, Yang, and Shao]{mprnet}
Syed~Waqas Zamir, Aditya Arora, Salman Khan, Munawar Hayat, Fahad~Shahbaz Khan, Ming-Hsuan Yang, and Ling Shao.
\newblock Multi-stage progressive image restoration.
\newblock \emph{The IEEE Conference on Computer Vision and Pattern Recognition (CVPR)}, pages 14821--14831, 2021.

\bibitem[Zamir et~al.(2022)Zamir, Arora, Khan, Hayat, Khan, and Yang]{restormer}
Syed~Waqas Zamir, Aditya Arora, Salman Khan, Munawar Hayat, Fahad~Shahbaz Khan, and Ming-Hsuan Yang.
\newblock Restormer: Efficient transformer for high-resolution image restoration.
\newblock \emph{The IEEE Conference on Computer Vision and Pattern Recognition (CVPR)}, pages 5728--5739, 2022.

\bibitem[Zhang et~al.(2021)Zhang, Wang, Maybank, and Tao]{exposuretrajectory}
Youjian Zhang, Chaoyue Wang, Stephen~J Maybank, and Dacheng Tao.
\newblock Exposure trajectory recovery from motion blur.
\newblock \emph{IEEE Transactions on Pattern Analysis and Machine Intelligence}, 44\penalty0 (11):\penalty0 7490--7504, 2021.

\end{thebibliography}
}

\clearpage
\onecolumn
\centerline{\Large{\textbf{\thetitle}}}
\vspace{1.0em}
\centerline{\large{Supplementary Material}}
\vspace{1.0em}

\appendix
\counterwithin{figure}{section}
\counterwithin{table}{section}
\section{Implementation Details}\label{sec:app_details}
\subsection{Our model variants}\label{sec:app_variants}
The default number of blocks of NAFNet~\cite{nafnet} is $36$ which consists of encoder blocks $\{1,1,1,28\}$, middle block $\{1\}$ and decoder blocks $\{1,1,1,28\}$. NAFNet with 32 widths and 64 widths are referred to as NAFNet-32 and NAFNet-64, respectively.
To be consistent with the computational cost of NAFNet, we build SegDeblur-S ($14.44$ GMACs) and SegDeblur-L ($62.68$ GMACs) for the realistic datasets such as RealBlur~\cite{realblur}, RSBlur~\cite{rsblur} and ReLoBlur~\cite{reloblur}.
Since FFTFormer~\cite{fftformer} shows the best performance on GoPro~\cite{gopro}, we build SegFFTFormer ($135.81$ GMACs) for GoPro.
Our discrete-to-continuous (D2C) converter of SegDeblur-S is based on NAFNet and consists of encoder blocks $\{8,8,8,22\}$, middle block $\{8\}$ and decoder blocks $\{8,8,8,8\}$ with $16$ widths ($10.07$ GMACs). Also, our D2C converter of SegDeblur-L is also based on NAFNet and consists of encoder blocks $\{5,5,5,18\}$, middle block $\{5\}$ and decoder blocks $\{5,5,5,5\}$ with $48$ widths ($58.31$ GMACs). For SegFFTFormer, we use the original FFTFormer ($131.45$ GMACs) for our D2C converter. On the other hand, our blur pixel discretizer is based on NAFNet and shared with our SegDeblur-S, SegDeblur-L and SegFFTFormer, which consists of encoder blocks $\{2,2,2,20\}$, middle block $\{2\}$ and decoder blocks $\{2,2,2,2\}$ with $16$ widths ($4.37$ GMACs). For our kernel estimator, we use U-Net~\cite{unet} with $16$ widths ($1.16$ GMACs), which performs four times of downsampling by stride $2$ operations. We remark that our kernel estimator is only used for training, such that it is not included in our computational cost. 
Overall, our models are summarized in Table~\ref{tbl:ourmodel}.

\begin{table}[ht]\centering
\scalebox{0.95}{
\begin{tabular}{c|c|c|c|c}
\hline \multirow{2}{*}{Our Model} & \multicolumn{2}{c|}{Model (GMACs)} & \multirow{2}{*}{Total GMACs} & \multirow{2}{*}{Total \# Params (M)}\\
\cline{2-3} & {Blur Pixel Discretizer} & {\;\;\;\;D2C Converter\;\;\;\;}  & 	   & \\
\hline {SegDeblur-S} &{$4.37$} &{$10.07$} & {$14.44$} & {$12.30$} \\						
{SegDeblur-L} &{$4.37$} &{$58.31$}  & {$62.68$} & {$55.40$}  \\
{SegFFTFormer} &{$4.37$} &{$131.45$}  & {$135.81$} & {$20.60$}  \\
\hline		
\end{tabular}
}
\vspace{-0.2cm}
\vspace{0.01in}
\caption{Our model variants. We present a detailed description of our individual variants.} \label{tbl:ourmodel}
\vspace{-0.2cm}
\end{table}

\section{Additional Ablation Study}\label{sec:app_ablation}
\subsection{Effects on deblurring network architecture}\label{sec:app_net}
We investigate that our blur segmentation map still works well on the other network architectures. To confirm this, we apply our blur segmentation map to the network architectures such as NAFNet~\cite{nafnet}, FFTFormer~\cite{fftformer} and FMIMOUNet~\cite{deeprft}. We train the deblurring model with RealBlur-J~\cite{realblur}. Note that the network architectures such as NAFNet, FFTFormer and FMIMOUNet are scaled down to around 16 GMACs. The results are given in Table~\ref{tbl:ablation_net}.

\begin{table}[ht]
    \centering
\scalebox{0.99}{
    \begin{tabular}{c|c|c}
\hline         &  \multirow{2}{*}{Prior}& RealBlur-J \\
\cline{3-3}         & &  \;PSNR $\uparrow$\;\;SSIM $\uparrow$\\ 
\hline         \multirow{2}{*}{NAFNet~\cite{nafnet}}&  & 31.99\;\;\;\;\;0.920\\  & \checkmark & $\textbf{32.53}$\;\;\;\;\;$\textbf{0.927}$\\
\hline     \multirow{2}{*}{FFTFormer~\cite{fftformer}}&  & 32.08\;\;\;\;\;0.917\\  & \checkmark & $\textbf{32.73}$\;\;\;\;\;$\textbf{0.926}$\\
\hline      \multirow{2}{*}{FMIMOUNet~\cite{deeprft}}&  & 31.25\;\;\;\;\;0.903\\  & \checkmark & $\textbf{32.30}$\;\;\;\;\;$\textbf{0.920}$\\
\hline
    \end{tabular}}
    \vspace{-0.2cm}
    \caption{Ablation study on network architectures.~\cite{nafnet,fftformer,deeprft}. All network architectures are based on $16$ GMACs.}
    \label{tbl:ablation_net}
    \vspace{-0.3cm}
\end{table}

\subsection{Effects on model size}\label{sec:app_modelsize}
Our model consists of blur pixel discretizer and D2C converter.  Here, we raise a fundamental question: which model size should we increase to improve performance?
To address this, we conduct experiments with various model sizes by scaling the size of blur pixel discretizer and D2C converter.
The results are shown in Table~\ref{tbl:ablation_modelsize}.
We observe that the small model size of the blur pixel discretizer is sufficient to provide high deblurring performance (compare the performance in 1st-3rd row results and 2nd-4th row results). This is because the classification task (e.g., blur pixel discretization) is easier than the regression task (e.g., discrete-to-continuous conversion). Meanwhile, the model size of the D2C converter is the most important ingredient to improve performance as shown in Table~\ref{tbl:ablation_modelsize}.
\begin{table}[ht]\centering
\scalebox{0.95}{
\begin{tabular}{c|c|c|c|c}
\hline \multicolumn{3}{c|}{MACs (G)} & \multirow{2}{*}{PSNR} & \multirow{2}{*}{SSIM}\\
\cline{1-3} {Blur Pixel Discretizer} & {\;\;\;\;D2C converter\;\;\;\;} & {\;\;\;\;\;Total\;\;\;\;\;}  & 	   & \\
\hline {4.37} &{10.07} &{14.44} & {$32.53$} & {$0.927$} \\						
{4.37} &{58.31} &{62.68} & {$\textbf{32.95}$} & {$\textbf{0.934}$} \\
{10.07} &{4.37} &{14.44} & {$32.18$} & {$0.919$} \\
{58.31} &{4.37}&{62.68} & {$32.65$} & {$0.923$} \\
\hline		
\end{tabular}
}
\vspace{-0.2cm}
\vspace{0.01in}
\caption{Ablation study on model sizes of our method. We make several combinations of our blur pixel discretizer and D2C converter to evaluate which model type contributes to more performance improvement when varying the size of each model.}
\label{tbl:ablation_modelsize}
\vspace{-0.2cm}
\end{table}

\subsection{Cross-data validation}\label{sec:app_cross}
To verify the generalization ability, we conduct the cross-data test. Namely, we train with RealBlur-J~\cite{realblur} and evaluate on RSBlur~\cite{rsblur}. As reported in Table~\ref{tbl:ablation_cross}, our SegDeblur-L gives better performance than NAFNet~\cite{nafnet} at similar computational cost. Although the performance gap ($0.2$ dB) seems small, this performance gap may require $4 \times$ increase in computational cost in RSBlur, as demonstrated in Table~\ref{tbl:rsblur}.

\begin{table}[!h]\centering
\scalebox{0.90}{
\begin{tabular}{c|c|c|c}
\hline \multirow{2}{*}{Methods}	& \multirow{2}{*}{GMACs}  & \multicolumn{2}{c}{RealBlur (Train) - RSBlur (Test)} \\
\cline{3-4}                     &                          & {PSNR $\uparrow$} & {SSIM $\uparrow$}\\
\hline {NAFNet~\cite{srndeblurnet}}	   & {$63.64$} & \;\;\;\;\;\;{$30.56$}\;\;\;\;\;\; & {$0.806$}\\
{UFPNet~\cite{ufp}}	           & {$243.33$} & {$30.75$} & {$0.812$}\\
{SegDeblur-L (ours)}	                   & {$62.68$} & {$\textbf{30.76}$} & {$\textbf{0.813}$}\\
\hline		
\end{tabular}
}
\vspace{0.02in}
\vspace{-0.3cm}
\caption{Ablation study on the cross-data validation. We train with RealBlur-J~\cite{realblur} and evaluate on RSBlur~\cite{rsblur} The best results are indicated in bold.}
\label{tbl:ablation_cross}
\vspace{-0.3cm}
\end{table}
\vspace{0.15in}
\vspace{-0.3cm}

\subsection{Different datasets for blur pixel discretizer and D2C converter}\label{sec:app_diff}
To confirm how different datasets affect the performance, we conduct experiments in which we use the blur pixel discretizers trained with RealBlur-J~\cite{realblur}, RSBlur~\cite{rsblur} and GoPro~\cite{gopro} and train the D2C converter with RealBlur-J. The results are presented in Table~\ref{tbl:ablation_diff}. The results show that it is crucial to use the same data in both blur pixel discretizer and D2C converter. On the other hand, using the different datasets leads to significant performance drop as shown in Table~\ref{tbl:ablation_diff}. We believe that this is due to vulnerability to unseen data in deep models as well as variations arising from different image sensors, lenses, ISPs and motion types. 

\begin{table}[!h]\centering
\scalebox{0.90}{
\begin{tabular}{c|c|c|c}
\hline \multicolumn{2}{c|}{Training Set}  & \multirow{2}{*}{PSNR $\uparrow$} & \multirow{2}{*}{SSIM $\uparrow$} \\
\cline{1-2}  Blur Pixel Discretizer	& D2C Converter  &  &  \\
\hline {GoPro~\cite{gopro}}	   & \multirow{3}{*}{RealBlur-J} & {$31.89$} & {$0.919$}\\
{RSBlur~\cite{gopro}}	           &  & {$31.93$} & {$0.921$}\\
{RealBlur-J~\cite{realblur}}	                   &  & {$\textbf{32.53}$} & {$\textbf{0.927}$}\\
\hline		
\end{tabular}
}
\vspace{0.02in}
\vspace{-0.3cm}
\caption{Ablation study on different datasets. To confirm the effect on different datasets for two models, we train with GoPro~\cite{gopro}, RSBlur~\cite{rsblur} and RealBlur-J~\cite{realblur} for our blur pixel discretizer and train with RealBlur-J for our D2C converter. The best results are indicated in bold.}
\label{tbl:ablation_diff}
\vspace{-0.3cm}
\end{table}
\vspace{0.15in}
\vspace{-0.3cm}

\subsection{Performance of large motion test set in RSBlur}\label{sec:app_largemotion}
As shown in Table~\ref{tbl:efficient}, we demonstrate that our method is robust to large motion test set in RealBlur-J~\cite{realblur}. To additionally verify the robustness of large motion scenarios for our method, we construct the large motion test set of RSBlur~\cite{rsblur} by extracting the largest motion $300$ image pairs. The results are presented in Table~\ref{tbl:ablation_largemotion}. Our efficient model, i.e., SegDeblur-S, improves PSNR in the large motion set ($30.04\rightarrow 30.39$ dB) compared with the counterpart, NAFNet-32. Furthermore, its large motion performance is also comparable to that of the larger model, NAFNet-64 ($30.39$ dB).

\begin{table}[!ht]\centering
\scalebox{0.95}{
\begin{tabular}{c|c|c|c}
\hline \multirow{2}{*}{Methods}	& \multirow{2}{*}{GMACs}  & \multicolumn{2}{c}{RSBlur}\\
\cline{3-4}                    &                          & {Total} & {Large}\\
\hline {NAFNet-32~\cite{nafnet}}	            & {16.25} & {$33.71$} &{$30.04$}\\
{NAFNet-64~\cite{nafnet}}	            & {63.64} & {$33.97$} &{$30.39$}\\
{SegDeblur-S (ours)}	& {14.44} &  {$33.96$} &{$30.39$}\\
{SegDeblur-L (ours)}	& {62.68} &  {$\textbf{34.21}$} &{$\textbf{30.82}$}\\
\hline		
\end{tabular}
}
\vspace{0.01in}
\vspace{-0.2cm}
\caption{Performance of large motion test set in RSBlur~\cite{rsblur}. ``Total'' means the whole set while ``Large'' denotes the large motion set.}
\label{tbl:ablation_largemotion}
\vspace{-0.3cm}
\end{table}

\subsection{Comparison with kernel-based methods}
As the kernel-based methods produce pixel-wise motion information such as motion trajectories and kernels, we can use such information as prior when training the deblurring model. To this end, we train the deblurring model (NAFNet-32) with such information and compare their results with our method. As shown in Table~\ref{tbl:ablation_kb}, simply introducing prior information does not highly improve performance compared to NAFNet-32. Basically, estimating motion information for every pixel is a huge ill-posed problem. Therefore, their methods may produce inaccurate information, which is not helpful for subsequent deblurring tasks.
On the other hand, our method overcomes the ill-posedness by introducing the latent sharp image and logarithmic fourier space as discussed in Section~\ref{sec:kest}, which leads to better deblurring performance as shown in Table~\ref{tbl:ablation_kb}.

\begin{table}[ht]
    \centering
\scalebox{0.99}{
    \begin{tabular}{c|c|c|c}
\hline         &  Prior&  GMACs&\;PSNR $\uparrow$\;\;SSIM $\uparrow$\\    
\hline         NAFNet-32~\cite{nafnet}&  &  16.25& 31.99\;\;\;\;\;0.920\\
         Adaptive Basis~\cite{adaptivebasis}&  \checkmark&  80.60& 32.08\;\;\;\;\;0.921\\
         Exposure Trajectory~\cite{exposuretrajectory}&  \checkmark&  72.58& 31.94\;\;\;\;\;0.919\\
         SegDeblur-S (ours)&  \checkmark&  14.44& \textbf{32.53}\;\;\;\;\;\textbf{0.927}\\
\hline
    \end{tabular}}
    \vspace{-0.2cm}
    \caption{Comparison with kernel-based methods~\cite{adaptivebasis,exposuretrajectory}.}
    \label{tbl:ablation_kb}
    \vspace{-0.3cm}
\end{table}

\section{Implementation details for mobile deployment}\label{sec:app_deploy}
We modify some network architecture of our model according to the operations supported by AI accelerators in order to accelerate our method, which is called SegDeblur-S+.
Specifically, SimpleGate, SCA, and LayerNorm in NAFNet~\cite{nafnet} are removed.
We train SegDeblur-S+ with a combination of RealBlur~\cite{realblur} and RSBlur~\cite{rsblur}. 
We deploy our SegDeblur-S+ using 32-bit floating point precision on GPU of Qualcomm SM8550 chipset in Samsung Galaxy S23 without any quantization. The average inference time of our method is measured using TensorFlow Lite Benchmark Tool.
Furthermore, we measure the averaged execution time in Samsung EnhanceX and Google Unblur based on the image size of 2000$\times$2000.
The comparison result for on-chip execution time is presented in Table~\ref{tbl:app_latency}.
Notably, our SegDeblur-S takes nearly twice of the execution time compared to the accelerated version of our method, SegDeblur-S+. This accounts for the importance of using the network architectures supported by AI accelerators for the mobile deployment.
The on-chip execution time of SegDeblur-S+ ($2.35$s) lags behind that of Samsung EnhanceX ($1.64$s) and Google Unblur ($2.03$s). Our current implementation is based on on-chip GPU, such that it can be more accelerated if deployed on Neural Processing Unit (NPU). We conjecture that Samsung EnhanceX and Google Unblur are implemented on NPU to achieve faster processing.

\begin{table}[!ht]\centering
\begin{tabular}{c|c|c|c}
\hline {Methods}		                    & {Device}			& {Time} & {GMACs}\\
\hline{Samsung EnhanceX}    &{-}			    & {$1.64$s} &{-}\\
{Google UnBlur}    &{-}			    & {$2.03$s} &{-}\\
\hline{SegDeblur-S+ (ours)}    &\multirow{2}{*}{GPU}			    & {$2.35$s} & {$14.05$}\\
{SegDeblur-S (ours)}    &			    & {$4.10$s} &{$14.44$}\\
\hline		
\end{tabular}
\vspace{0.01in}
\vspace{-0.2cm}
\caption{On-chip execution time. We measure the averaged execution time on image size of $2000\times2000$.}
\label{tbl:app_latency}
\vspace{-0.2cm}
\end{table}

\section{Generalization to real-world blur images}\label{sec:app_realworld}
\begin{figure*}[!ht]
\begin{minipage}[!ht]{.51\linewidth}
  \centering
  \centerline{\includegraphics[width=8.6cm]{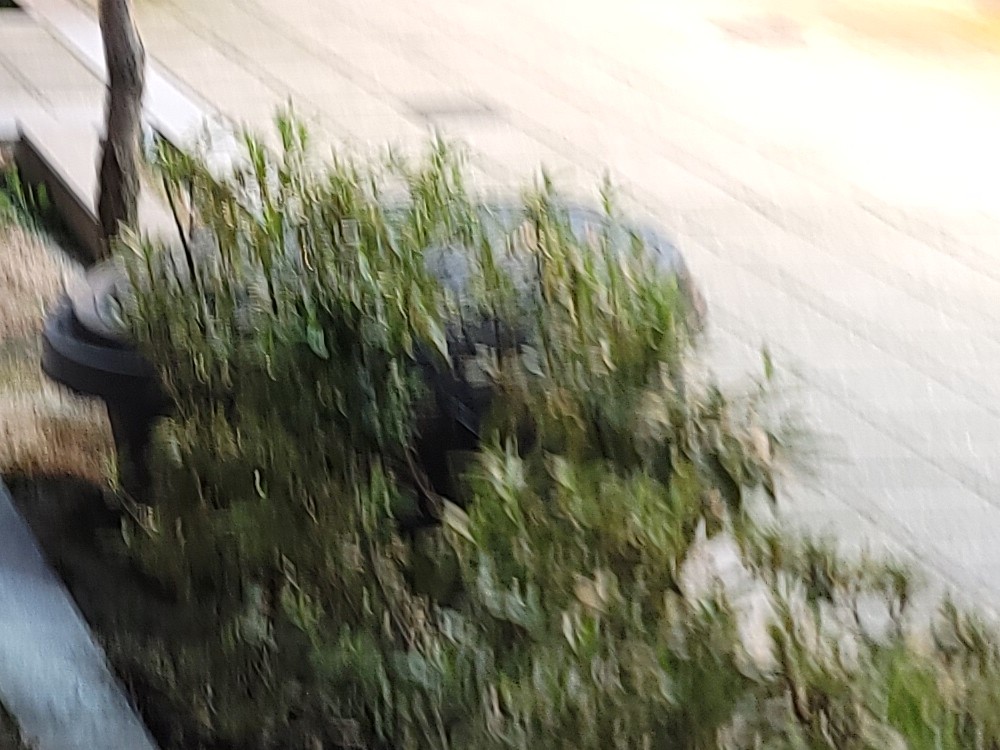}}
  \centerline{(a) Blur Input}\medskip
  \centerline{\includegraphics[width=8.6cm]{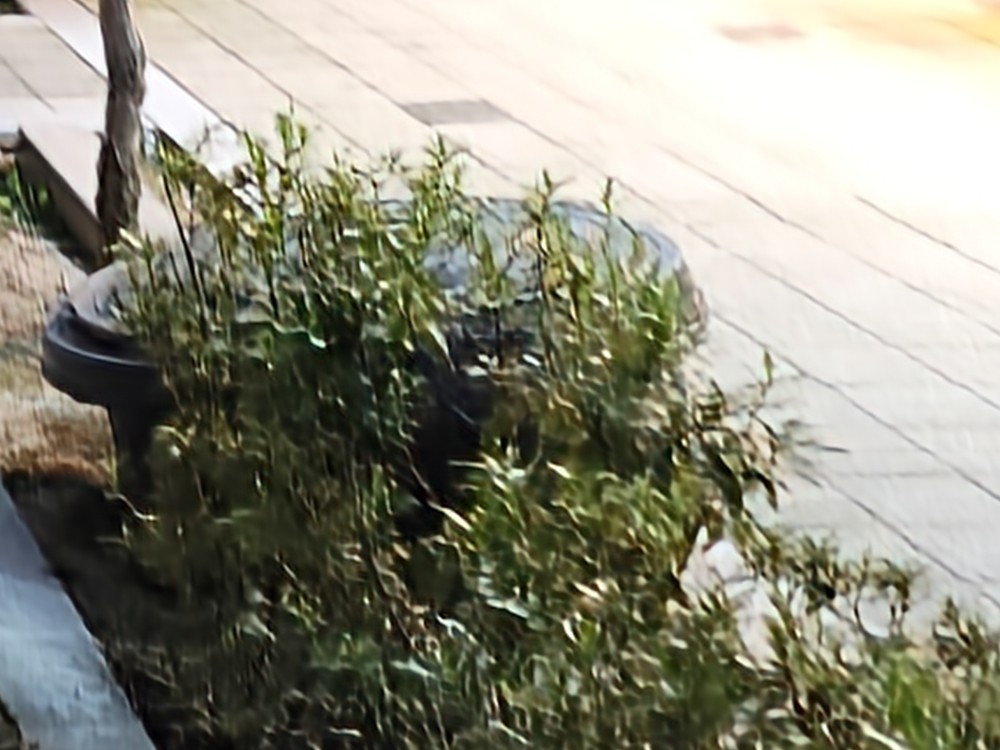}}
  \centerline{(c) Google Unblur}\medskip
  \centerline{\includegraphics[width=8.6cm]{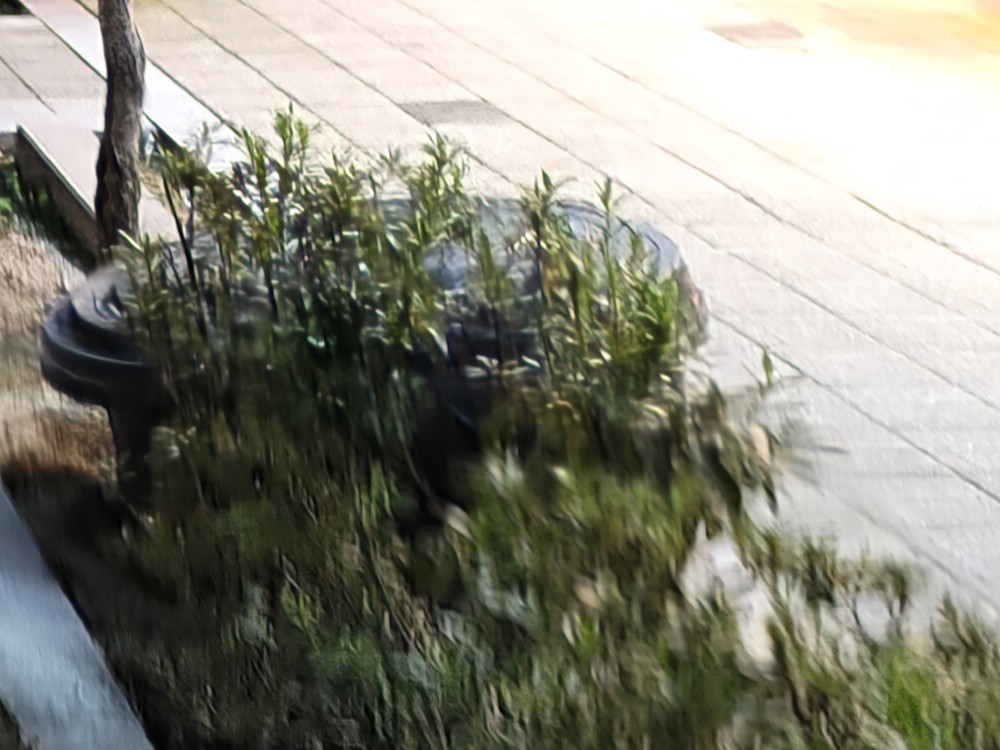}}
  \centerline{(e) FFTFormer-16~\cite{fftformer}}\medskip
\end{minipage}
\begin{minipage}[!ht]{.51\linewidth}
  \centering
  \centerline{\includegraphics[width=8.6cm]{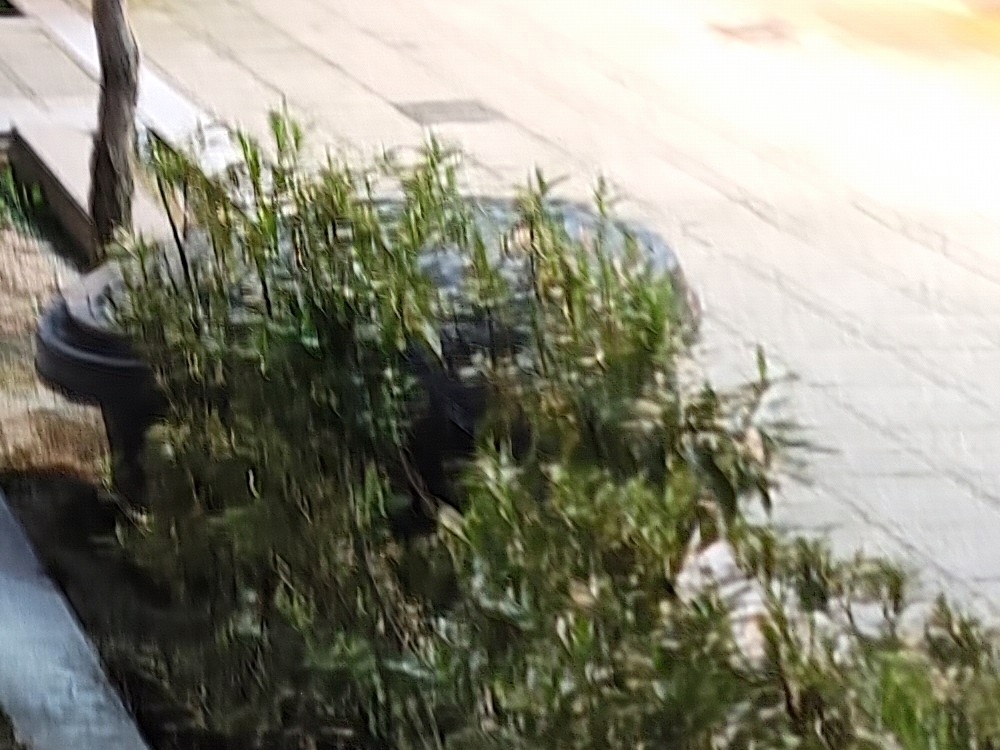}}
  \centerline{(b) Samsung EnhanceX}\medskip
  \centerline{\includegraphics[width=8.6cm]{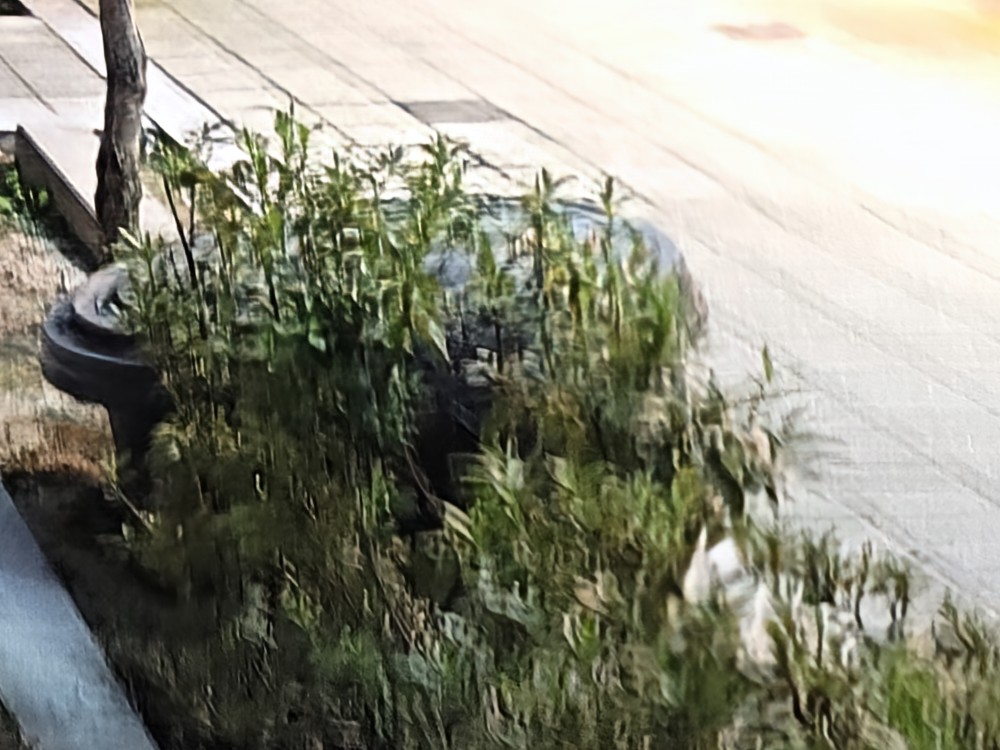}}
  \centerline{(d) NAFNet-32~\cite{nafnet}}\medskip
  \centerline{\includegraphics[width=8.6cm]{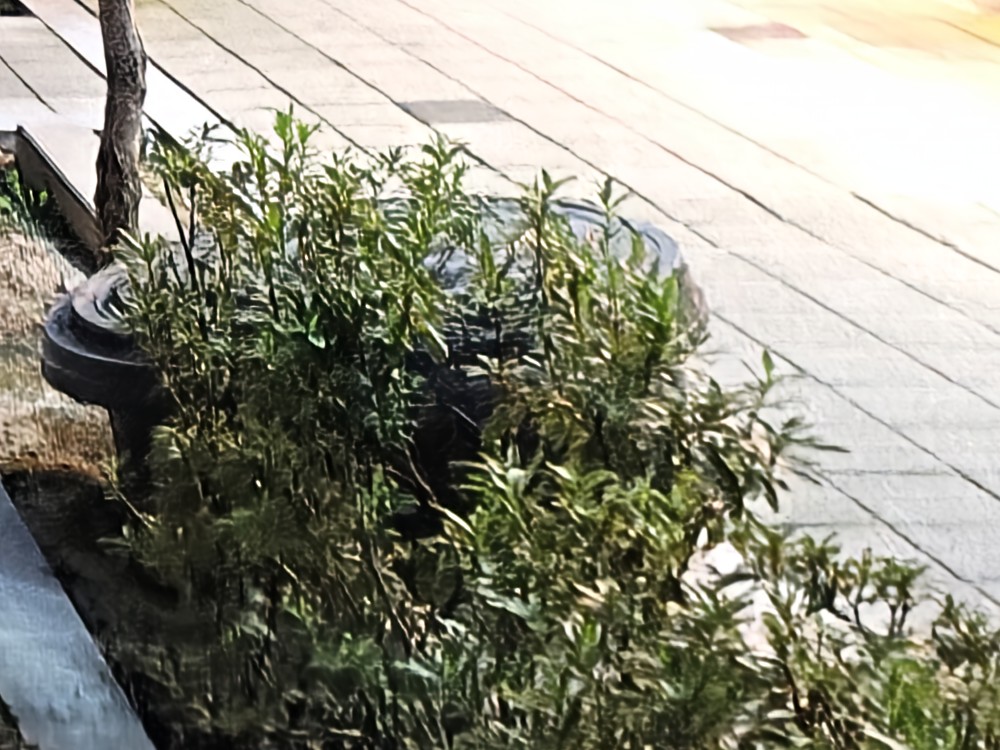}}
  \centerline{(f) SegDeblur-S (ours)}\medskip
\end{minipage}
\caption{Visual comparison results on real-world blur images.}
\label{fig:supp_realworld1}
\end{figure*}

\begin{figure*}[!ht]
\begin{minipage}[!ht]{.51\linewidth}
  \centering
  \centerline{\includegraphics[width=8.6cm]{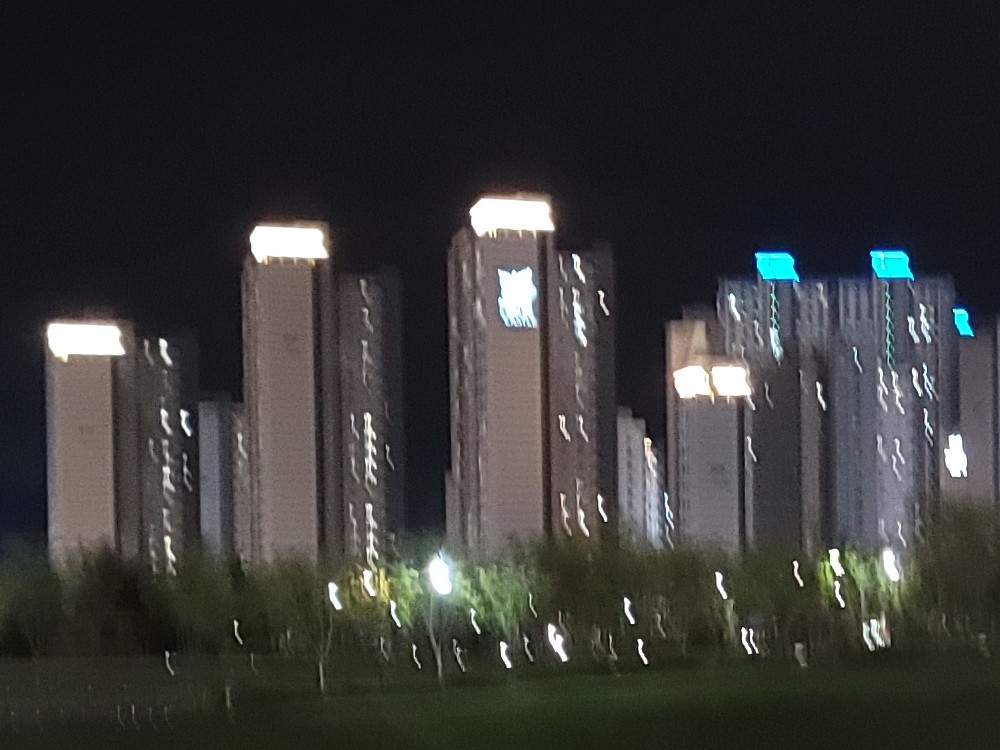}}
  \centerline{(a) Blur Input}\medskip
  \centerline{\includegraphics[width=8.6cm]{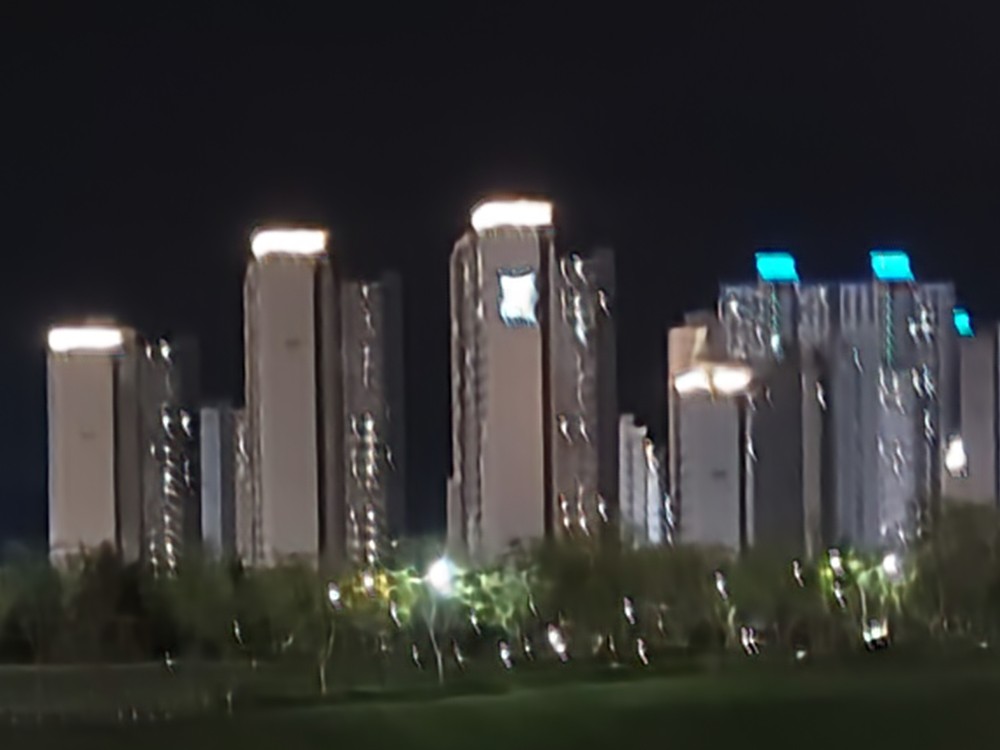}}
  \centerline{(c) Google Unblur}\medskip
  \centerline{\includegraphics[width=8.6cm]{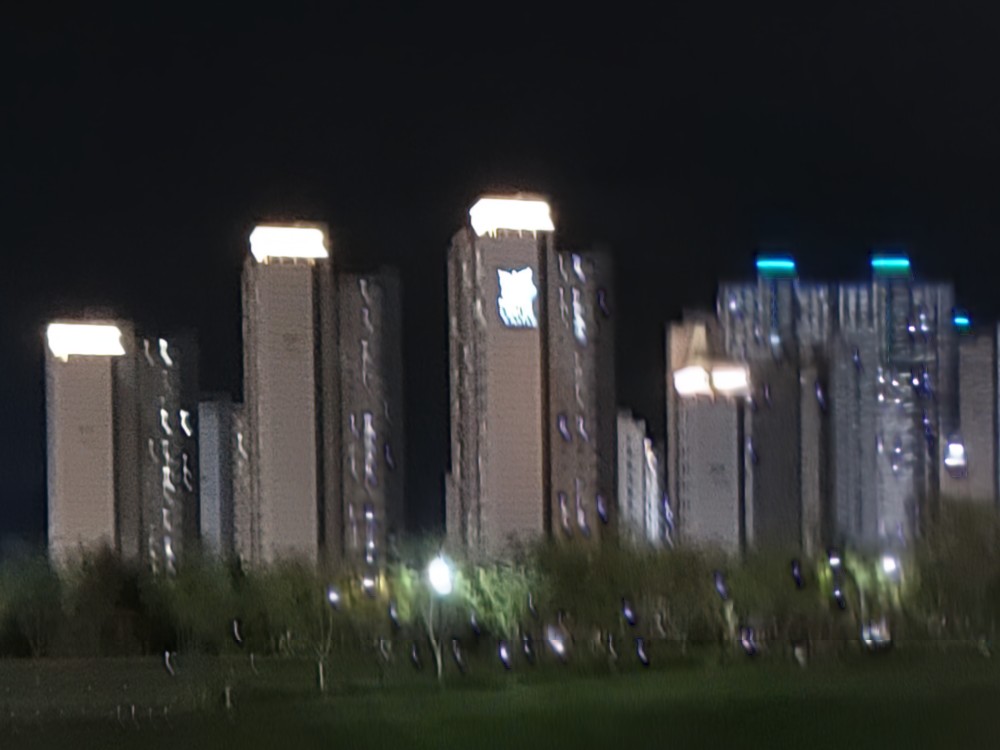}}
  \centerline{(e) FFTFormer-16~\cite{fftformer}}\medskip
\end{minipage}
\begin{minipage}[!h]{.51\linewidth}
  \centering
  \centerline{\includegraphics[width=8.6cm]{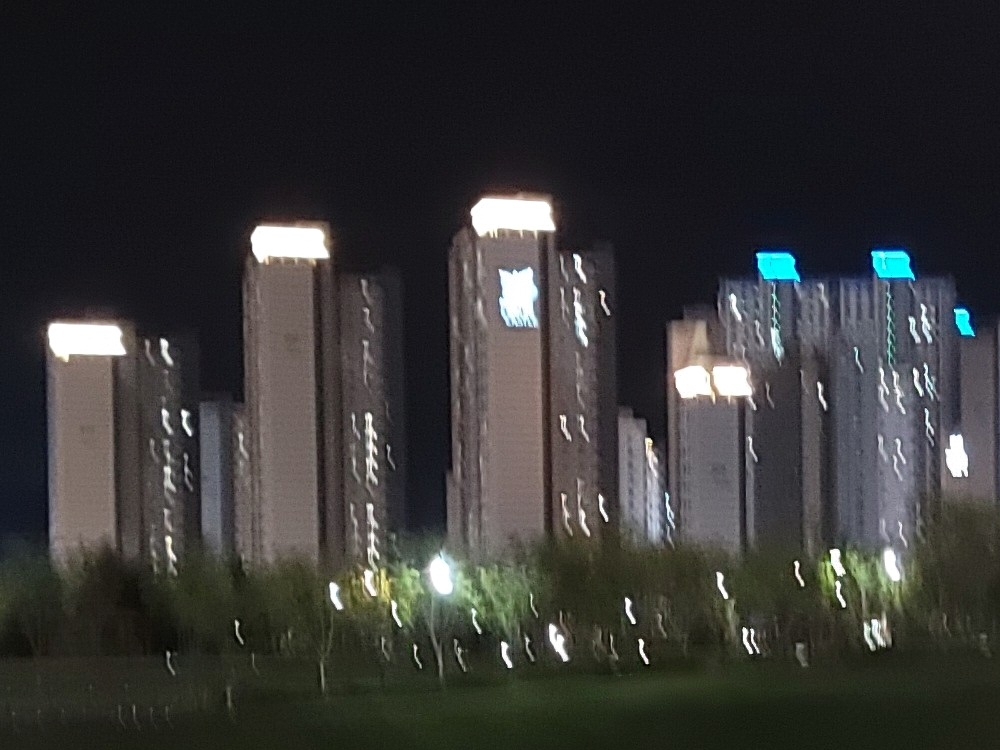}}
  \centerline{(b) Samsung EnhanceX}\medskip
  \centerline{\includegraphics[width=8.6cm]{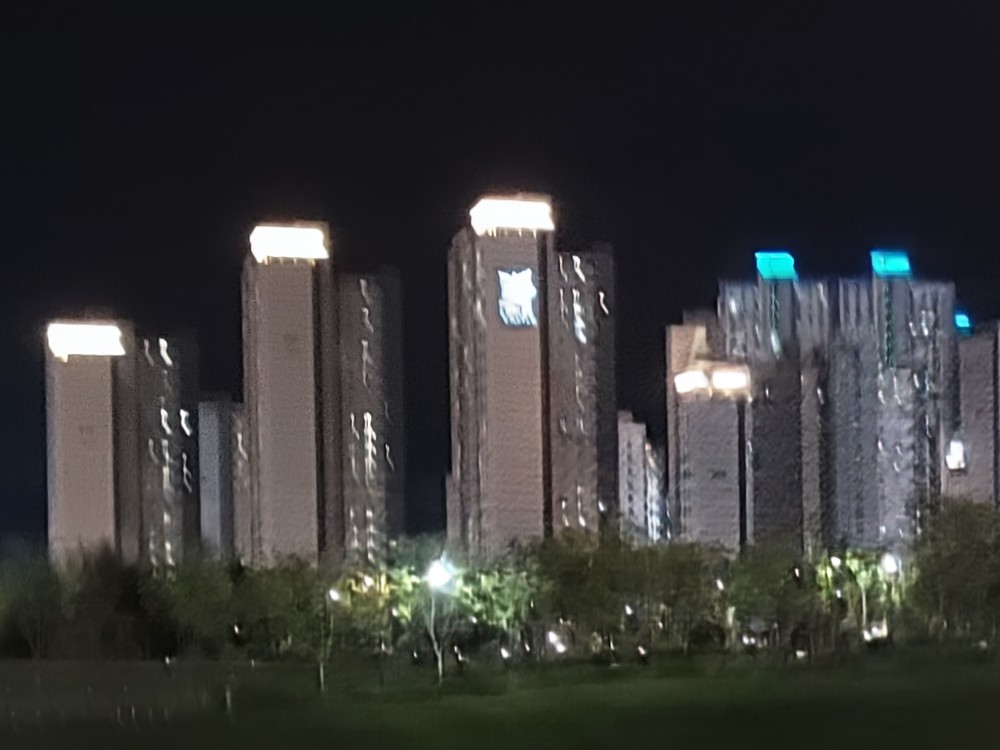}}
  \centerline{(d) NAFNet-32~\cite{nafnet}}\medskip
  \centerline{\includegraphics[width=8.6cm]{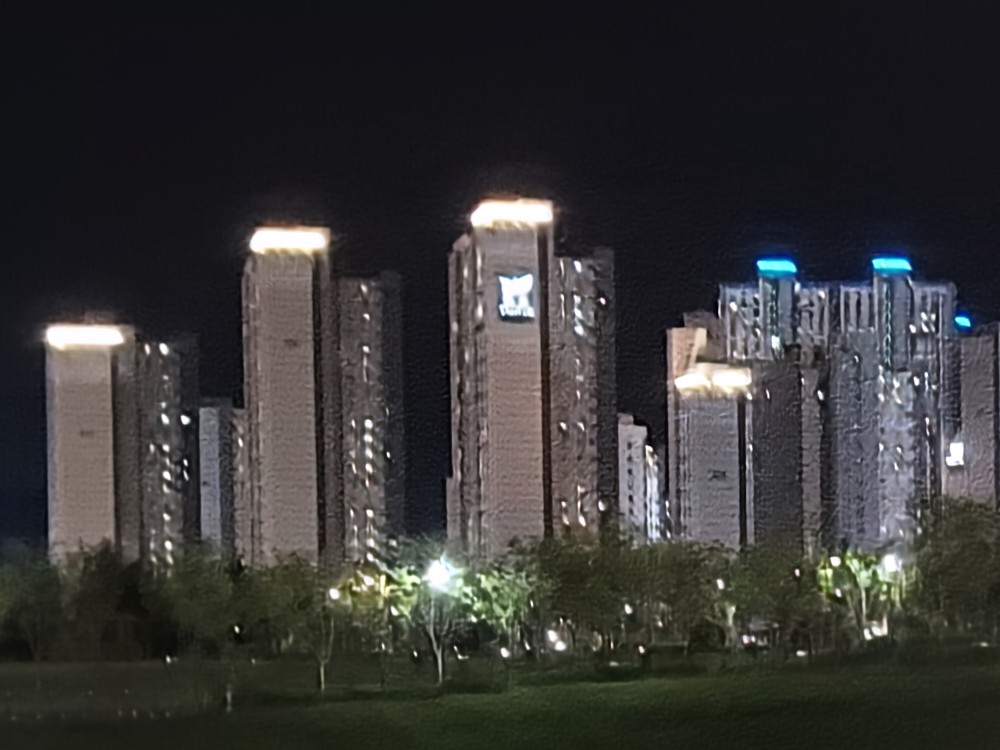}}
  \centerline{(f) SegDeblur-S (ours)}\medskip
\end{minipage}
\caption{Visual comparison results on real-world blur images.}
\label{fig:supp_realworld2}
\end{figure*}

\begin{figure*}[!ht]
\begin{minipage}[!ht]{.51\linewidth}
  \centering
  \centerline{\includegraphics[width=8.6cm]{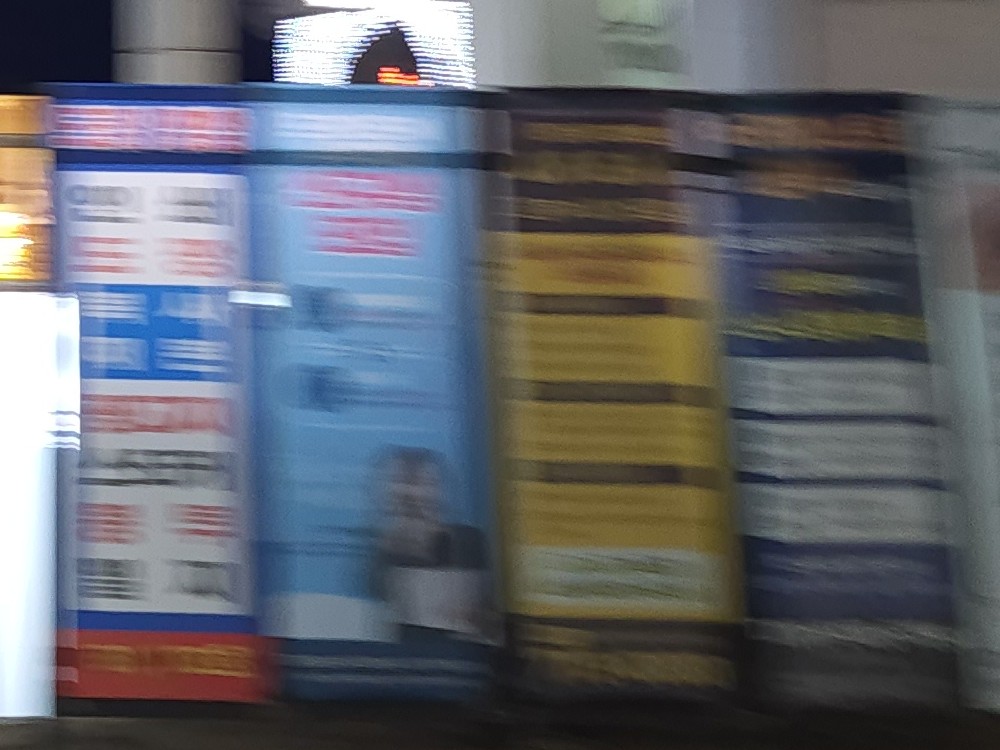}}
  \centerline{(a) Blur Input}\medskip
  \centerline{\includegraphics[width=8.6cm]{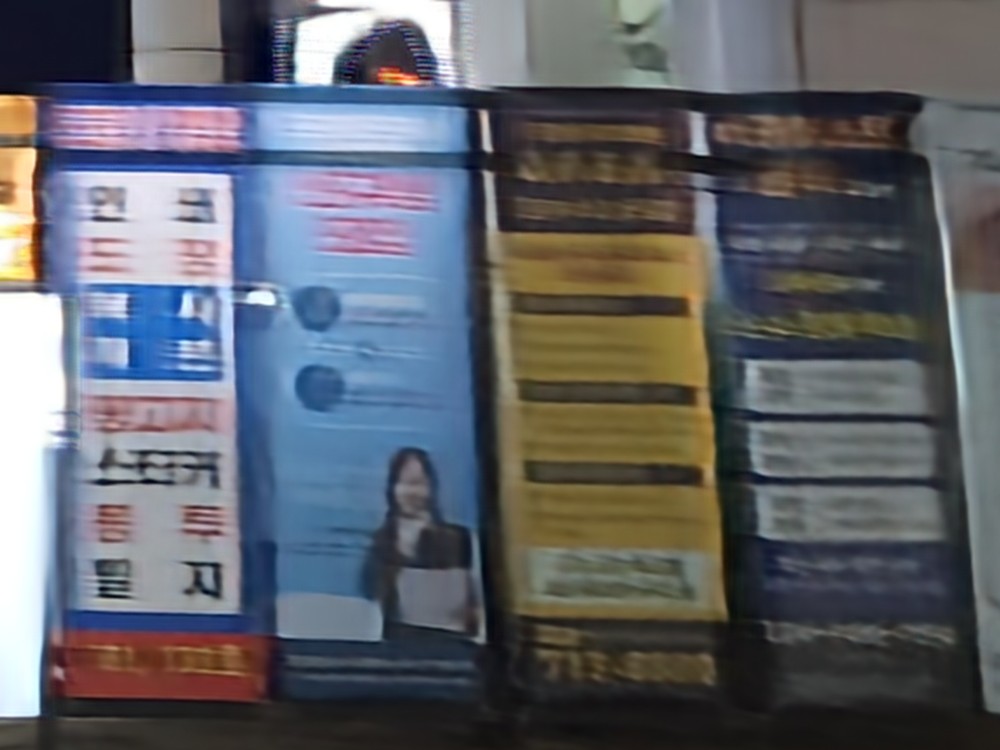}}
  \centerline{(c) Google Unblur}\medskip
  \centerline{\includegraphics[width=8.6cm]{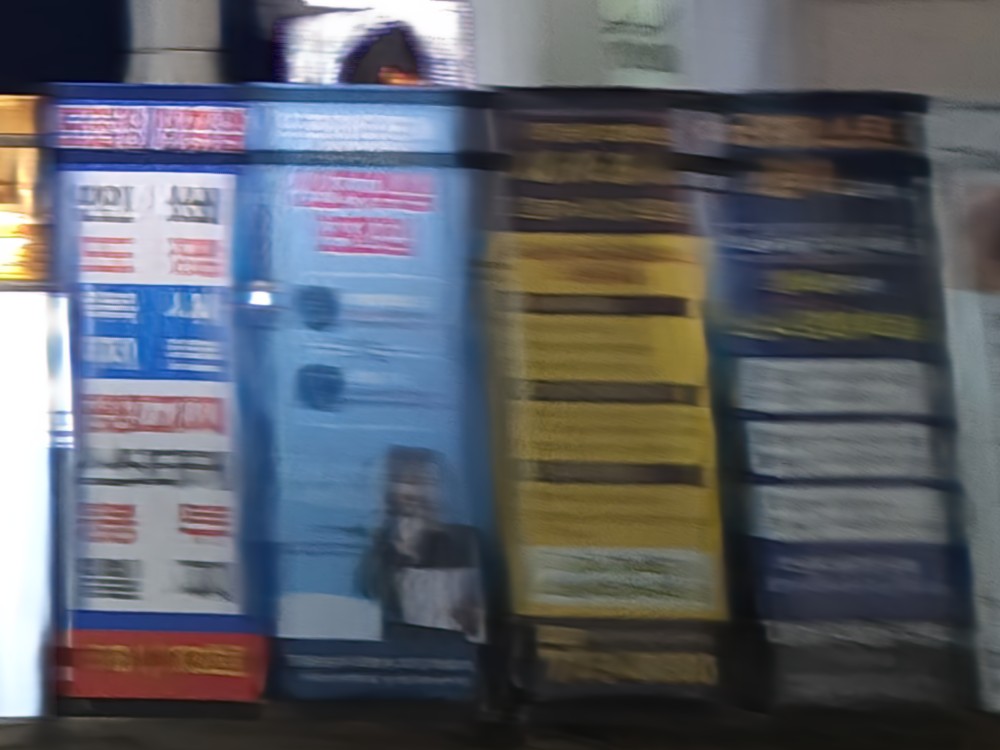}}
  \centerline{(e) FFTFormer-16~\cite{fftformer}}\medskip
\end{minipage}
\begin{minipage}[!ht]{.51\linewidth}
  \centering
  \centerline{\includegraphics[width=8.6cm]{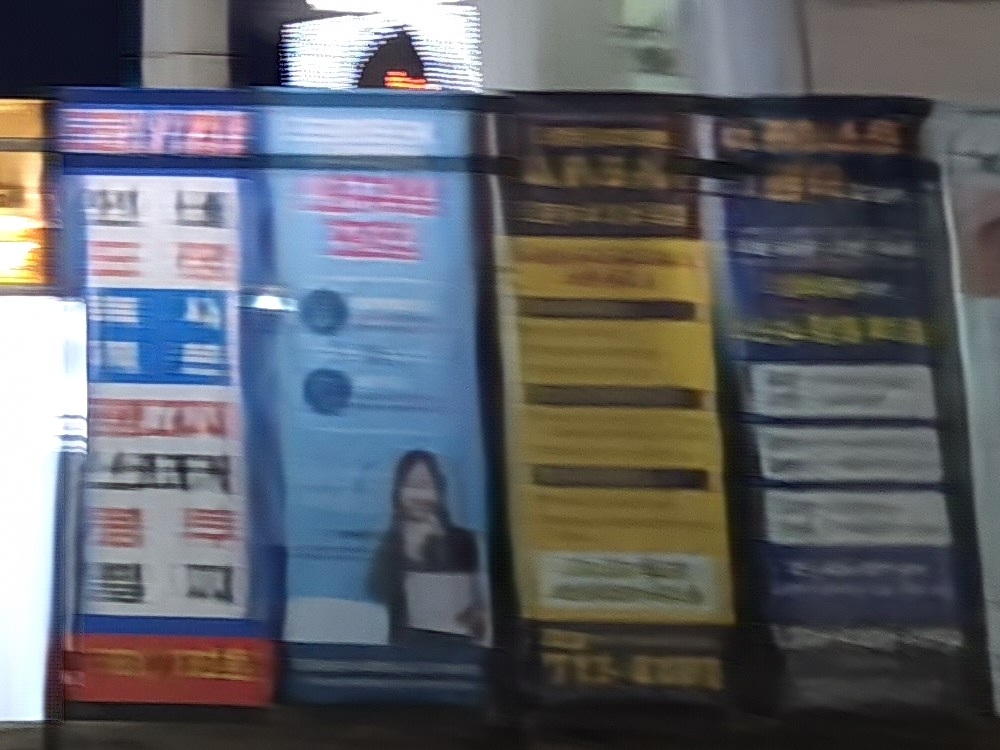}}
  \centerline{(b) Samsung EnhanceX}\medskip
  \centerline{\includegraphics[width=8.6cm]{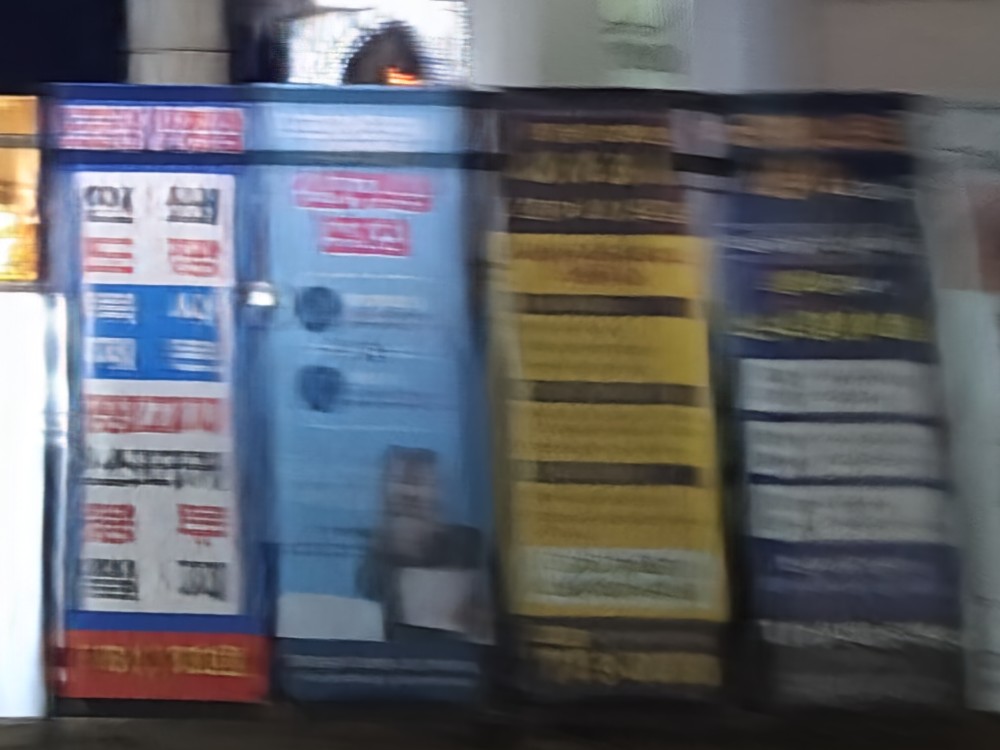}}
  \centerline{(d) NAFNet-32~\cite{nafnet}}\medskip
  \centerline{\includegraphics[width=8.6cm]{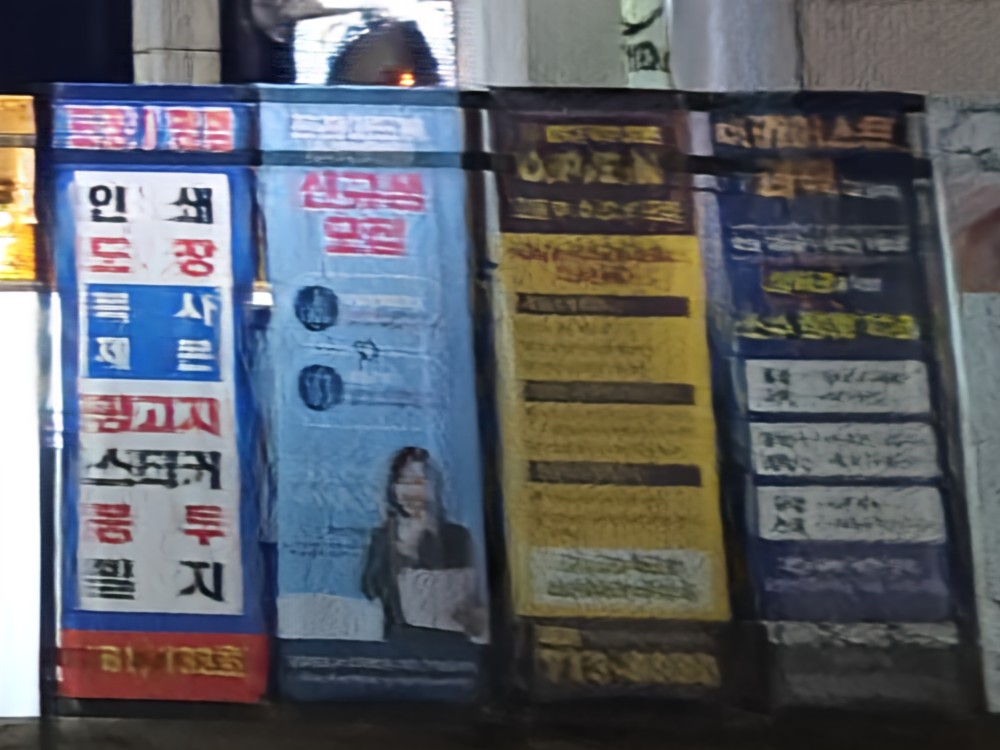}}
  \centerline{(f) SegDeblur-S (ours)}\medskip
\end{minipage}
\caption{Visual comparison results on real-world blur images.}
\label{fig:supp_realworld3}
\end{figure*}

{We present more real-world examples to compare our method with the recent works and commercial applications. We capture real-world blur images with natural hand motions by Samsung Galaxy Note S20 Ultra. As shown in Fig.~\ref{fig:supp_realworld1}, \ref{fig:supp_realworld2} and \ref{fig:supp_realworld3}, the real-world blur images are well-reconstructed by our method. Meanwhile, the recent works and deblurring applications work well on some blur images but some other blur images are not well-recovered.}


\clearpage
\section{Visual results on blur segmentation map}\label{sec:app_blursegmap}

\begin{figure*}[!ht]
  \centering
\begin{minipage}[!ht]{.240\linewidth}
  \centering
  \centerline{\includegraphics[width=4.2cm]{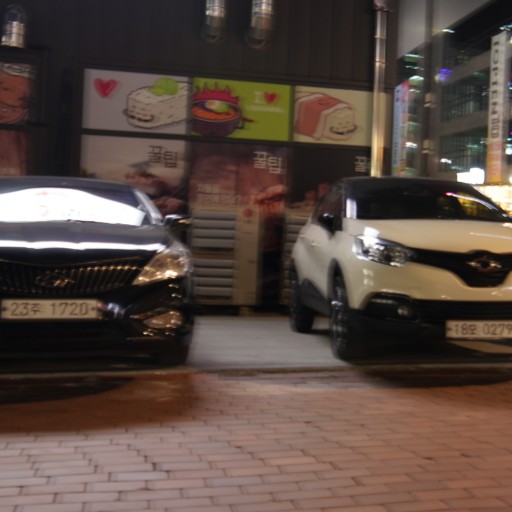}}
  \centerline{\includegraphics[width=4.2cm]{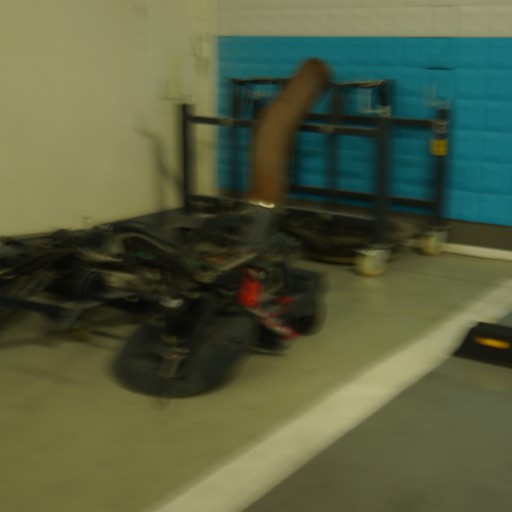}}
  \centerline{\includegraphics[width=4.2cm]{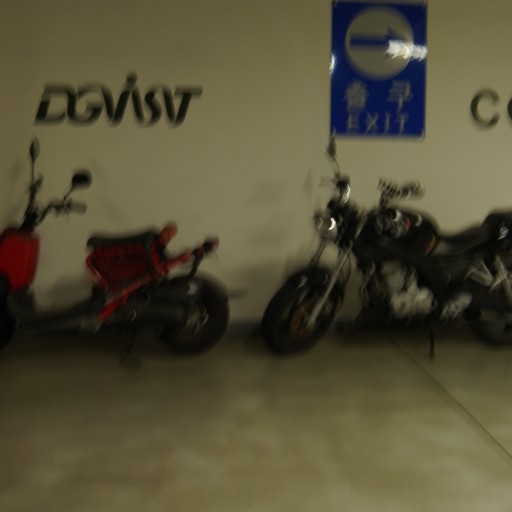}}
  \centerline{\includegraphics[width=4.2cm]{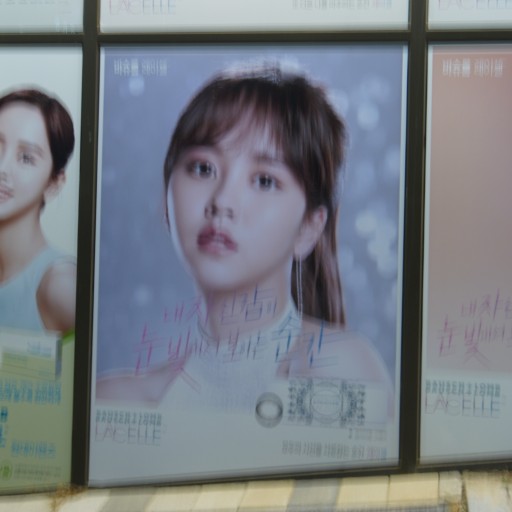}}
  \centerline{(a) Blur Image}\medskip
\end{minipage}
\begin{minipage}[!ht]{.240\linewidth}
  \centering
  \centerline{\includegraphics[width=4.2cm]{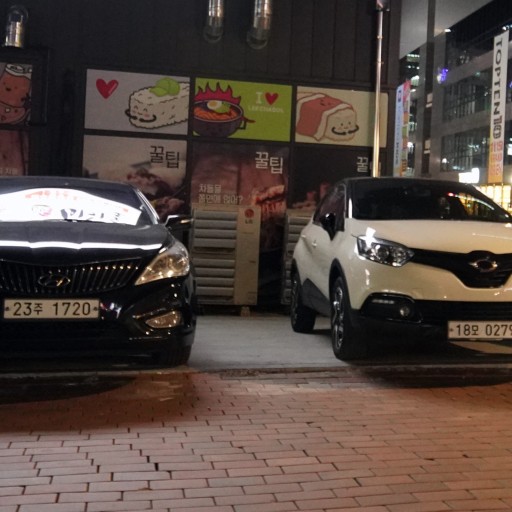}}
  \centerline{\includegraphics[width=4.2cm]{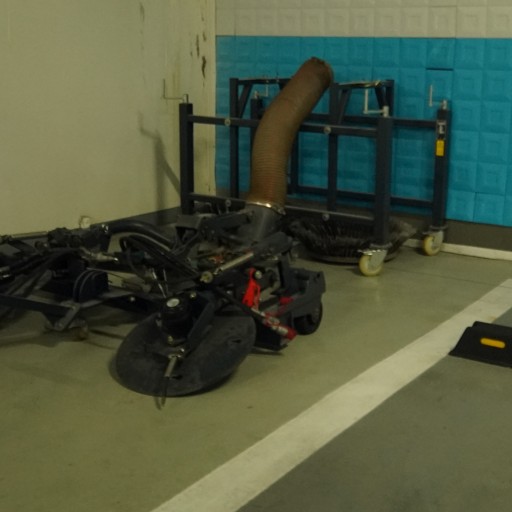}}
  \centerline{\includegraphics[width=4.2cm]{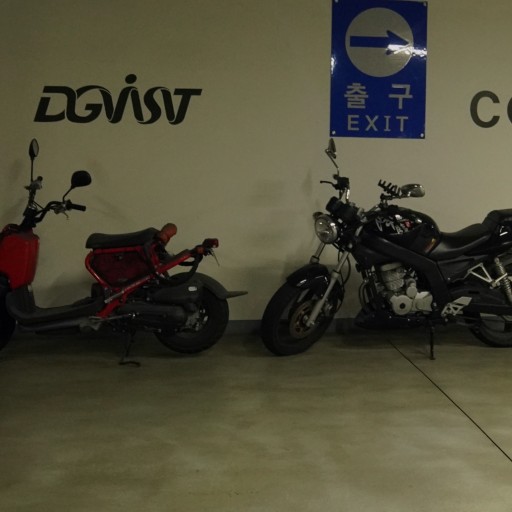}}
  \centerline{\includegraphics[width=4.2cm]{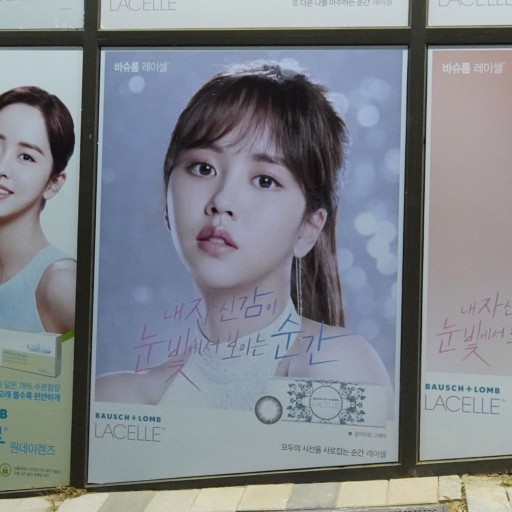}}
  \centerline{(b) Sharp Image}\medskip
\end{minipage}
\begin{minipage}[!ht]{.240\linewidth}
  \centering
  \centerline{\includegraphics[width=4.2cm]{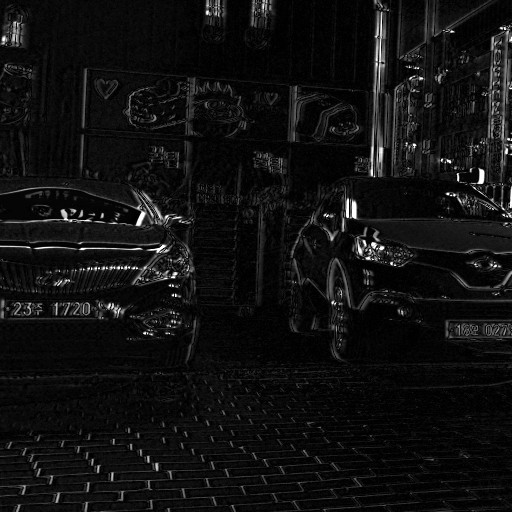}}
  \centerline{\includegraphics[width=4.2cm]{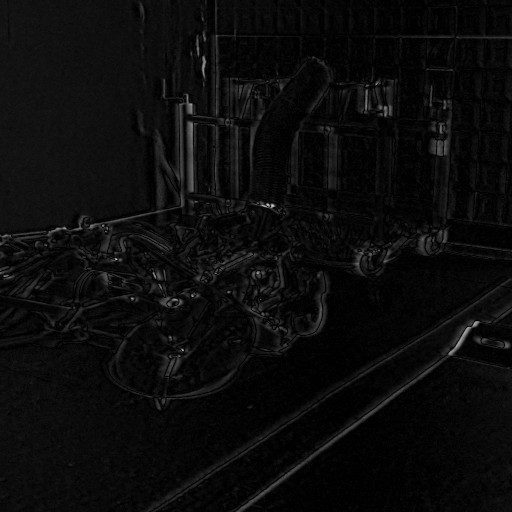}}
  \centerline{\includegraphics[width=4.2cm]{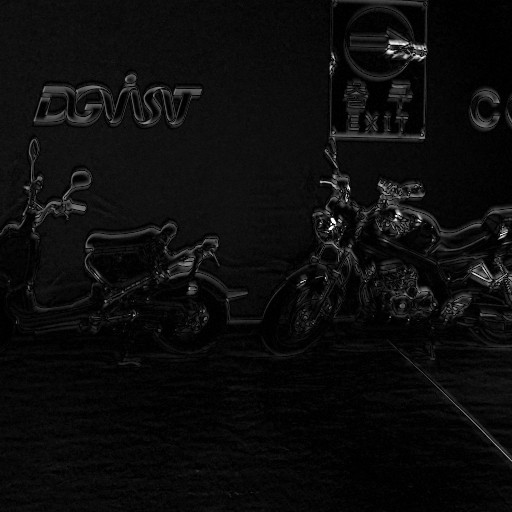}}
  \centerline{\includegraphics[width=4.2cm]{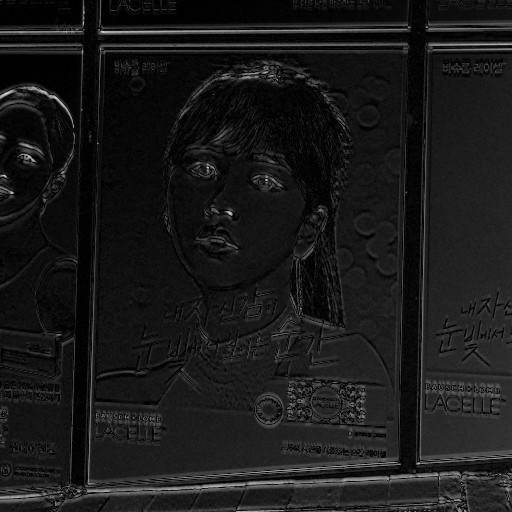}}
  \centerline{(c) Image Residual Error}\medskip
\end{minipage}
\begin{minipage}[!ht]{.240\linewidth}
  \centering
  \centerline{\includegraphics[width=4.2cm]{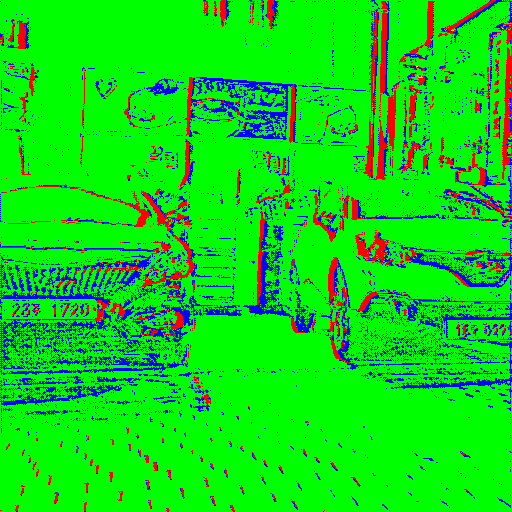}}
  \centerline{\includegraphics[width=4.2cm]{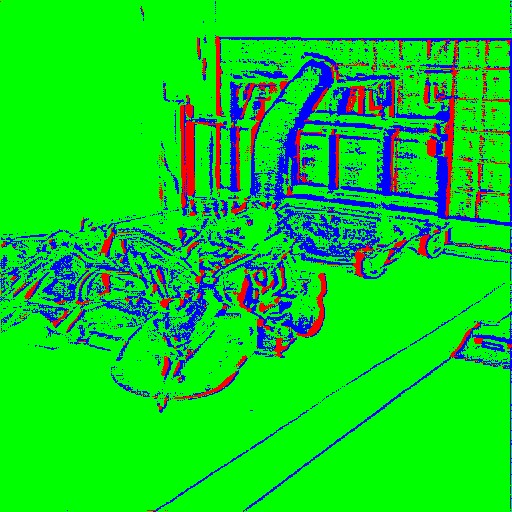}}
  \centerline{\includegraphics[width=4.2cm]{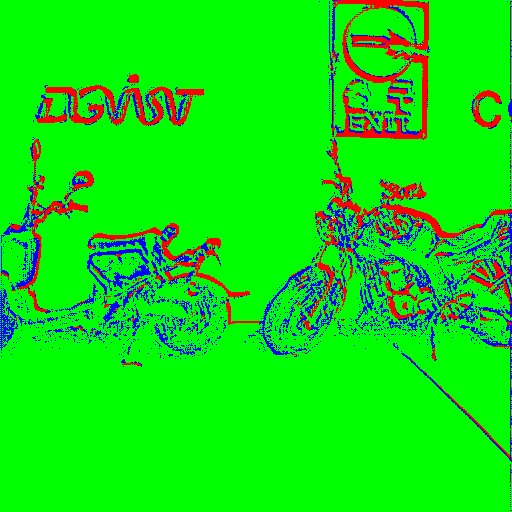}}
  \centerline{\includegraphics[width=4.2cm]{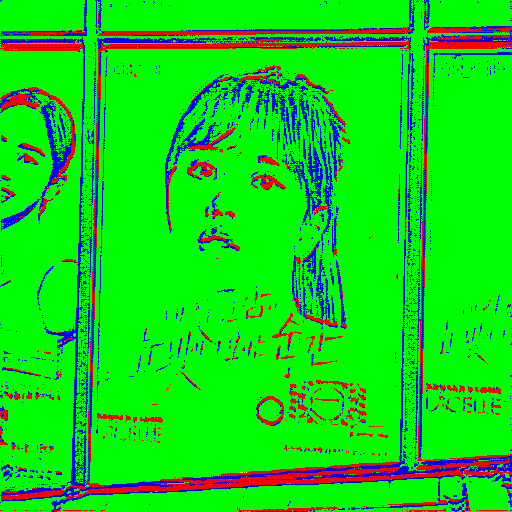}}
  \centerline{(d) Blur Segmentation Map}\medskip
\end{minipage}
\caption{Blur segmentation map results on camera motion examples for RealBlur~\cite{realblur}.}
\label{fig:supp_blurmap1}
\vspace{-0.3cm}
\end{figure*}

\begin{figure*}[!ht]
  \centering
\begin{minipage}[!ht]{.240\linewidth}
  \centering
  \centerline{\includegraphics[width=4.2cm]{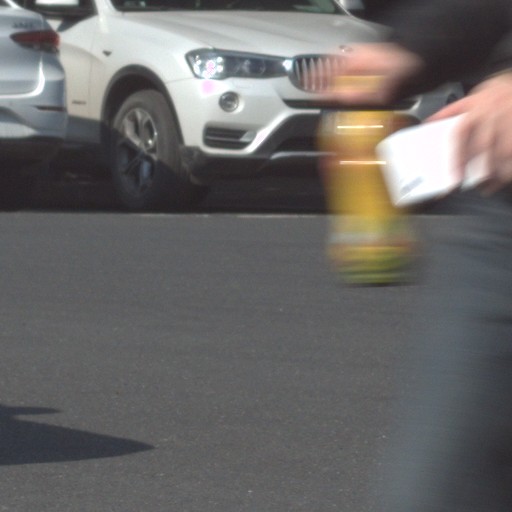}}
  \centerline{\includegraphics[width=4.2cm]{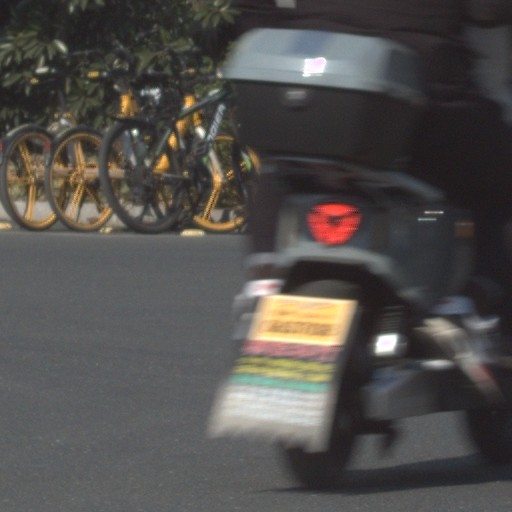}}
  \centerline{\includegraphics[width=4.2cm]{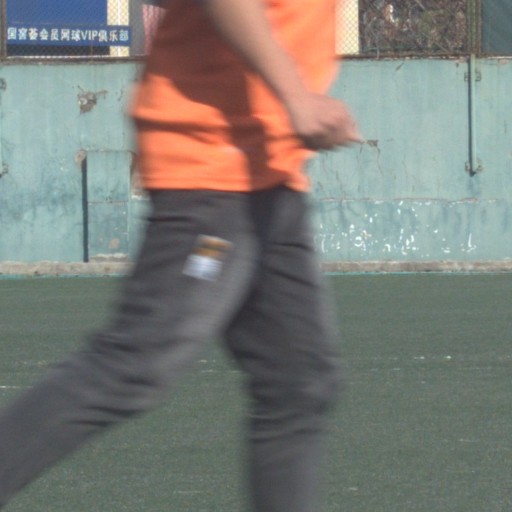}}
  \centerline{\includegraphics[width=4.2cm]{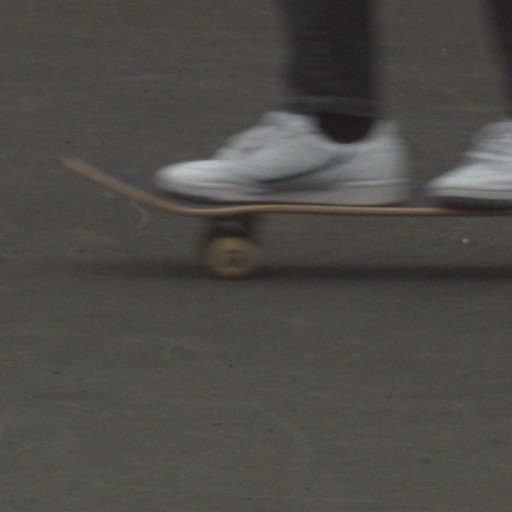}}
  \centerline{(a) Blur Image}\medskip
\end{minipage}
\begin{minipage}[!ht]{.240\linewidth}
  \centering
  \centerline{\includegraphics[width=4.2cm]{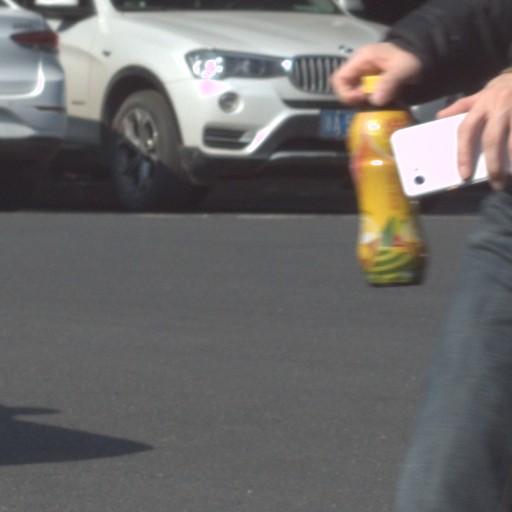}}
  \centerline{\includegraphics[width=4.2cm]{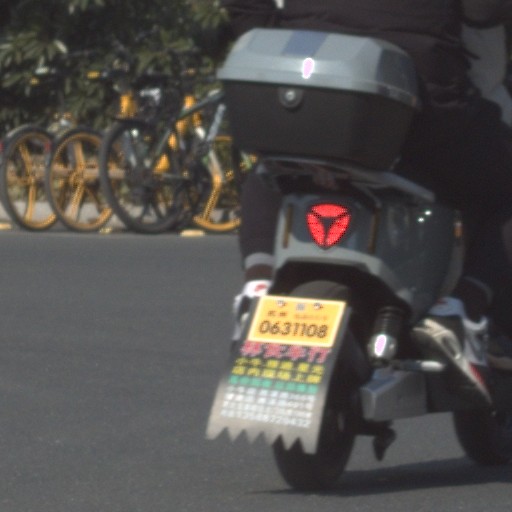}}
  \centerline{\includegraphics[width=4.2cm]{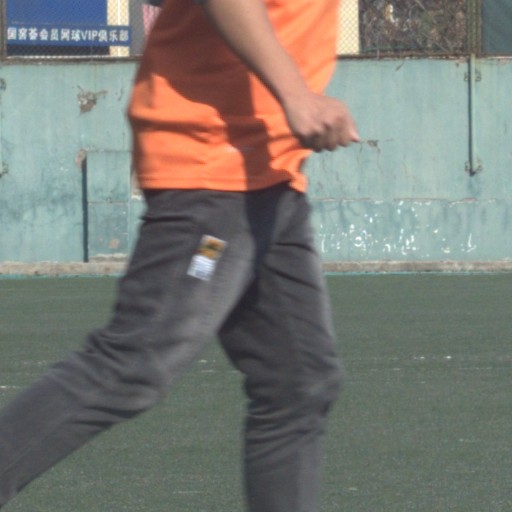}}
  \centerline{\includegraphics[width=4.2cm]{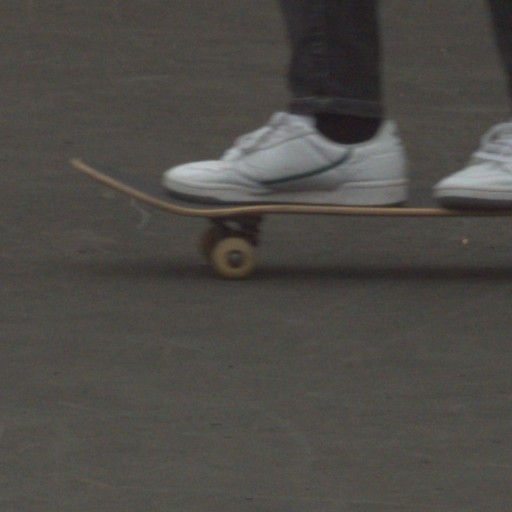}}
  \centerline{(b) Sharp Image}\medskip
\end{minipage}
\begin{minipage}[!ht]{.240\linewidth}
  \centering
  \centerline{\includegraphics[width=4.2cm]{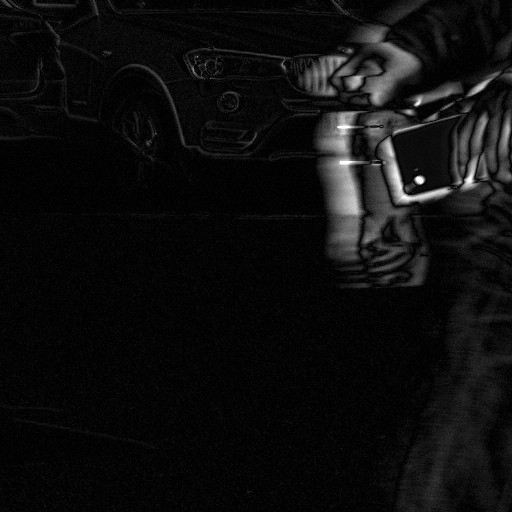}}
  \centerline{\includegraphics[width=4.2cm]{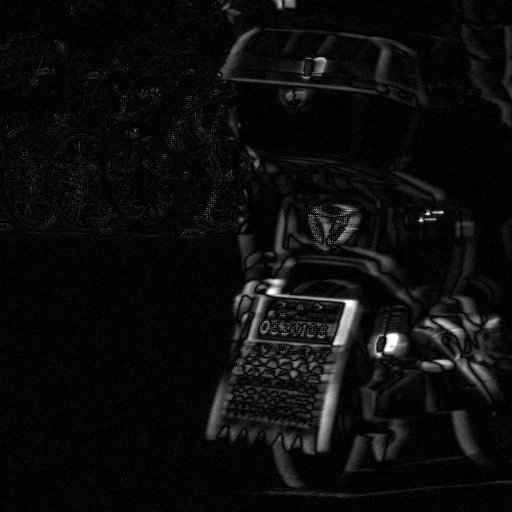}}
  \centerline{\includegraphics[width=4.2cm]{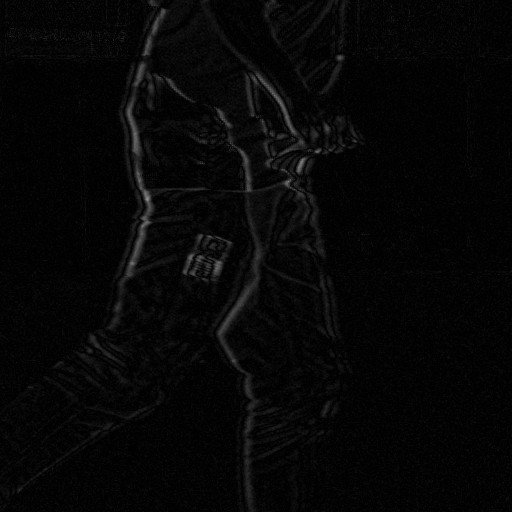}}
  \centerline{\includegraphics[width=4.2cm]{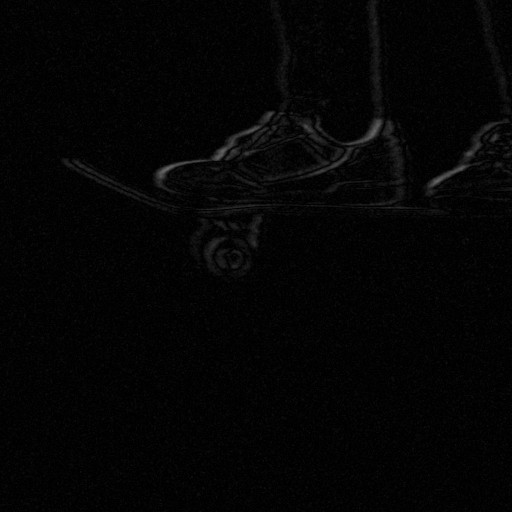}}
  \centerline{(c) Image Residual Error}\medskip
\end{minipage}
\begin{minipage}[!ht]{.240\linewidth}
  \centering
  \centerline{\includegraphics[width=4.2cm]{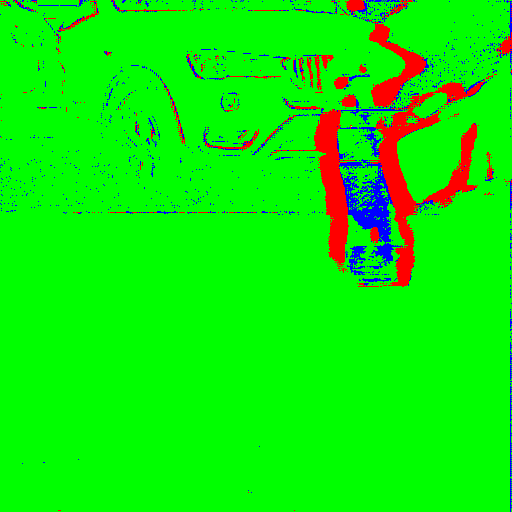}}
  \centerline{\includegraphics[width=4.2cm]{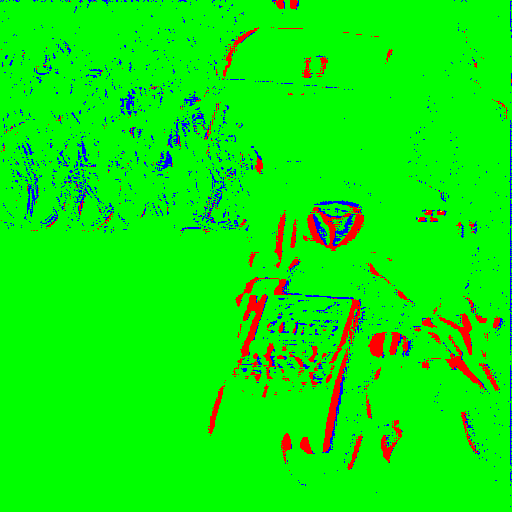}}
  \centerline{\includegraphics[width=4.2cm]{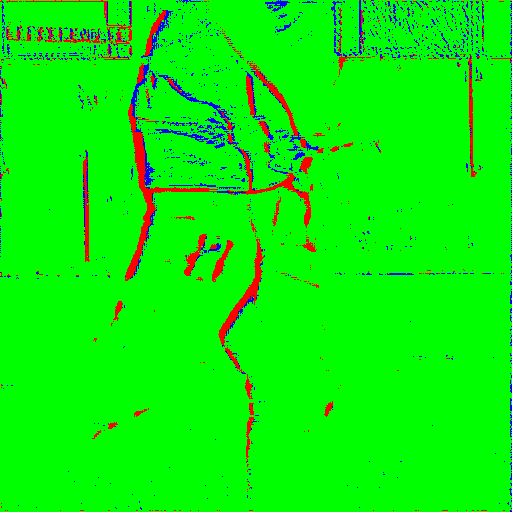}}
  \centerline{\includegraphics[width=4.2cm]{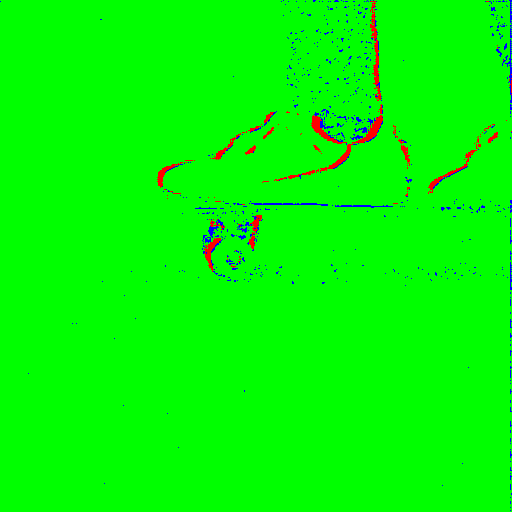}}
  \centerline{(d) Blur Segmentation Map}\medskip
\end{minipage}
\caption{Blur segmentation map results on object motion examples for ReLoBlur~\cite{reloblur}.}
\label{fig:supp_blurmap2}
\vspace{-0.3cm}
\end{figure*}

\clearpage
\section{Qualitative results for large deblurring models}\label{sec:app_qualitative_large}
\vspace{-0.3cm}
\begin{figure*}[!ht]
  \centering
\begin{minipage}[!ht]{.48\linewidth}
  \centering
  \centerline{\includegraphics[width=7.5cm]{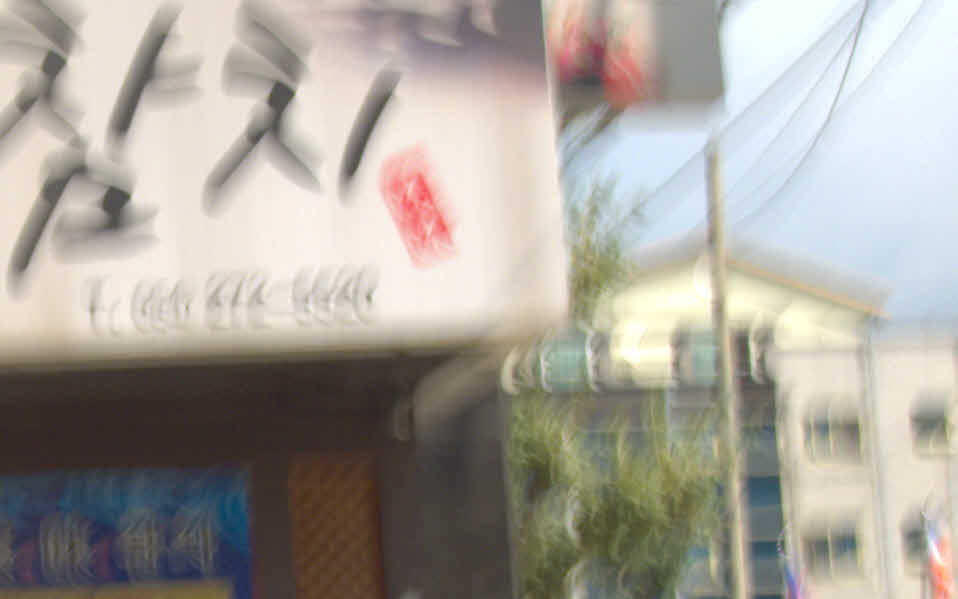}}
  \centerline{(a) Blur Input}\medskip
  \centerline{\includegraphics[width=7.5cm]{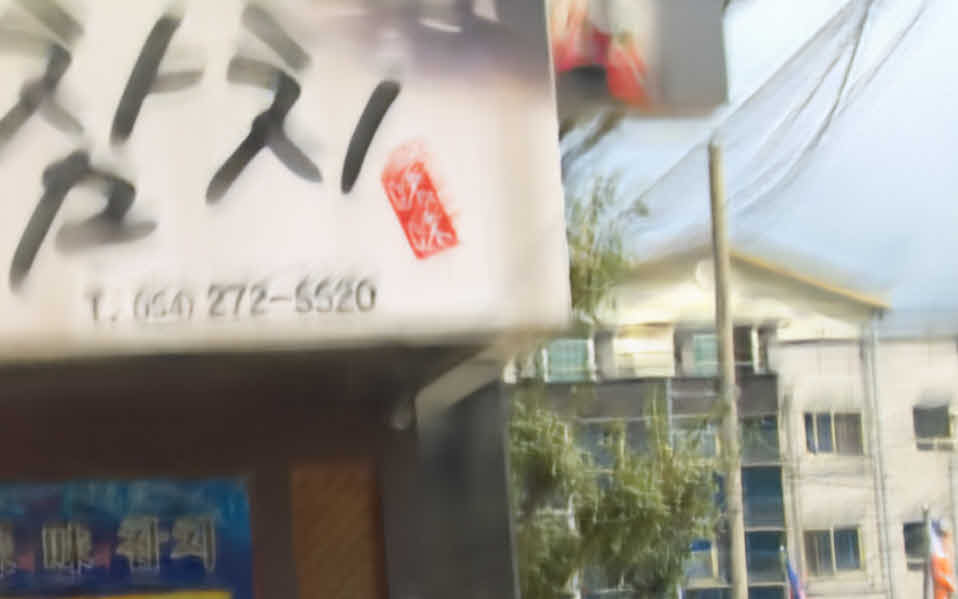}}
  \centerline{(c) UFPNet~\cite{ufp}}\medskip
\end{minipage}
\begin{minipage}[!ht]{.48\linewidth}
  \centering
  \centerline{\includegraphics[width=7.5cm]{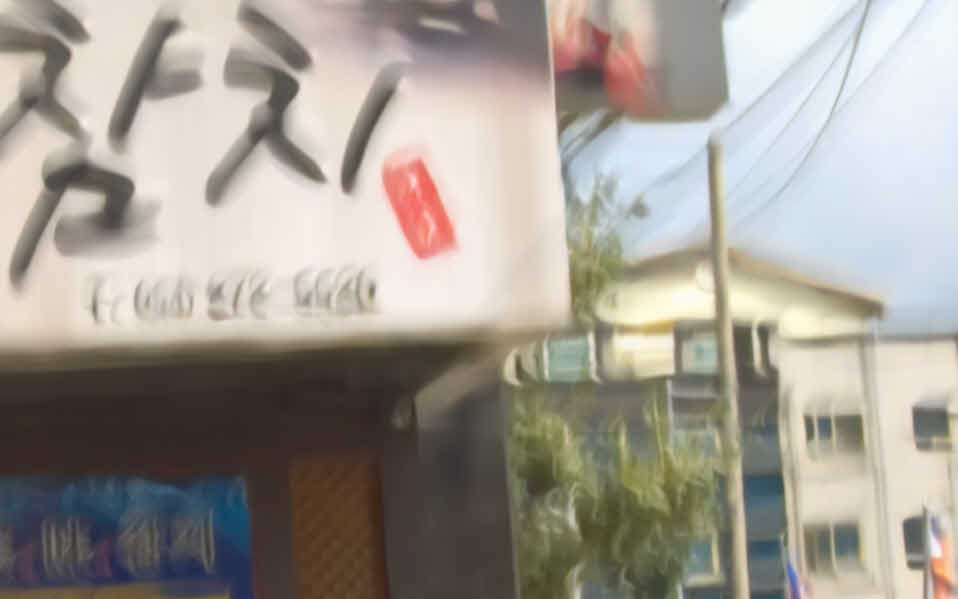}}
  \centerline{(b) MAXIM-3S~\cite{maxim}}\medskip
  \centerline{\includegraphics[width=7.5cm]{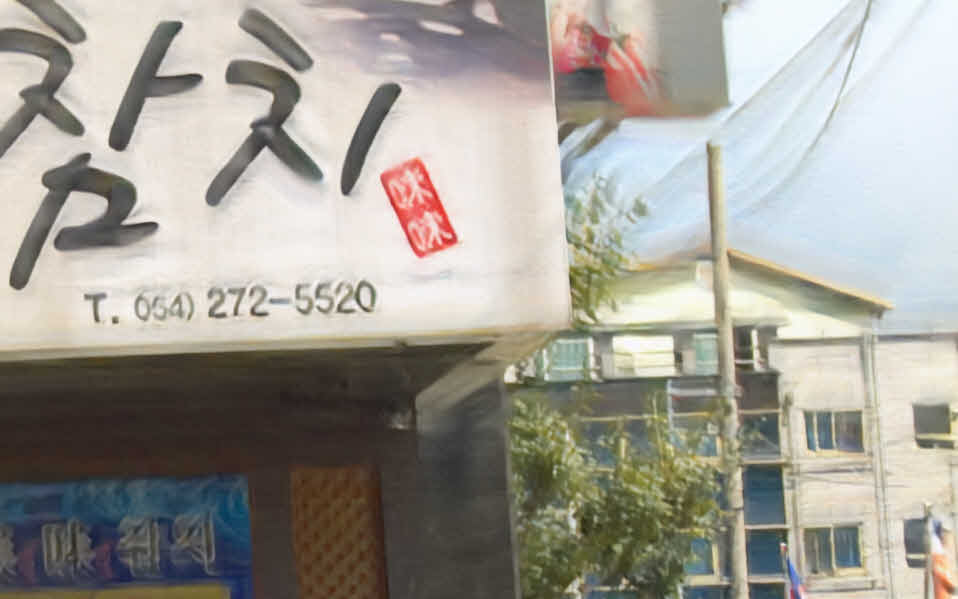}}
  \centerline{(d) SegDeblur-L (ours)}\medskip
\end{minipage}
\vspace{-0.3cm}
\end{figure*}

\begin{figure*}[!ht]
  \centering
\begin{minipage}[!ht]{.48\linewidth}
  \centering
  \centerline{\includegraphics[width=7.5cm]{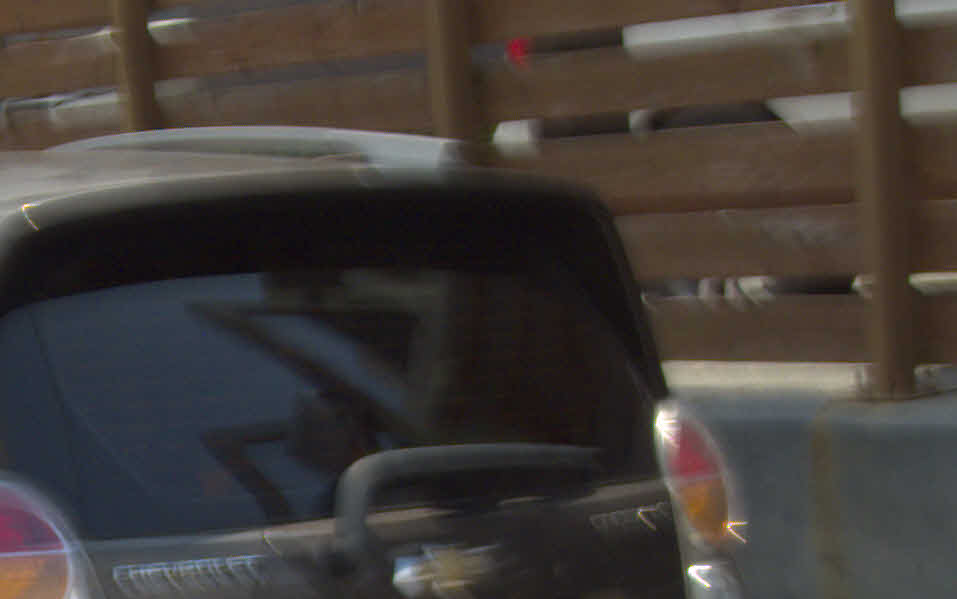}}
  \centerline{(a) Blur Input}\medskip
  \centerline{\includegraphics[width=7.5cm]{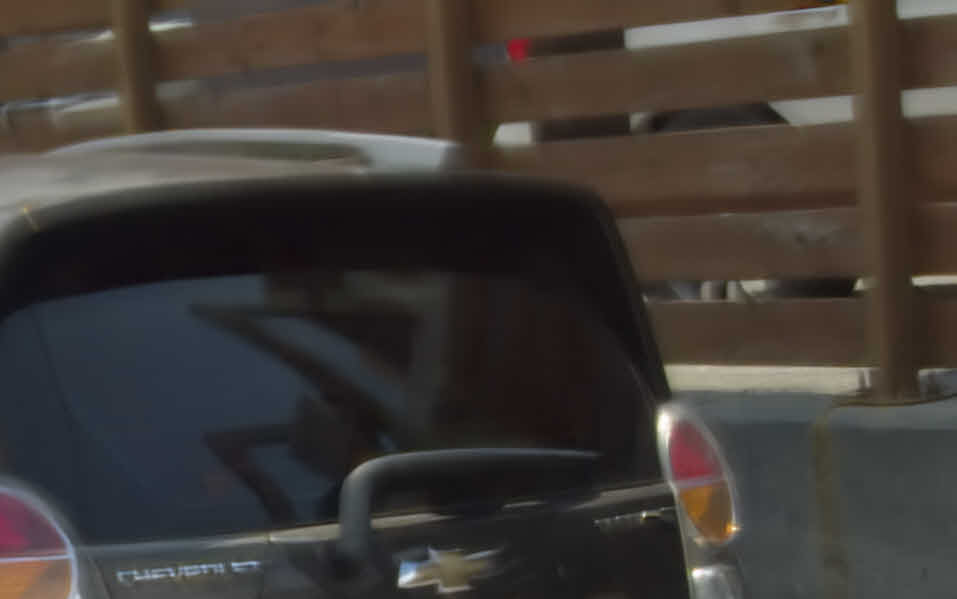}}
  \centerline{(c) UFPNet~\cite{ufp}}\medskip
\end{minipage}
\begin{minipage}[!ht]{.48\linewidth}
  \centering
  \centerline{\includegraphics[width=7.5cm]{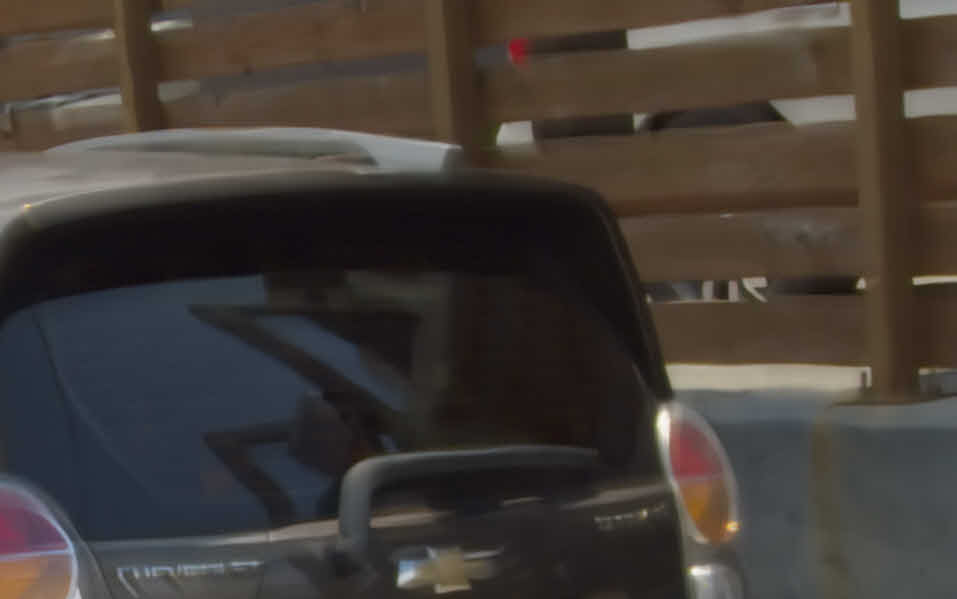}}
  \centerline{(b) MAXIM-3S~\cite{maxim}}\medskip
  \centerline{\includegraphics[width=7.5cm]{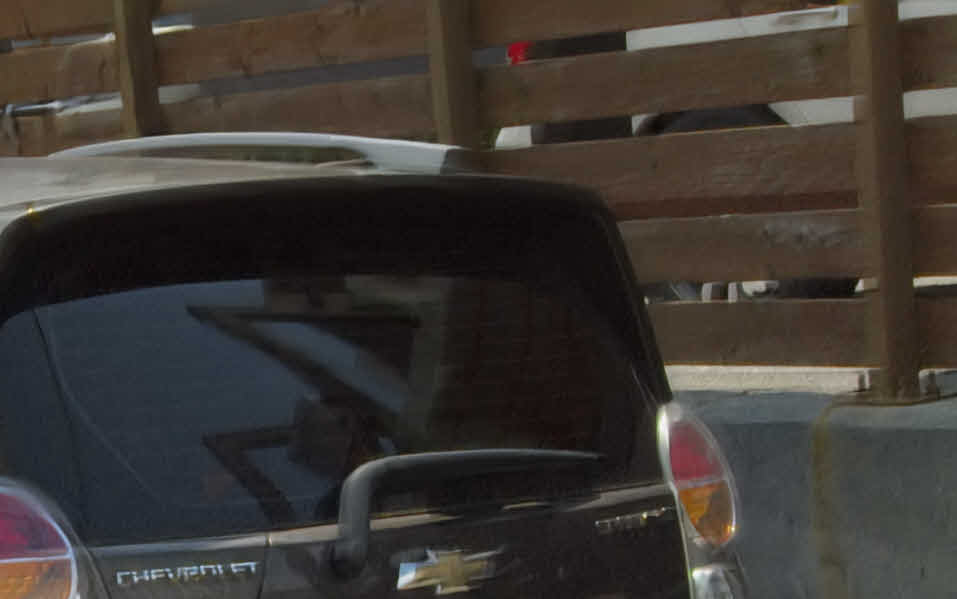}}
  \centerline{(d) SegDeblur-L (ours)}\medskip
\end{minipage}
\vspace{-0.5cm}
\end{figure*}

\vspace{-0.5cm}
\begin{figure*}[!ht]
  \centering
\begin{minipage}[!ht]{.48\linewidth}
  \centering
  \centerline{\includegraphics[width=7.5cm]{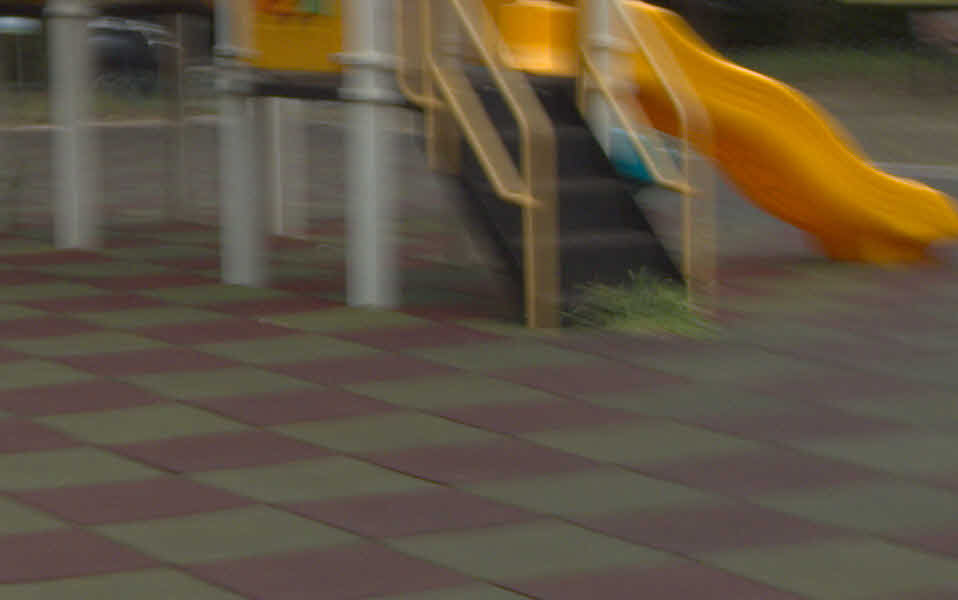}}
  \centerline{(a) Blur Input}\medskip
  \centerline{\includegraphics[width=7.5cm]{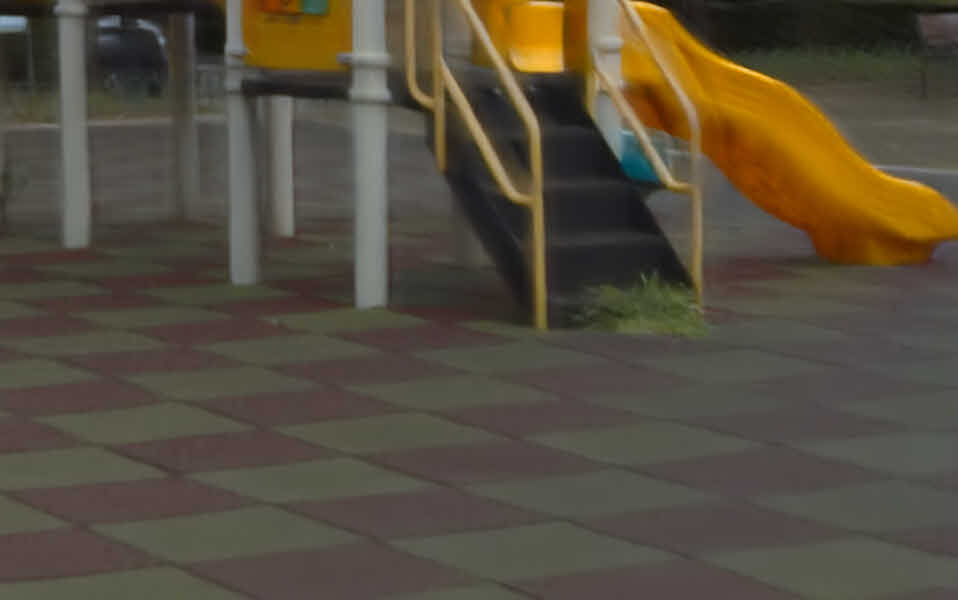}}
  \centerline{(c) UFPNet~\cite{ufp}}\medskip
\end{minipage}
\begin{minipage}[!ht]{.48\linewidth}
  \centering
  \centerline{\includegraphics[width=7.5cm]{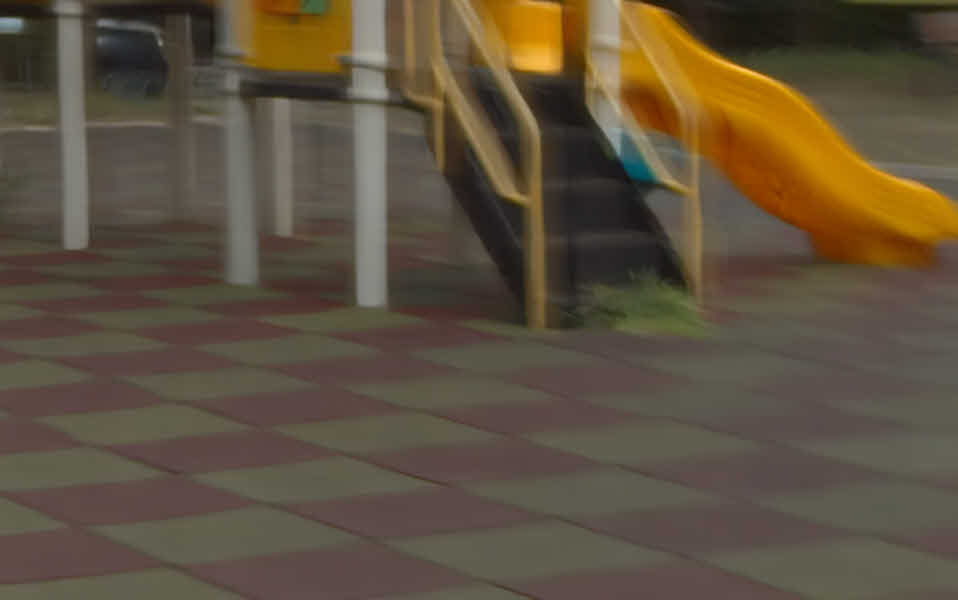}}
  \centerline{(b) MAXIM-3S~\cite{maxim}}\medskip
  \centerline{\includegraphics[width=7.5cm]{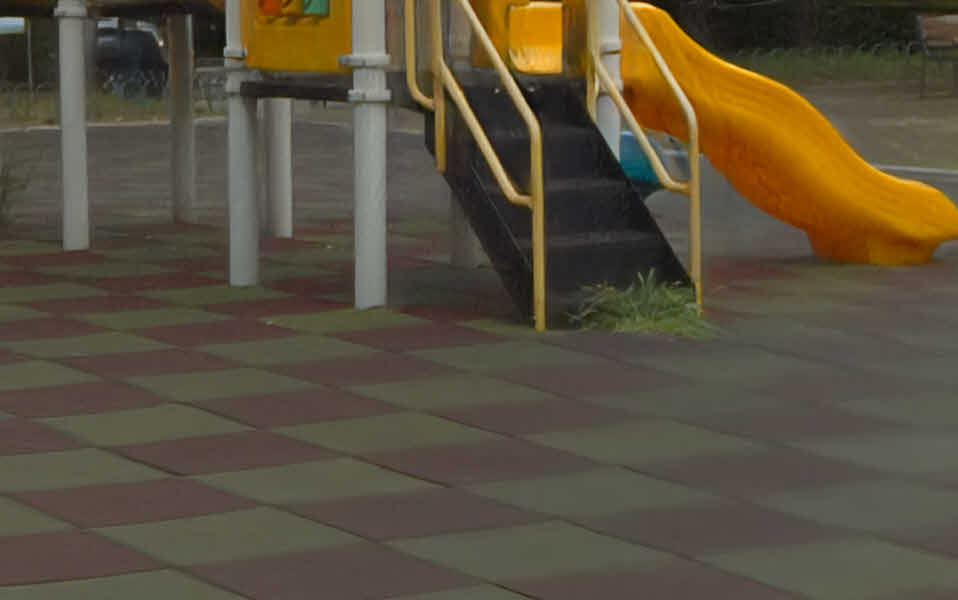}}
  \centerline{(d) SegDeblur-L (ours)}\medskip
\end{minipage}
\end{figure*}

\vspace{-0.5cm}
\begin{figure*}[!ht]
  \centering
\begin{minipage}[!ht]{.48\linewidth}
  \centering
  \centerline{\includegraphics[width=7.5cm]{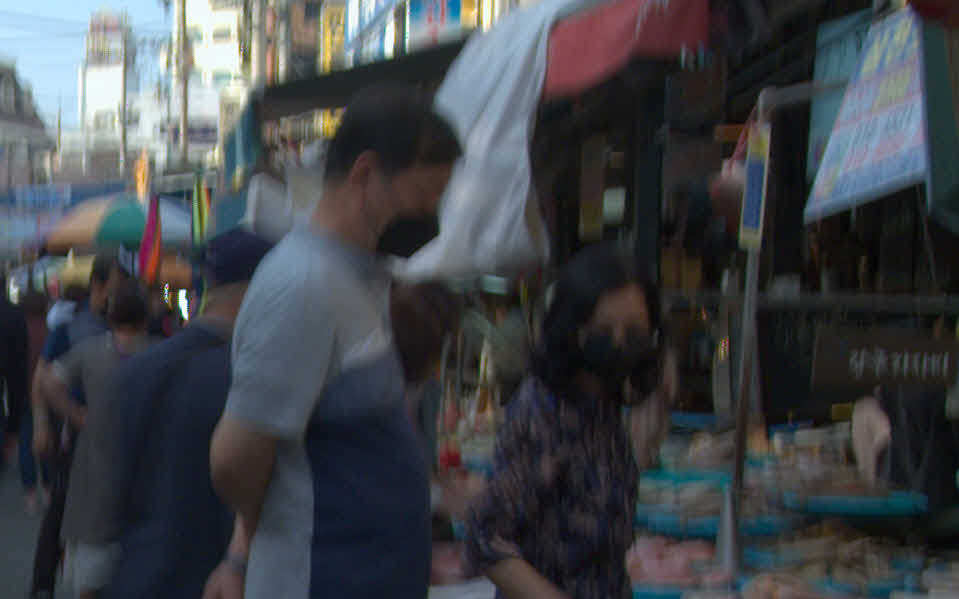}}
  \centerline{(a) Blur Input}\medskip
  \centerline{\includegraphics[width=7.5cm]{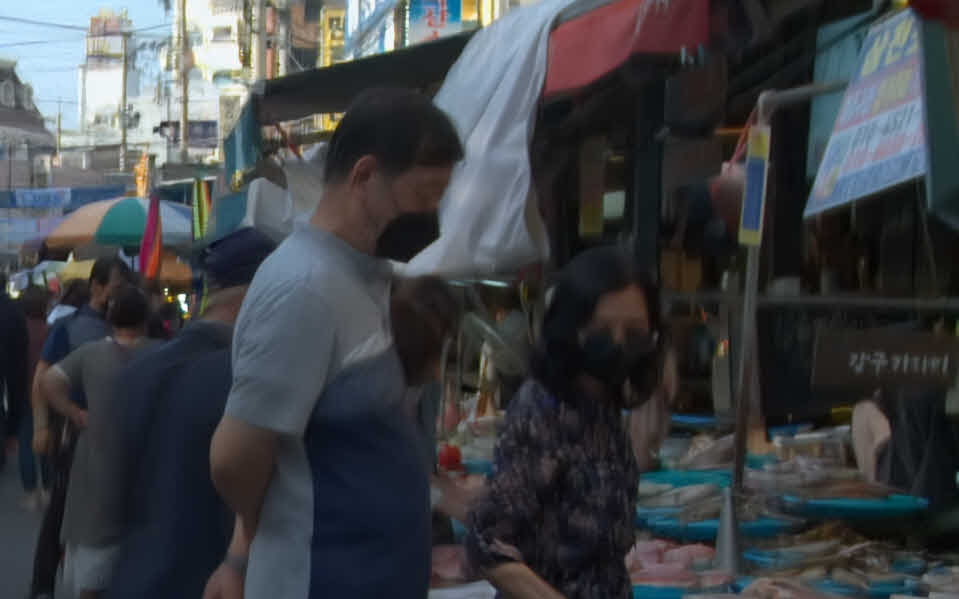}}
  \centerline{(c) UFPNet~\cite{ufp}}\medskip
\end{minipage}
\begin{minipage}[!ht]{.48\linewidth}
  \centering
  \centerline{\includegraphics[width=7.5cm]{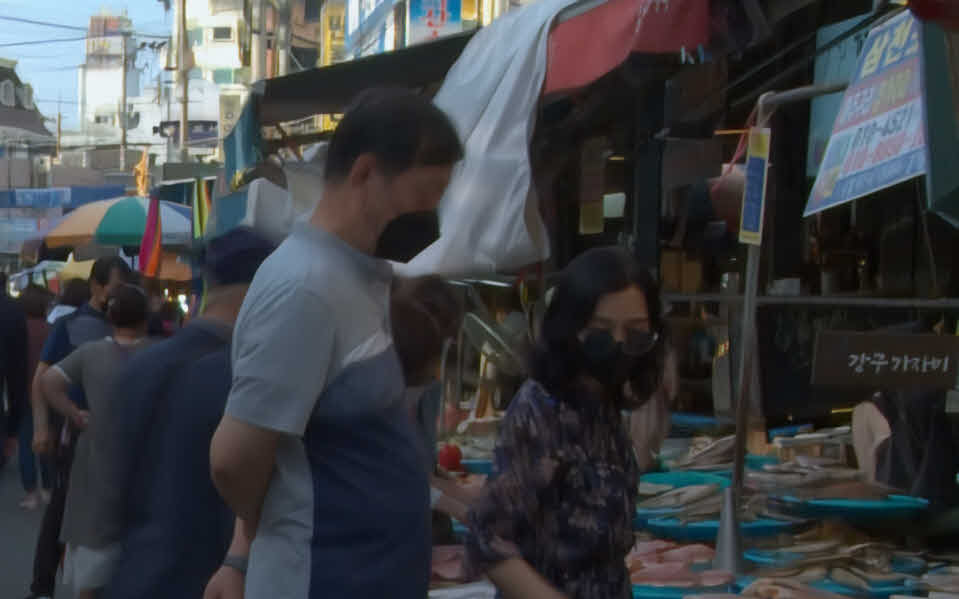}}
  \centerline{(b) MAXIM-3S~\cite{maxim}}\medskip
  \centerline{\includegraphics[width=7.5cm]{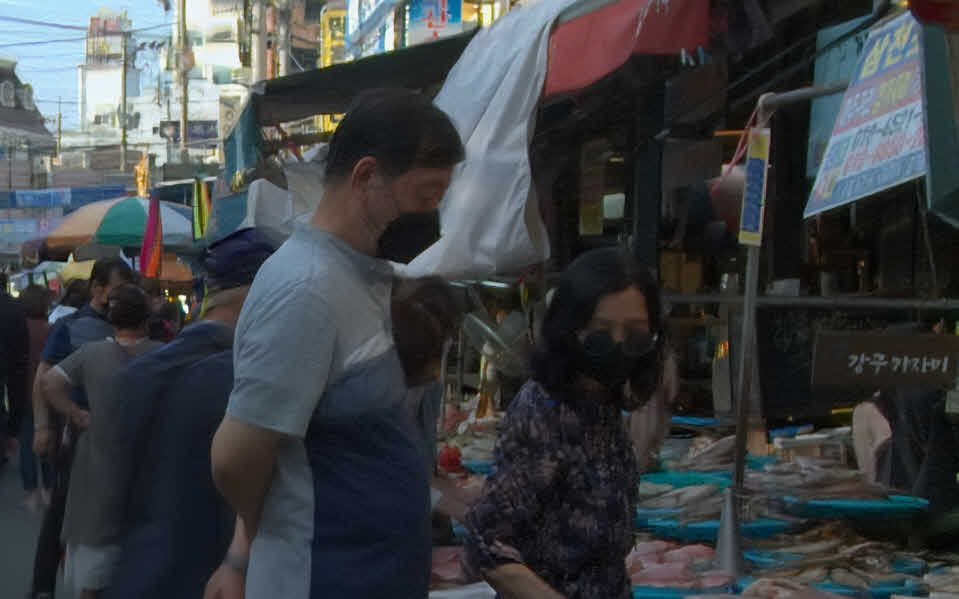}}
  \centerline{(d) SegDeblur-L (ours)}\medskip
\end{minipage}
\vspace{-0.2cm}
\caption{Qualitative results for large deblurring models in RSBlur~\cite{rsblur}.}
\end{figure*}

\clearpage
\section{Qualitative results for efficient deblurring models}\label{sec:app_qualitative}
\begin{figure*}[!ht]
  \centering
\begin{minipage}[!ht]{.245\linewidth}
  \centering
  \centerline{\includegraphics[width=4.3cm]{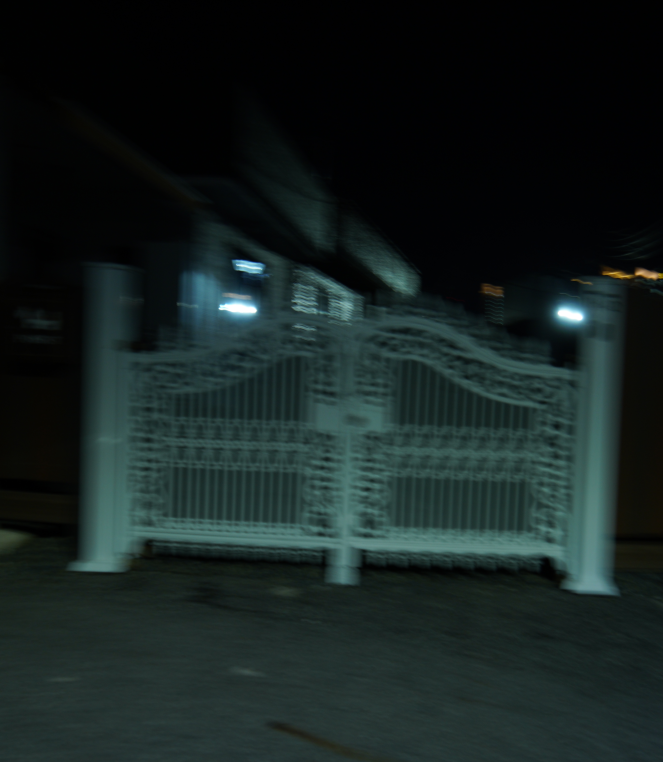}}
  \centerline{\includegraphics[width=4.3cm]{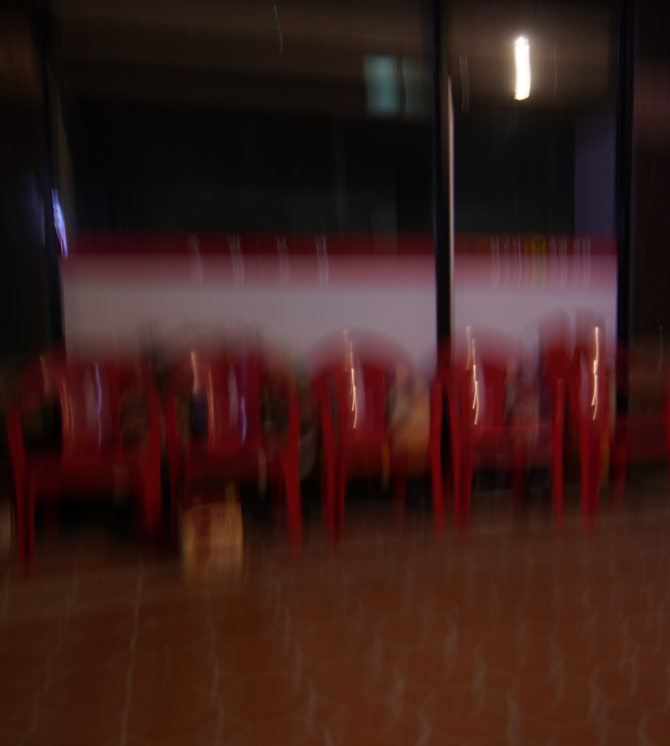}}
  \centerline{\includegraphics[width=4.3cm]{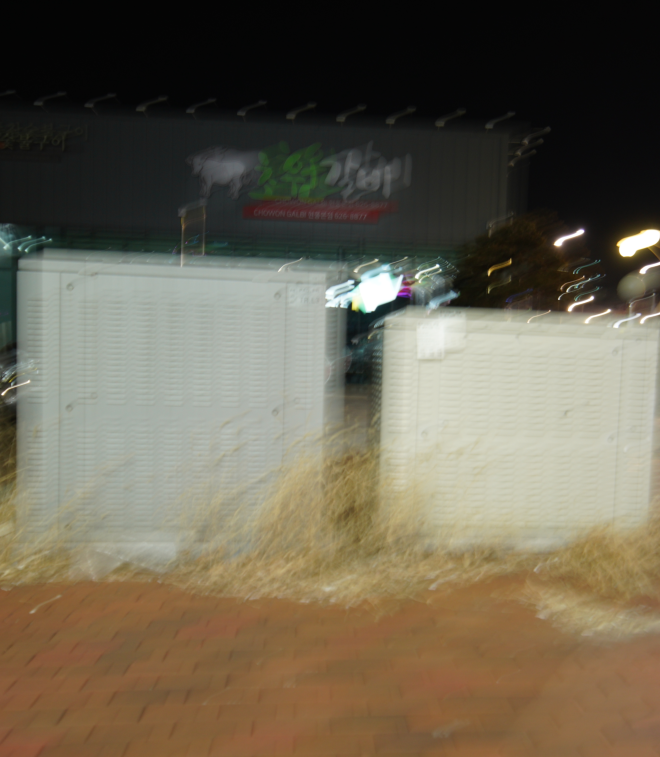}}
  \centerline{\includegraphics[width=4.3cm]{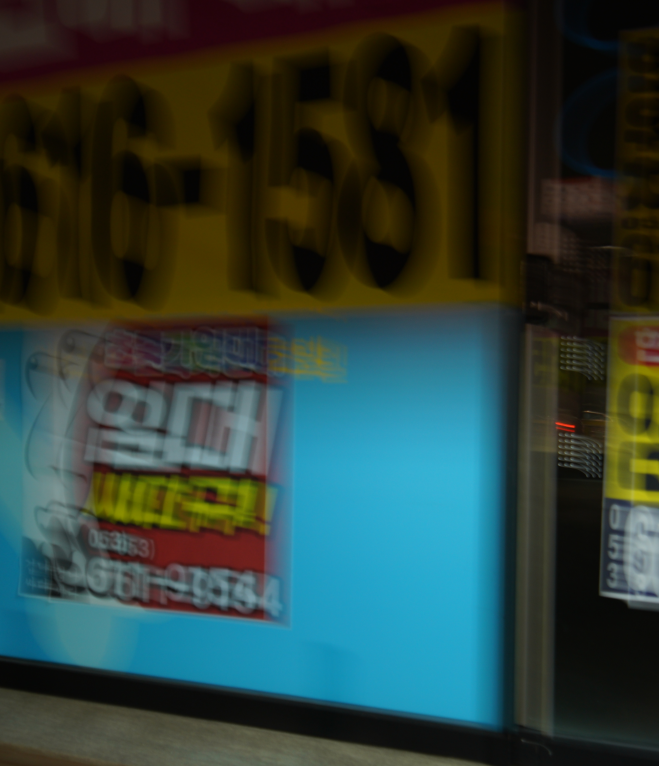}}
  \centerline{(a) Blur Input}\medskip
\end{minipage}
\begin{minipage}[!ht]{.245\linewidth}
  \centering
  \centerline{\includegraphics[width=4.3cm]{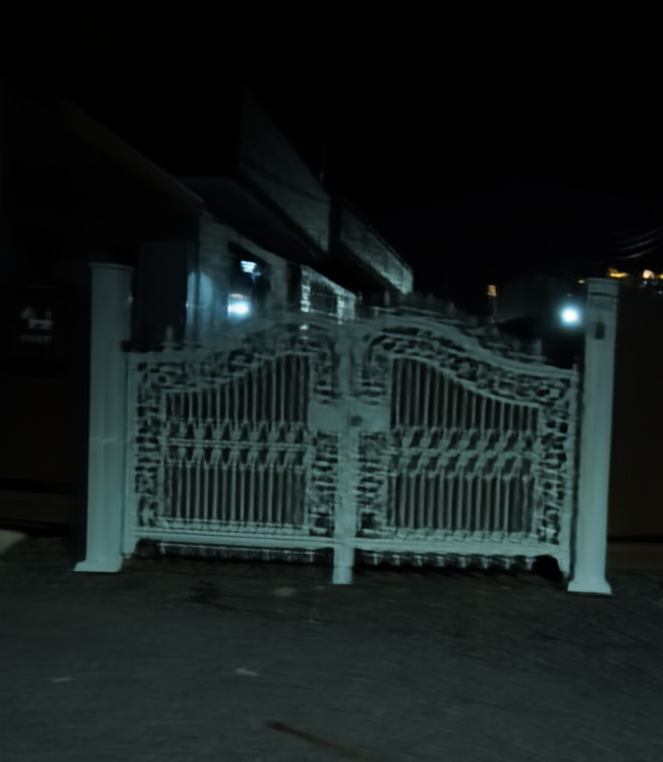}}
  \centerline{\includegraphics[width=4.3cm]{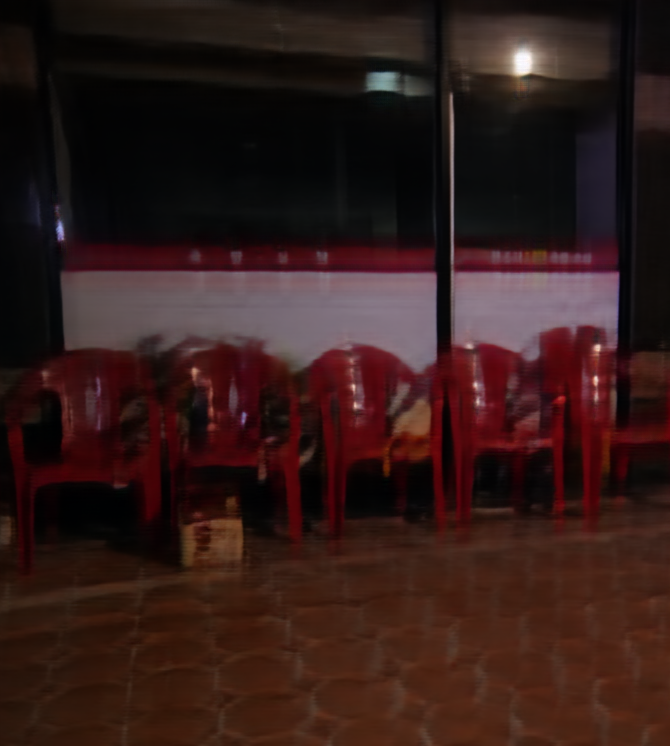}}
  \centerline{\includegraphics[width=4.3cm]{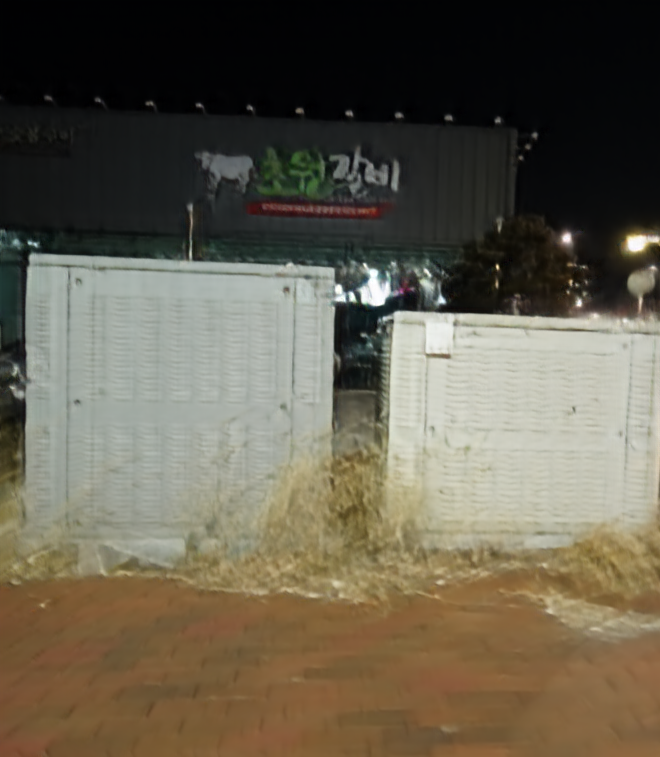}}
  \centerline{\includegraphics[width=4.3cm]{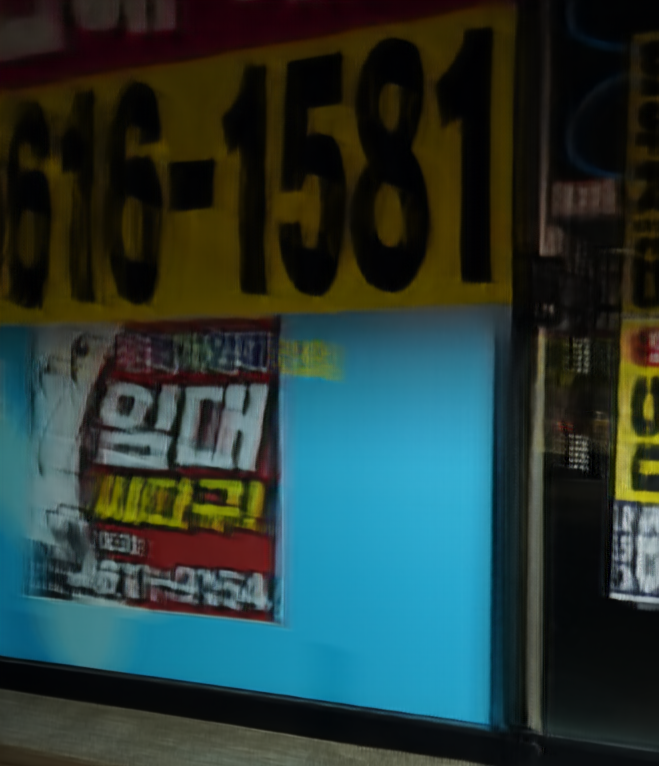}}
  \centerline{(b) NAFNet-32~\cite{nafnet}}\medskip
\end{minipage}
\begin{minipage}[!ht]{.245\linewidth}
  \centering
  \centerline{\includegraphics[width=4.3cm]{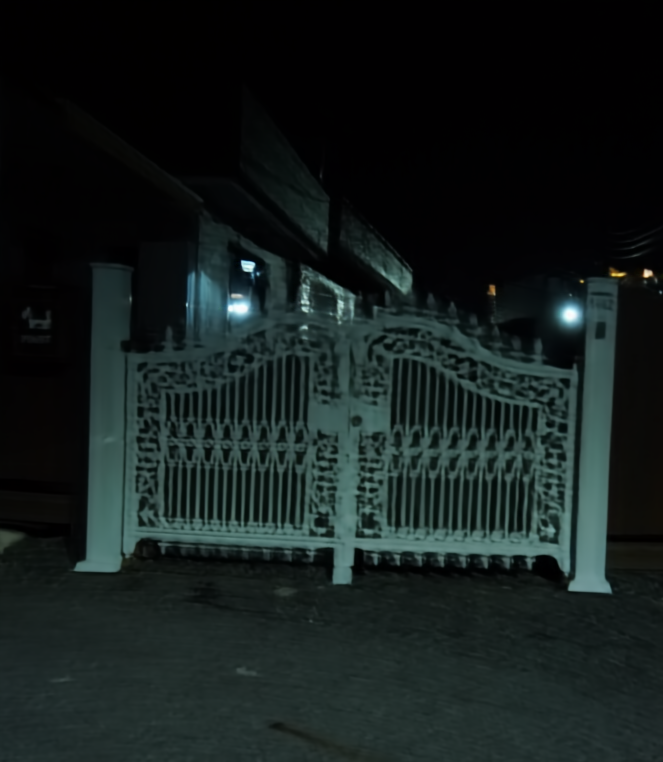}}
  \centerline{\includegraphics[width=4.3cm]{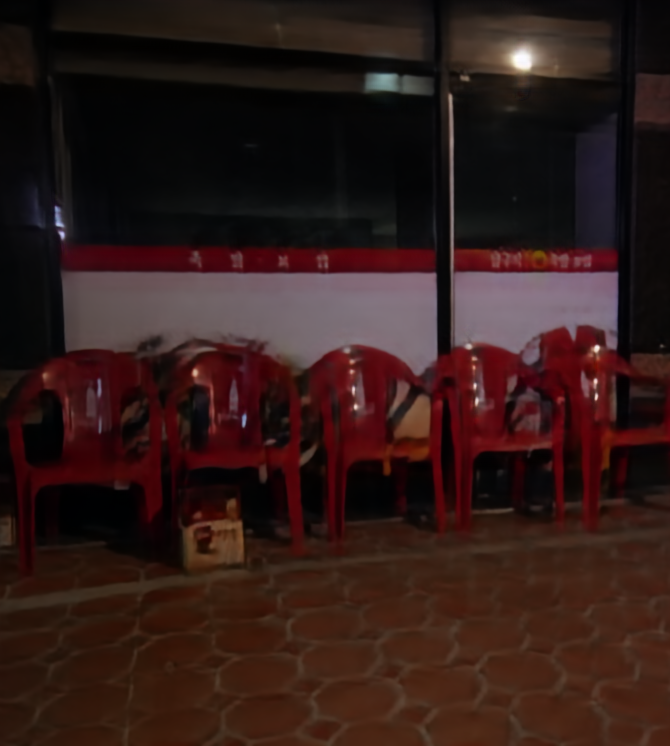}}
  \centerline{\includegraphics[width=4.3cm]{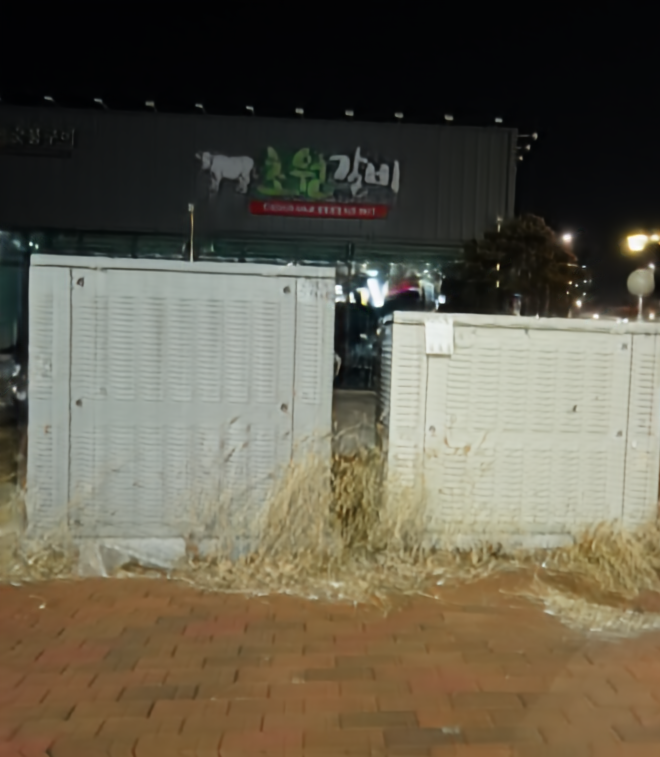}}
  \centerline{\includegraphics[width=4.3cm]{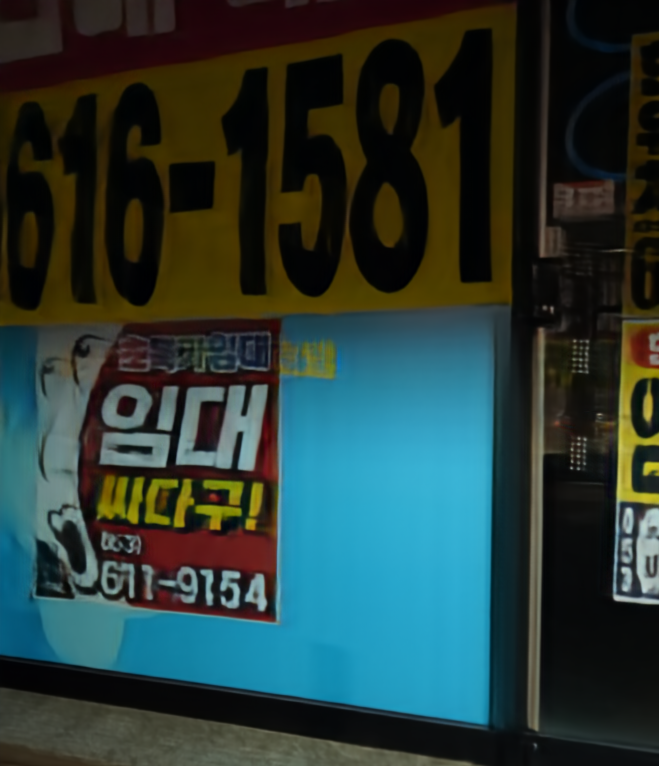}}
  \centerline{(c) SegDeblur-S (ours)}\medskip
\end{minipage}
\begin{minipage}[!ht]{.245\linewidth}
 \centering
 \centerline{\includegraphics[width=4.3cm]{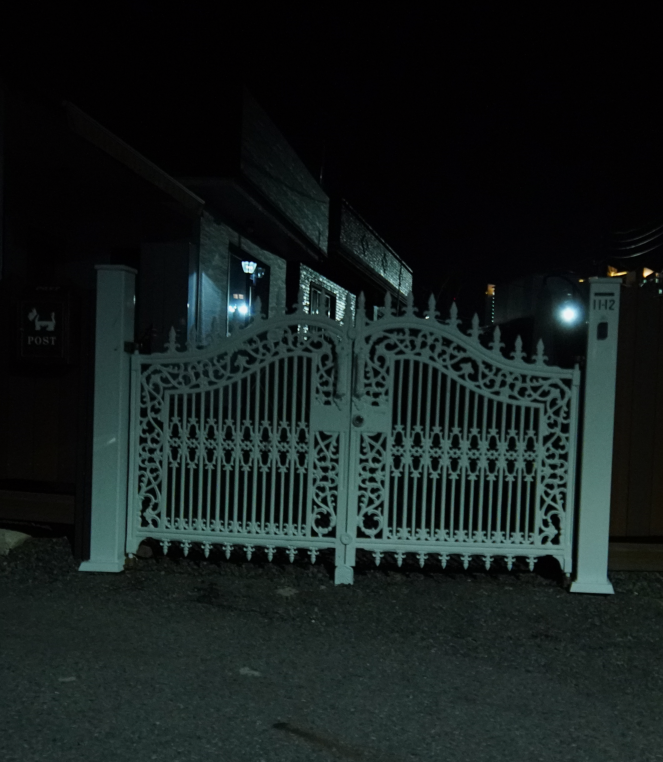}}
 \centerline{\includegraphics[width=4.3cm]{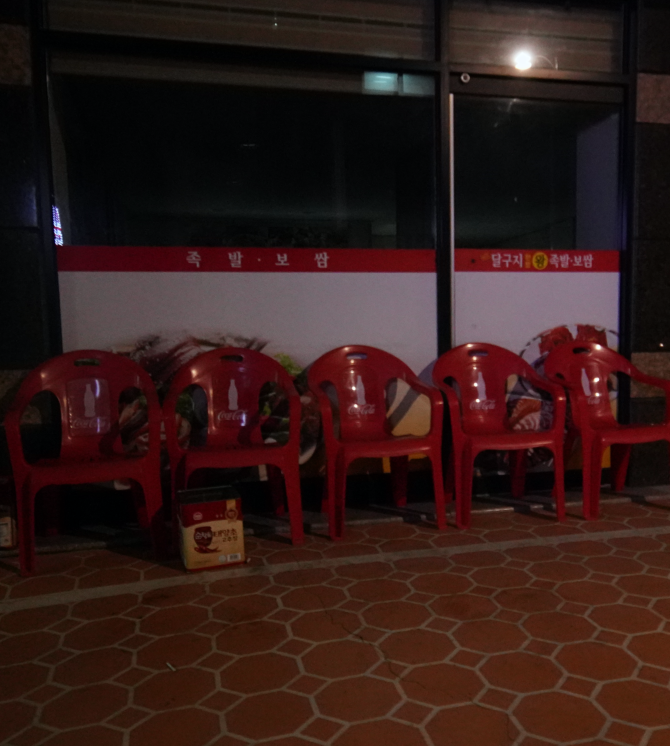}}
 \centerline{\includegraphics[width=4.3cm]{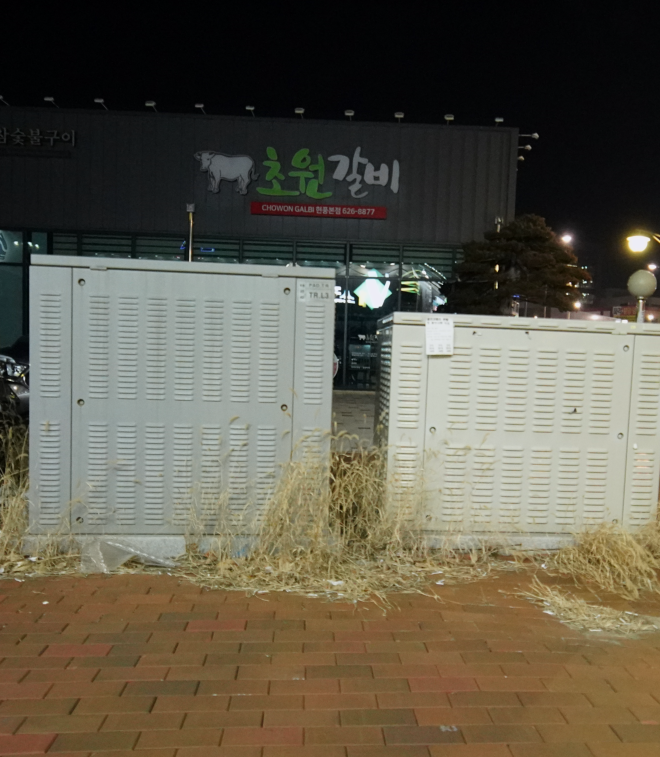}}
 \centerline{\includegraphics[width=4.3cm]{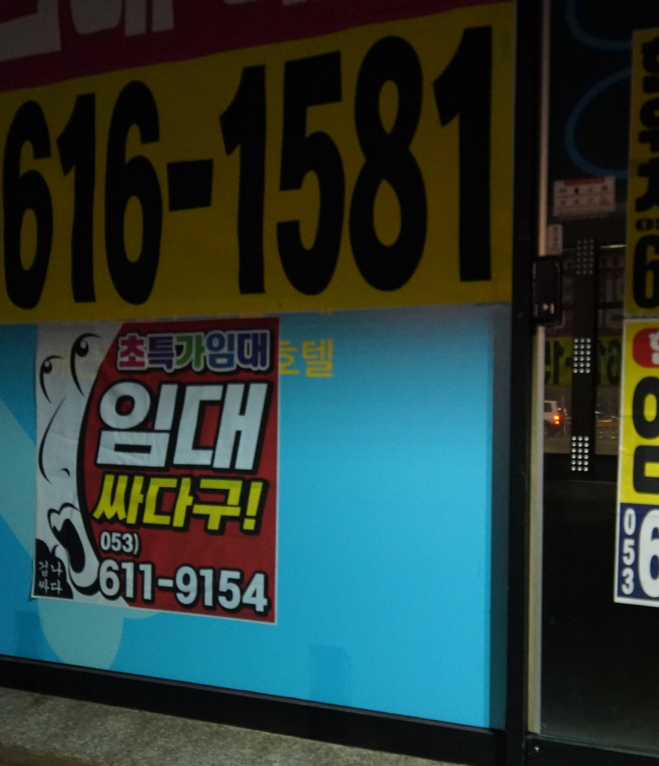}}
 \centerline{(d) Ground-Truth Sharp}\medskip
\end{minipage}
\caption{Qualitative results for efficient deblurring models in RealBlur~\cite{realblur}.}
\label{fig:supp_realblur1}
\vspace{-0.3cm}
\end{figure*}

\begin{figure*}[!ht]
  \centering
\begin{minipage}[!ht]{.48\linewidth}
  \centering
  \centerline{\includegraphics[width=7.5cm]{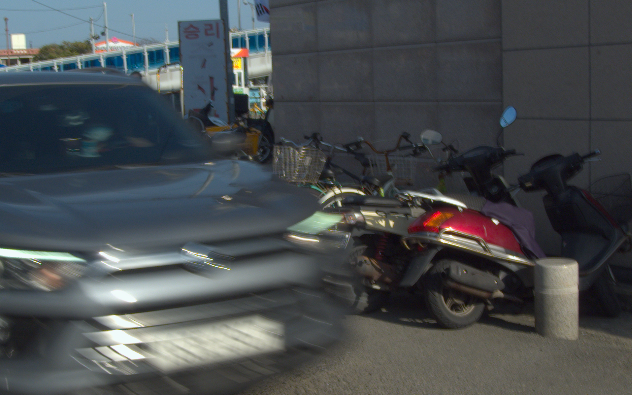}}
  \centerline{(a) Blur Input}\medskip
  \centerline{\includegraphics[width=7.5cm]{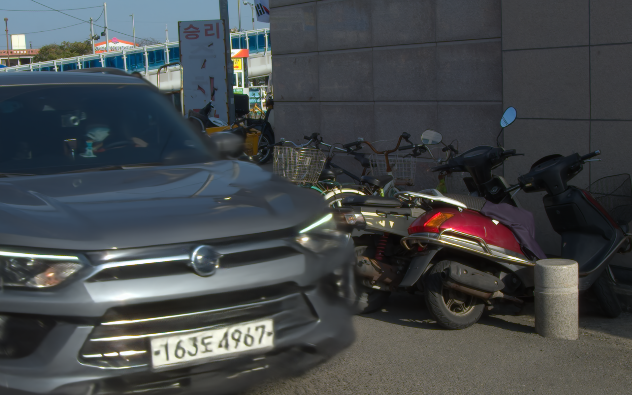}}
  \centerline{(c) SegDeblur-S (ours)}\medskip
\end{minipage}
\begin{minipage}[!ht]{.48\linewidth}
  \centering
  \centerline{\includegraphics[width=7.5cm]{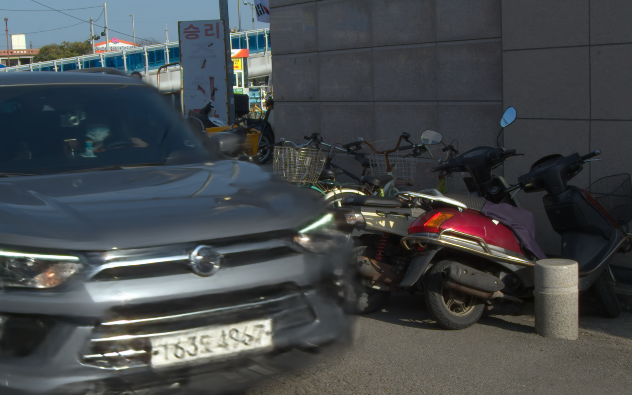}}
  \centerline{(b) FFTFormer-16~\cite{fftformer}}\medskip
  \centerline{\includegraphics[width=7.5cm]{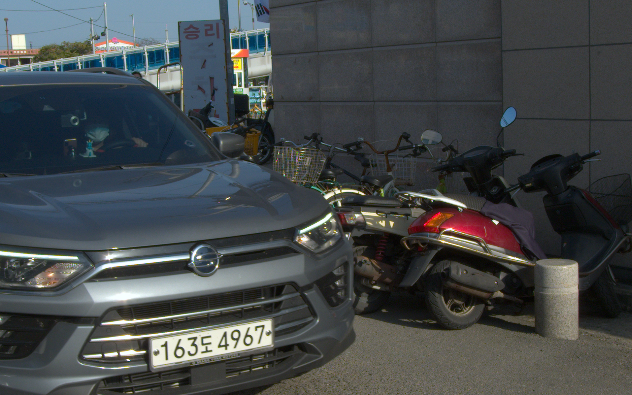}}
  \centerline{(d) Ground-Truth Sharp}\medskip
\end{minipage}
\end{figure*}
\begin{figure*}[!ht]
  \centering
\begin{minipage}[!ht]{.48\linewidth}
  \centering
  \centerline{\includegraphics[width=7.5cm]{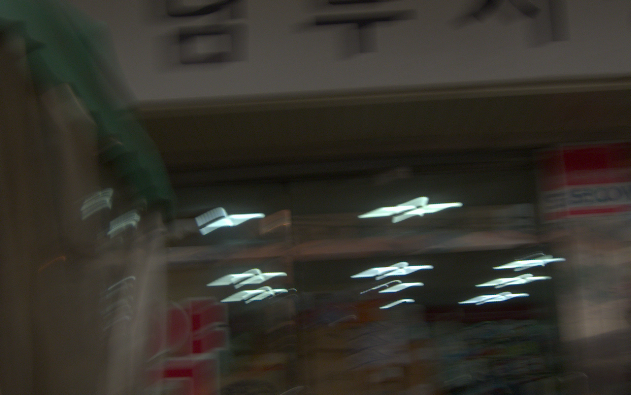}}
  \centerline{(a) Blur Input}\medskip
  \centerline{\includegraphics[width=7.5cm]{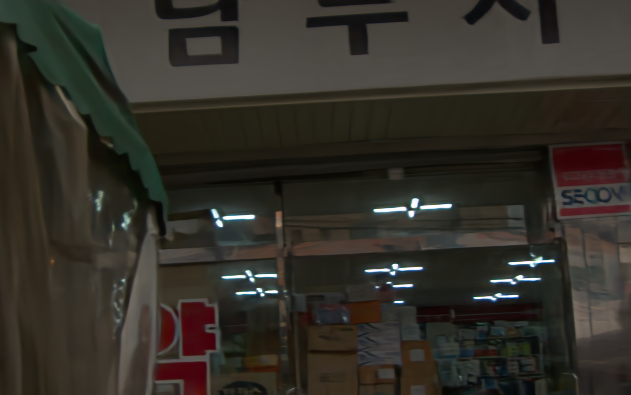}}
  \centerline{(c) SegDeblur-S (ours)}\medskip
\end{minipage}
\begin{minipage}[!ht]{.48\linewidth}
  \centering
  \centerline{\includegraphics[width=7.5cm]{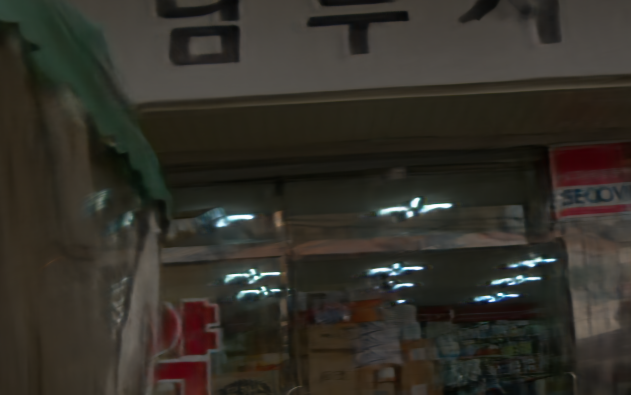}}
  \centerline{(b) FFTFormer-16~\cite{fftformer}}\medskip
  \centerline{\includegraphics[width=7.5cm]{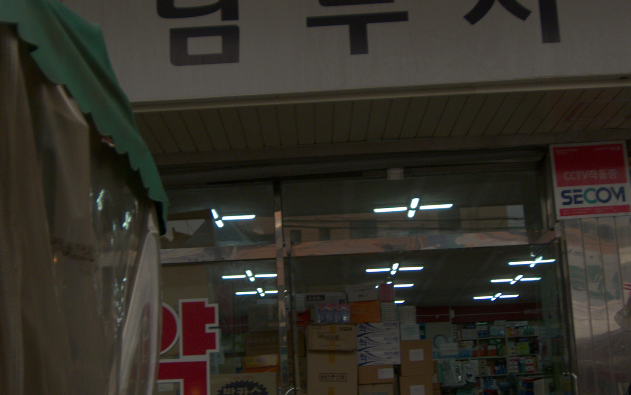}}
  \centerline{(d) Ground-Truth Sharp}\medskip
\end{minipage}
\label{fig:supp_rsblur1}
\vspace{-0.3cm}
\end{figure*}

\begin{figure*}[!ht]
  \centering
\begin{minipage}[!ht]{.48\linewidth}
  \centering
  \centerline{\includegraphics[width=7.5cm]{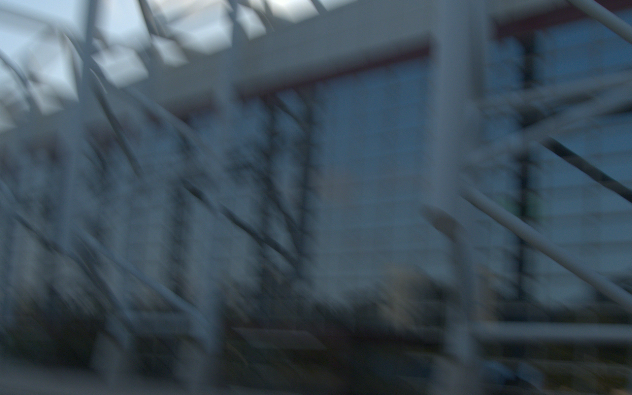}}
  \centerline{(a) Blur Input}\medskip
  \centerline{\includegraphics[width=7.5cm]{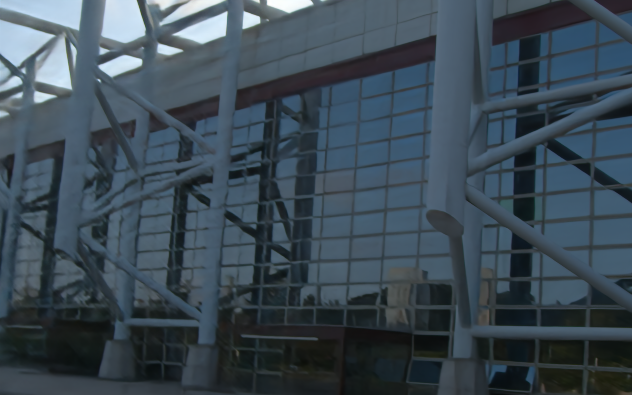}}
  \centerline{(c) SegDeblur-S (ours)}\medskip
\end{minipage}
\begin{minipage}[!ht]{.48\linewidth}
  \centering
  \centerline{\includegraphics[width=7.5cm]{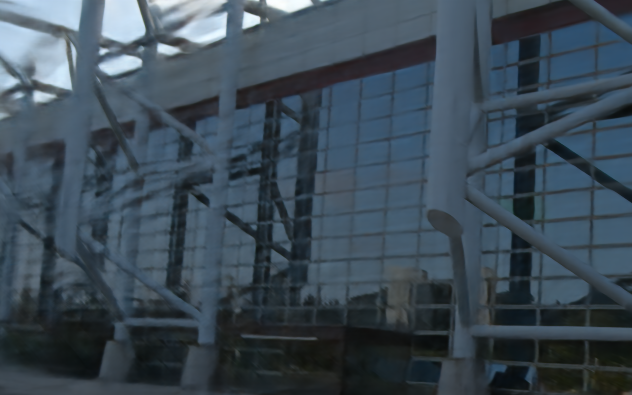}}
  \centerline{(b) FFTFormer-16~\cite{fftformer}}\medskip
  \centerline{\includegraphics[width=7.5cm]{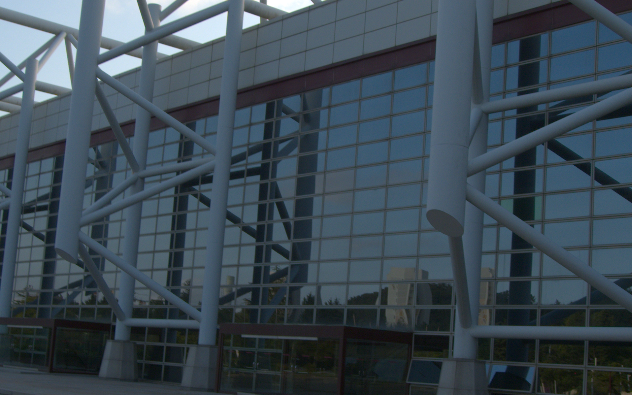}}
  \centerline{(d) Ground-Truth Sharp}\medskip
\end{minipage}
\end{figure*}
\begin{figure*}[!ht]
  \centering
\begin{minipage}[!ht]{.48\linewidth}
  \centering
  \centerline{\includegraphics[width=7.5cm]{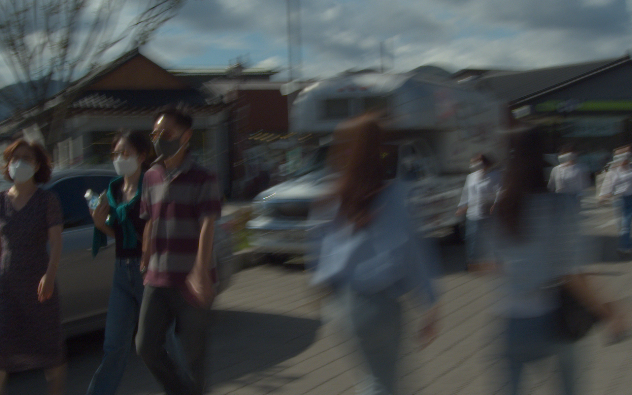}}
  \centerline{(a) Blur Input}\medskip
  \centerline{\includegraphics[width=7.5cm]{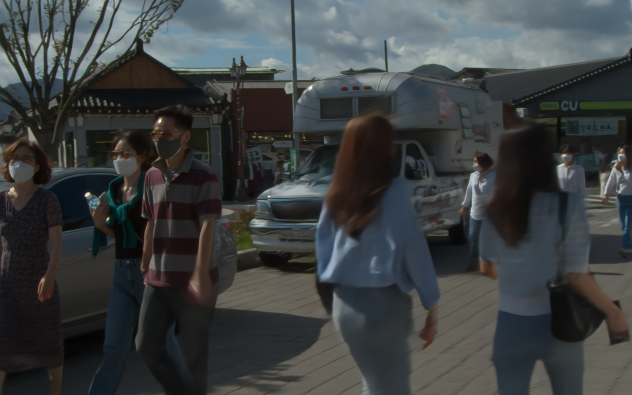}}
  \centerline{(c) SegDeblur-S (ours)}\medskip
\end{minipage}
\begin{minipage}[!ht]{.48\linewidth}
  \centering
  \centerline{\includegraphics[width=7.5cm]{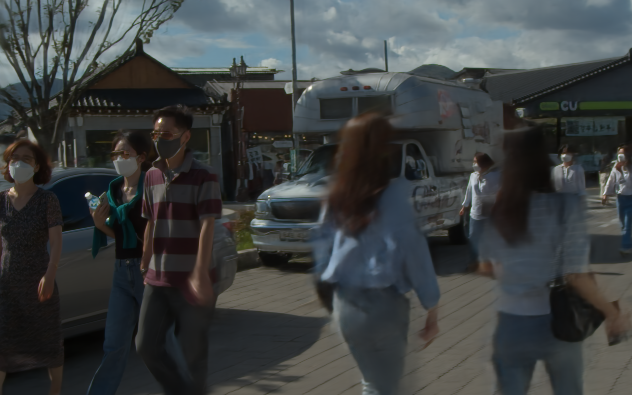}}
  \centerline{(b) FFTFormer-16~\cite{fftformer}}\medskip
  \centerline{\includegraphics[width=7.5cm]{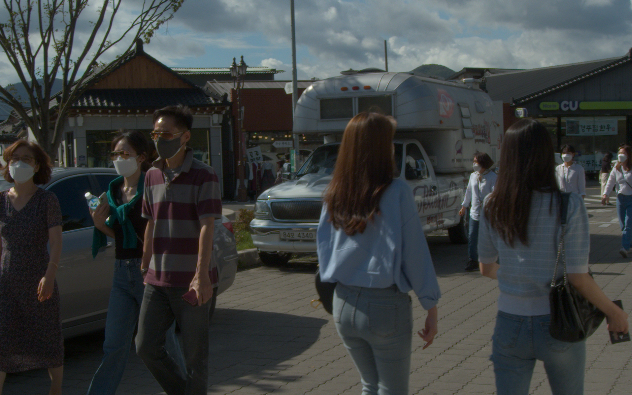}}
  \centerline{(d) Ground-Truth Sharp}\medskip
\end{minipage}
\caption{Qualitative results for efficient deblurring models in RSBlur~\cite{rsblur}.}
\label{fig:supp_rsblur2}
\vspace{-0.3cm}
\end{figure*}

\end{document}